%% file: main.tex
\documentclass[journal]{IEEEtran}
\usepackage{amsmath,amsfonts}
\usepackage{algorithmic}
\usepackage{algorithm}
\usepackage{array}
\usepackage[caption=false,font=normalsize,labelfont=sf,textfont=sf]{subfig}
\usepackage{textcomp}
\usepackage{stfloats}
\usepackage{url}
\usepackage{verbatim}
\usepackage{graphicx}
\usepackage{cite}
\usepackage{booktabs} 
\usepackage{multirow} 
\usepackage[table]{xcolor} 
\usepackage{adjustbox} 
\usepackage{bbm} 

\usepackage{silence}
\begin{document}

\title{Online Learning under Haphazard Input Conditions: A Comprehensive Review and Analysis}

\author{Rohit Agarwal, Arijit Das, Alexander Horsch, Krishna Agarwal, and Dilip K. Prasad
\thanks{R. Agarwal \{agarwal.102497@gmail.com\}, A. Horsch, K. Agarwal, and D.K. Prasad are with UiT The Arctic University of Norway, Tromsø.}
\thanks{A. Das is with the Indian Institute of Technology (Indian School of Mines), Dhanbad, India, and carried out this work as an intern at Bio-AI Lab, UiT The Arctic University of Norway, Tromsø.}
}



\maketitle

\begin{abstract}
The domain of online learning has experienced multifaceted expansion owing to its prevalence in real-life applications. Nonetheless, this progression operates under the assumption that the input feature space of the streaming data remains constant. In this survey paper, we address the topic of online learning in the context of haphazard inputs, explicitly foregoing such an assumption. We discuss, classify, evaluate, and compare the methodologies that are adept at modeling haphazard inputs, additionally providing the corresponding code implementations and their carbon footprint. Moreover, we classify the datasets related to the field of haphazard inputs and introduce evaluation metrics specifically designed for datasets exhibiting imbalance. The code of each methodology can be found at \url{https://github.com/Rohit102497/HaphazardInputsReview}. 
\end{abstract}



\section{Introduction}

Over the past few years, online learning has garnered significant attention in modeling real-life applications that produce streaming data \cite{8955936}. It is assumed that the input feature space in such an application is always constant. Nonetheless, a plethora of applications exist --- ranging from urban disaster monitoring systems to movie review classification and email spam detection --- that generate dimension-varying streaming data (see Figure \ref{fig:onlinelearning}). These types of inputs have been designated as haphazard inputs \cite{agarwal2023auxdrop}. This paper presents a comprehensive survey of the literature on haphazard inputs, aiming to propel further advancements within this sphere of research.

The literature features a limited number of publications that address the problem of dimension-variant inputs. Among these, Beyazit et al. \cite{beyazit2019online} introduced the problem as the varying feature space, wherein the feature space of the input data keeps changing. He et al.\cite{he2019online} introduced it as a problem of the capricious data streams with an arbitrarily varying feature space. Agarwal et al.\cite{agarwal2023auxdrop} further refined the problem by defining it as `haphazard inputs', where the dimension of the input varies at each time instance without any prior information about the future data. Furthermore, they delineate six characteristics of haphazard inputs: streaming data, missing data, missing features, obsolete features, sudden features, and an unknown number of total features. In this survey, we adopt the haphazard input terminology because of its clear definition, description of different characteristics, and segregation from other research domains \cite{agarwal2023auxdrop}.

\input{FigureTex/Fig_OnlineLearning}

\textbf{Motivation:} The field of haphazard inputs is not new; its genesis can be traced back to the foundational research conducted in 2005 by Katakis et al. \cite{katakis2005utility}. Despite its established presence, the models designed to address haphazard inputs are scattered sporadically within the literature. A recent survey paper by He et al. \cite{he2023towards} provides a succinct analysis of six models, comparing them across four datasets with a focus on feature apportionment strategies. Despite this contribution, there remains an evident gap in the literature concerning several critical aspects. These include comprehensive dataset categorization, the inclusion of pertinent datasets, the examination of additional models, the utilization of more effective metrics, and, crucially, the establishment of benchmarks and the promotion of open-source resources. Addressing these gaps is essential for advancing the field and facilitating the development of more robust and generalized models. These identified gaps motivated us to write this survey paper. Furthermore, this survey extends by analyzing the carbon footprint of models, discussing allied subfields, suggesting models from other fields adaptable for haphazard inputs, and outlining future research directions and potential applications.

This review adheres, albeit in a non-rigid manner, to the Preferred Reporting Items for Systematic Reviews and Meta-Analyses (PRISMA) guidelines \cite{page2021prisma} for identifying the relevant literature. The PRISMA flowchart is presented in Figure \ref{fig:prisma} which elucidates the stages of relevant literature identification and the number of articles studied for this survey. An extensive keyword search was conducted on Google Scholar to ascertain relevant studies. The references of these articles, as well as the studies citing them, were meticulously examined to uncover additional relevant literature. This iterative process was pursued exhaustively.  The first literature search was conducted over a period from the 20th to the 25th of September, 2023. Consequently, 16 studies were identified, among which 9 models have been recognized as capable of processing haphazard inputs: NB3\cite{katakis2005utility}, FAE \cite{wenerstrom2006temporal}, DynFo \cite{schreckenberger2022dynamic}, ORF\textsuperscript{3}V \cite{schreckenberger2023online}, OLVF \cite{beyazit2019online}, OCDS \cite{he2019online}, OVFM \cite{he2021online}, Aux-Net \cite{agarwal2023auxiliary}, and Aux-Drop \cite{agarwal2023auxdrop}. We performed the final literature search on March 8th, 2024, to identify any new methods that emerged in the last few months. This search yielded three articles: DCDF2M \cite{sajedi2024data}, OIL \cite{lee2023study}, and OLCF \cite{zhou2023online}. We chose to exclude these methods from our analysis due to factors such as the unavailability of code and their deviation from the defining characteristics of haphazard inputs. We provide a concise summary of these methods in Supplementary Section III.

\input{FigureTex/Fig_PrismaGuideline}

\subsection{Notable Contributions of This Survey}
\begin{enumerate}
    \item{\textit{Model Taxonomy:} We introduce a novel classification schema for models adept at processing haphazard inputs. This taxonomy divides models into four distinct categories predicated upon their algorithmic approach.}
    \item{\textit{Dataset Taxonomy:} We systematically categorized the datasets into three types based on the number of instances they contain, facilitating a more structured approach to dataset analysis.}
    \item{\textit{Imbalance Metrics:} We adopt robust metrics --- AUROC, AUPRC, and balanced accuracy --- for the comparative evaluation of models on imbalanced datasets. It is the inaugural application of these metrics within the haphazard inputs field.}
    \item{\textit{Overall Comparison Metrics:} We propose 5 comprehensive metrics for the overall comparison of methodologies. These encompass performance, data scalability, prediction consistency, speed, and feature scalability.}
    \item{\textit{Benchmark:} We establish a benchmark for all examined models, accompanied by the open-sourcing of the associated codes, thereby contributing to the reproducibility and transparency of research.}
    \item{\textit{Carbon Footprint:} We assess the carbon footprint of individual models and the collective environmental impact of our research, thereby supporting the United Nations' Sustainable Development Goals (SDGs) \cite{UNSDG}.}
    \item{\textit{Possible Architecture Ideas:} We elucidate potential architectural concepts for deep learning models that may be tailored to effectively handle haphazard inputs.}
    \item{\textit{Subfields:} We explore the subfields related to haphazard inputs, providing insights into their interconnection with the primary field of study.}
    \item{\textit{Possible Applications:} We articulate a vision for the potential applications of haphazard inputs, with the aspiration that the research community will continue to advance this domain.}
\end{enumerate}

\subsection{Outline of This Article}
The outline of the rest of this article is as follows: Section \ref{sec:problemstatement} presents the problem of haphazard inputs, Section \ref{sec:datasets} explores the datasets referenced in the literature, along with their categorization, Section \ref{sec:metrics} outlines the metrics employed for the comparative analysis of the models, Section \ref{sec:models} discusses various models within the scope of their categorization, Section \ref{sec:experiments} details the experimental procedures and their results, Section \ref{sec:observations} provides some observations based on the results, Section \ref{sec:adaptablemodels} presents ideas on adapting models from other fields models to address haphazard inputs, Section \ref{sec:subfield} delves into the subfields associated with haphazard inputs, Section \ref{sec:applications} introduces potential applications for haphazard inputs, Section \ref{sec:future_directions} explores future research directions, and Section \ref{sec:conclusion} provides the concluding remarks of the paper.

\section{Haphazard Inputs}
\label{sec:problemstatement}

Haphazard inputs refer to the problem of dimension-varying input streams in an online setting. It can have the following six characteristics: (1) \textit{Streaming data}: The data arrives sequentially and continuously throughout time. It cannot be stored and needs to be processed as soon as it arrives leading to an online setting \cite{hoi2021online}. (2) \textit{Missing data}: Some of the known features/sensors do not contain any information in some instances \cite{emmanuel2021survey}. This data may be imputed by inducing noise and bias. (3) \textit{Missing features}: Some features do not arrive from the outset but it is known that they will be available in future instances. In some cases, prior information regarding these features may be available. (4) \textit{Sudden features}: No prior information about these features exists and hence imputation of any form cannot be employed for them. (5) \textit{Obsolete features}: These features stop arriving from a specific time instance, but the cessation of their arrival remains unknown. (6) \textit{Unknown number of features}: The potential for an indeterminate number of sudden features to emerge, coupled with the possibility of existing features becoming obsolete, results in an unknown total number of features.

Mathematically, haphazard inputs can be represented by $X_t \in R^{D_t}$, where $D_t$ is the dimension of the inputs at time $t$. The mapping function between the input and the output is defined by $f_t: X_t \rightarrow y_t$, where $f_t$ is the learnable function (or model) that is tasked with predicting the output $y_t$, based on the haphazard input $X_t$. Since the model ($f_t$) learns in an online setting, $f_t$ produces an output $\hat{y}_t$ as $f_t(X_t) = \hat{y}_t$ at time $t$. The model incurs an instantaneous loss $l_t = L(y_t, \hat{y}_t)$, where $L$ is the loss function determining the closeness between the prediction and the ground truth. Subsequently, the model updates its parameters based on $l_t$, thereby yielding the updated model $f_{t+1}$ for the next time instance. This process is demonstrated in Figure \ref{fig:onlinelearning}(a).

Let us denote the set of features available at time $t$ by $\mathbb{F}_t$ and the set of all the features whose information is available up to and including time $t$ by $\Bar{\mathbb{F}}_t$. There can be many relationships between $\mathbb{F}_t$ and $\Bar{\mathbb{F}}_t$. For instance, if $\mathbb{F}_t \cap  \Bar{\mathbb{F}}_{t-1} = \phi$, it implies that all features in $\mathbb{F}_t$ are sudden features. If $\mathbb{F}_t \cap \Bar{\mathbb{F}}_{t-1} = \mathbb{F}_t$, this indicates that all features in $\mathbb{F}_t$ were seen before. In the scenario where $\mathbb{F}_t \cap \Bar{\mathbb{F}}_{t-1} = \Bar{\mathbb{F}}_{t-1}$, it is understood that all features encountered prior to time $t$ are still present, with the potential addition of new sudden features.  These are merely illustrative examples, and other configurations of the relationship between $\mathbb{F}_t$ and $\Bar{\mathbb{F}}_t$ are also possible. To elucidate the aforementioned potential characteristics of haphazard inputs, an example is presented in Figure \ref{fig:CharacteristicsExample}.

\input{FigureTex/Fig_CharacteristicsExample}

In Figure \ref{fig:CharacteristicsExample}, at the initial time instance 1, $\mathbb{F}_1 = \{F_1, F_2, F_4\}$ and $\Bar{\mathbb{F}}_1 = \{F_1, F_2, F_3, F_4\}$. Note that, despite the availability of prior information, $F_3$ is absent from the current feature set; this absence classifies it as a missing feature. The model can impute $F_3$ on the cost of potentially introducing bias and noise. At time instance 2, $\mathbb{F}_2 = \{F_1, F_2, F_4\}$. Thus $\mathbb{F}_2 \cap \Bar{\mathbb{F}}_{1} = \mathbb{F}_2$ indicating that all features in $\mathbb{F}_2$ were previously observed. At time instance 3, the feature set $\mathbb{F}_3 = \{F_1, F_3\}$, which implies that features $F_2$ and $F_4$ are not present and thus constitute missing data at this particular juncture. Proceeding to time instance 4, $\mathbb{F}_4 = \{F_1, F_2, F_3, F_5\}$, where $F_5$ emerges as a sudden feature, having no prior information. It is also relevant to observe that feature $F_4$ has become obsolete, as it is no longer received beyond this point, and there is no indication of its obsolescence. Additionally, it is underscored that the actual total number of features at each time step remains unknown.

The disparate characteristics of haphazard inputs present a variety of challenges, both individually and collectively. The issue of missing data may be addressed through imputation techniques; however, it is critical to acknowledge that such techniques might introduce bias into the dataset. Furthermore, in situations where a feature is consistently absent across numerous instances, the imputation process may predominantly result in the incorporation of noise. The missing feature can be imputed by leveraging prior information, although the resultant noise could be substantial. Obsolete features may be subject to extrapolation; however, this approach may lead to a degradation in efficacy over time as the extrapolated values become increasingly less representative of the underlying data distribution \cite{hahn1977hazards}. Sudden features could be initially modeled as Gaussian noise until their actual values become available, but this is likely to result in sub-optimal model performance. The uncertainty surrounding the total number of features complicates the use of fixed online models, as they would require the pre-definition of a maximum number of features, potentially leading to an influx of noise or the exclusion of relevant features. The limitations of imputation, extrapolation, and the use of Gaussian noise as strategies to contend with the characteristics of haphazard inputs are exemplified in Table \ref{tab:comparisonOfmodels}, which illustrates the ineffectiveness of these techniques in handling the nuanced complexities of haphazard inputs.

\input{Tables/Tab_CharacterisitcsOfHaphazardInputs}

\section{Datasets}
\label{sec:datasets}

This survey encompasses an examination of binary class datasets that have been employed to train and evaluate existing methodologies. A total of 20 datasets are considered within the scope of this analysis. The description of the datasets is presented in Table \ref{tab:datasetdescription}. It is noteworthy that a majority of the datasets exhibit class imbalances, a factor that is captured within the aforementioned table. Among these datasets, 5 are identified as real data, which inherently possesses the characteristics of being haphazard in nature. The other 15 datasets are synthetic, a classification attributed to datasets wherein values have been deliberately omitted to simulate the condition of haphazard inputs. We will discuss in section \ref{sec:dataset_preparation}, the process of transforming synthetic datasets to create haphazard inputs. We also discuss the creation of datasets which are in raw form in section \ref{sec:dataset_creation}.

We systematically categorized the datasets based on the number of instances they contain. A dataset is classified into one of the following three categories: Small ($\leq$10k instances), Medium (10k $<$ instances $\leq$ 100k), and Large ($>$ 100k instances). In accordance with this taxonomy, 14 datasets have been classified as Small, 3 as Medium, and 3 as Large datasets, as outlined in Table \ref{tab:datasetdescription}. The link to each dataset is presented in Section IV of the Supplementary document.

Within the compendium of datasets documented in the literature, this survey elects to exclude 3 datasets that were previously analyzed in OCDS \cite{he2019online}, specifically the splice, HAPT, and dna datasets. The rationale for their exclusion is twofold. Firstly, the splice dataset is designed for a ternary classification task, while the HAPT dataset involves twelve distinct classes. To maintain a focus on binary classification tasks and ensure methodological consistency, these multi-class datasets are deemed incongruent with the objectives of this survey. Secondly, upon closer inspection, certain ambiguities have been identified within the dna dataset, which have prompted the decision to omit it from the current analysis.

\subsection{Dataset Creation}
\label{sec:dataset_creation}

Some datasets needed to be transformed from their raw form. Here we discuss these datasets and their transformations.

\paragraph{crowdsense} The crowdsense \cite{camprodon2019smart} collection originally comprises eight distinct datasets, each sharing identical input attributes, but annotated with divergent sets of labels. To align with the binary classification framework of this survey, we selectively incorporated two datasets from the crowdsense series that are characterized by binary labels — specifically, the third and the fifth datasets. Consequently, these datasets are referred to as crowdsense(c3) and crowdsense(c5), respectively. In their raw form, both datasets encompass 789 instances alongside 1086 features. However, upon meticulous examination, it was discerned that 132 features and 3 instances were devoid of any data, prompting their exclusion from the datasets. Post-curation, the refined datasets --- crowdsense(c3) and crowdsense(c5) --- each consist of 786 instances, and 954 features, and are distinctly binary in class structure.

\paragraph{spamassassin} The spamassassin dataset provides emails paired with corresponding labels that categorize each email as either spam or non-spam. We extract the date and body from each email, followed by converting all text to lowercase, substituting URLs and numbers with a generic string placeholder, and removing punctuation. During this process, we discovered that 5 emails did not contain any information, therefore, we dropped them. The remaining emails were then chronologically ordered based on their timestamp. The body of the email is then vectorized, resulting in a high-dimensional feature space consisting of 152257 attributes. To further refine the dataset, we select the 7500 most frequently occurring features for inclusion in the final analysis.

\input{Tables/Tab_DatasetDescription4}

\paragraph{imdb} The imdb dataset \cite{maas2011learning} comprises a total of 50000 instances, evenly divided into two subsets: training and testing, each containing 25000 instances. In accordance with established conventions in the literature, this study focuses on the training subset. Furthermore, attention is directed toward the 7500 most prevalent features within the imdb dataset, which are represented by the initial 7500 rows. This approach aligns with the standard practices of feature selection to ensure comparability and relevance to the field of study.

\paragraph{diabetes\_us} In the original diabetes\_us dataset \cite{strack2014impact}, missing or unrecorded values are denoted by the symbols "?", "nan", and "-1". A multitude of attributes with categorical data characterizes this dataset. To facilitate analysis, these categorical values are converted into numerical equivalents through the following procedures: \textit{(1) age:} The dataset presents age in the bracket of 10s ([0,10], [10,20], ...). We adopt the median value of each bracket as a representative figure and replace the categorical bracket with this value. \textit{(2) Weight:} Similarly to age, weight is categorized in intervals of 25. The median value of each interval is used to substitute the categorical bracket. \textit{(3) 34 Features:} There are 34 attributes within the dataset that are inherently categorical. These are systematically transformed into numerical values by assigning a unique integer, ranging from 1 up to the total number of categories present within each respective feature. This numeric encoding allows for the quantitative analysis of previously qualitative data.

\section{Metrics}
\label{sec:metrics}
The model efficacy is predominantly quantified through two metrics: the total number of errors and the overall accuracy. This study employs these metrics to facilitate comparative analysis among models. Additionally, certain articles reference the cumulative error rate (CER) \cite{he2021online} and the loss value \cite{agarwal2023auxiliary} as comparative indices. Given that CER is essentially the complement of accuracy (CER = 1 - accuracy), this analysis will solely report accuracy to avoid redundancy. The calculation of loss is contingent upon the distinct loss functions inherent to each model; therefore, loss metrics are excluded from our survey. As shown in Table \ref{tab:datasetdescription}, a class imbalance exists within many datasets. This necessitates the inclusion of the Area Under the Precision-Recall Curve (AUPRC), the Area Under the Receiver Operating Characteristic Curve (AUROC), and balanced accuracy as additional comparative metrics. A description of each metric is given below:

\input{FigureTex/Fig_ModelEmergence}

\paragraph{Number of Errors} The number of wrongly classified instances is considered the number of errors, and is given by
\begin{equation}
    \text{Number of Errors} = N - \sum_{t=1}^{N} \mathbbm{1}_{y_t}(\hat{y}_t),
\end{equation}
where $N$ is the number of instances and $\mathbbm{1}$ is the indicator function denoting
\begin{equation}
    \mathbbm{1}_{y_t}(\hat{y}_t) = 
    \begin{cases}
        1 & \text{if } \hat{y}_t = y_t, \\
        0 & \text{otherwise}.
    \end{cases}
\end{equation}
    
\paragraph{Accuracy} The accuracy of a model is quantified as the number of correct predictions divided by the total number of instances. Following the notations from the previous paragraph, accuracy can be mathematically represented as
\begin{equation}
    \text{Accuracy} = \frac{\sum_{t=1}^{N} \mathbbm{1}_{y_t}(\hat{y}_t)}{N} 
\end{equation}

\paragraph{AUROC} AUROC \cite{branco2016survey} serves as an indicator of a model's capacity to discriminate between positive and negative classes. In the context of binary classification, varying the decision threshold yields a spectrum of True Positive Rates (TPR) and False Positive Rates (FPR), which together constitute the Receiver Operating Characteristic (ROC) curve. The area under this ROC curve gives AUROC.

\paragraph{AUPRC} AUPRC \cite{saito2015precision} is similar to AUROC, however, it is derived from a distinct graphical representation, namely the precision-recall (PR) curve, as opposed to the TPR versus FPR plot utilized in the AUROC. The AUPRC metric is predominantly utilized in scenarios involving imbalanced datasets, where the primary concern is the accurate identification of positive labels. 

\paragraph{Balanced Accuracy} It is the arithmetic mean of specificity and sensitivity \cite{5597285}. It is a robust metric in the presence of imbalanced datasets. A model that demonstrates superior efficacy on the majority class at the expense of the minority class may present a deceptively high overall accuracy. However, such a model would yield a diminished balanced accuracy, as this measure would be adversely impacted by the lower value of either sensitivity or specificity, reflecting the model's inadequacy in one of the classes.

\section{Models}
\label{sec:models}

The models are classified according to the algorithmic strategies they implement to address the haphazard inputs. Initial efforts date back to 2005-06 to model haphazard inputs via naive Bayes \cite{john1995estimating}. More recently, the field has gained traction, with researchers actively developing models for haphazard inputs utilizing linear classifiers \cite{zhang2002recommender}, decision stumps \cite{iba1992induction}, and deep learning algorithms \cite{lecun2015deep}. We have documented the evolution of these models in a timeline showcased in Figure \ref{fig:timeline}. This progression of research exhibits clustering of efforts, beginning with naive Bayes and subsequently transitioning to linear classifiers, then to decision stumps, and most recently, to the incorporation of deep learning.

\paragraph{Qualitative Comparison} We present a qualitative comparison of the models based on various factors in Table \ref{tab:modelcomparison}. We examine each model's foundational assumptions and their ability to reconstruct unavailable features. We also elucidate the deterministic nature of the models by assessing the impact of randomness. A non-deterministic model will yield variable outcomes with different initial seeds, necessitating multiple executions to accurately evaluate its efficacy. In contrast, a deterministic model consistently produces identical results, regardless of the seed value. We found that most models except OVFM and Aux-Drop lack open-source availability. To address this, we provide codes for these models, namely, NB3, FAE, OLVF, OCDS, DynFo, ORF\textsuperscript{3}V, and Aux-Net. Additionally, we have updated the OVFM code to rectify issues stemming from outdated Python libraries. Finally, we provide an overall qualitative comparison across 5 different breadths --- performance, data scalability, prediction consistency, speed, and feature scalability --- and discuss them in detail in Section \ref{sec:results}. Next, we discuss each of the models in the haphazard inputs field.

\input{Tables/Tab_classificationbasedonassumption_results_2}

\subsection{Naive Bayes}

Naive  Bayes is a simple probabilistic classification algorithm based on the Bayes theorem \cite{mcnamara2006bayes} coupled with strong assumptions of independence between the features \cite{hastie2009elements, lowd2005naive}. It is based on conditional probability and calculates the probability of each class given the input features as $P(c_j|x_1, ..., x_n)$ where $c_j$ is the $j^\text{th}$ class, $x_i$ is the $i^\text{th}$ feature and $n$ is the total number of features. The class with the highest probability is designated as the prediction class given by $\arg\max_j P(c_j|x_1, ..., x_n)$. Considering Bayes theorem, the conditional probability is given by:
\begin{equation}
    \label{eq:bayesalgo}
    P(c_j|\mathbf{x}) = \frac{P(c_j)P(\mathbf{x}|c_j)}{P(\mathbf{x})},
\end{equation}
where $\mathbf{x} = (x_1, ..., x_n)$ is a vector. The product of prior $P(c_j)$ and likelihood $P(\mathbf{x}|c_j)$ can be given by the joint probability as $P(c_j, x_1, ..., x_n) = P(c_j)P(\mathbf{x}|c_j)$. This can be further simplified using the chain rule of conditional probability and considering the assumption that all the $x_i$'s are mutually independent as $P(c_j)\prod_{i=1}^{n} P(x_i|c_j)$. Thus, designing a classifier involves predicting the highest probable class, \emph{i.e.}, $\arg\max_j P(c_j|\mathbf{x})$. Since $P(\mathbf{x})$ is constant, the naive Bayes classifier can be given by
\begin{equation}
    \label{eq:naivebayes}
    \hat{y} = \arg \max_j P(c_j)\prod_{i=1}^{n} P(x_i|c_j)
\end{equation}

NB3 \cite{katakis2005utility} used naive Bayes coupled with $\chi^2$ statistics \cite{miller1982maximally} to model haphazard inputs. Specifically, they modeled textual data streams to classify an email as spam or legitimate. Each word in a mail is considered a feature and the number of times a feature is seen in the mail is taken as the feature value. The statistics of all the features seen thus far are stored. Whenever a new mail arrives, the statistics of new features in this mail are stored for the future. Since it is high-dimensional data, $\chi^2$ statistics are employed to select top $n$ features for classification. Information gain \cite{forman2003extensive} and mutual information \cite{yang1997comparative} can also be used in place of $\chi^2$ statistics. Based on the top $n$ features, eq. \ref{eq:naivebayes} gives a final prediction where 
\begin{gather}
    P(c_j) = \frac{\text{Number of $c_j$ emails}}{\text{Number of emails}}, \text{and}\\
    P(x_i|c_j) = \frac{\text{Number of $x_i$ in all $c_j$ emails}}{\text{Number of $c_j$ emails}}.
\end{gather}
Thus using $\chi^2$ statistics and naive Bayes classifier, NB3 \cite{katakis2005utility} dynamically adds features, updates current features statistics, and selects a feature subset to handle haphazard inputs.

Another addition to the naive Bayes category is Feature Adaptive Ensemble (FAE) \cite{wenerstrom2006temporal}. FAE is an incremental approach, drawing upon the foundational concept of NB3, which capitalizes on the naive Bayes classifier. Contrary to NB3, FAE does not rely on a singular classifier; instead, it implements an ensemble of naive Bayes classifiers to make predictions. This model dynamically discards outdated learners and incorporates new ones that align with the emergent features introduced by streaming data. At each time instance, FAE determines the final prediction by identifying the class that accumulates the maximum aggregate weight across all learners.

Naive Bayes models operate deterministically, predicated on the assumption that features are independent. These models require pretraining with a set of initial instances.

\subsection{Linear Classifiers}

A linear classifier \cite{zhang2002recommender} is a classification method designed as a linear combination of features. If $\mathbf{x}$ is a feature vector, then the predicted output is given by 
\begin{equation}
    \hat{y} = h(W\cdot\mathbf{x}),
    \label{eq:linear}
\end{equation}
where $W$ is the weight vector and $\cdot$ is the scalar product. The threshold function $h$ maps $W\cdot\mathbf{x}$ to an output class as
\begin{equation}
    h(W\cdot\mathbf{x}) = 
    \begin{cases}
        1 & \text{if } W\cdot\mathbf{x} > \theta, \\
        0 & \text{otherwise},
    \end{cases}
    \label{eq:linear_thershold}
\end{equation}
where $\theta$ is the scalar threshold. The optimal values of $W$ are determined by the Empirical Risk Minimization principle \cite{vapnik1991principles}. There are three methods in this category: Online Learning from Varying Feature Spaces (OLVF), Online Learning from Capricious Data Streams (OCDS), and Online Learning in Variable Feature Spaces with Mixed Data (OVFM).

OLVF \cite{beyazit2019online} partitions the features in incoming instances into three distinct feature spaces: existing, shared, and new. The existing feature space is exclusive to prior instances, while the new feature space is unique to the current instance. The shared feature space encompasses features present in both previous and current instances. OLVF adopts empirical risk minimization complemented by adaptive constraints for the training of its linear weights. The model employs the shared feature space for making predictions, maintaining a threshold ($\theta$) value of 0 in equation \ref{eq:linear_thershold}.

OCDS \cite{he2019online} operates under the premise that there exists a relatedness among features. The model utilizes graph-based methods to infer unobserved features from those that are observed. It combines the prediction derived from the observed features ($\mathbf{x_t}$) and unobserved features ($\mathbf{\Tilde{x}_t}$) to produce an ensemble prediction. This process is mathematically expressed as $\hat{y}_t = h(kW_t\cdot\mathbf{x_t} + (1-k)\Tilde{W}_t\cdot\mathbf{\Tilde{x}_t})$, where $k$ is the parameter that modulates the influence of both $\mathbf{x_t}$ and $\mathbf{\Tilde{x}_t}$, and the threshold ($\theta$) value is set to 0.

OVFM \cite{he2021online} examines the challenge of haphazard inputs within a feature space comprising mixed data types, including Boolean, ordinal, and continuous variables. While bearing similarities to OCDS, OVFM distinguishes itself in the methodology employed to learn unobservable features. Specifically, OVFM implements a Gaussian copula \cite{hoff2007extending} to capture feature relatedness. OVFM also sets the threshold ($\theta$) value to 0.

\input{FigureTex/Fig_DecisionStump}

OLVF operates without any specific assumptions, while OCDS and OVFM proceed on the premise that features exhibit relatedness. This underlying assumption enables both models to reconstruct features when they are missing. Although this process resembles imputation, it allows the linear classifier to manage haphazard inputs effectively. It is important to acknowledge, however, that feature reconstruction may introduce noise, particularly with features that have become obsolete. Additionally, OVFM's reliance on a storage buffer for feature reconstruction constitutes a limitation, as the ideal online learning setting precludes the retention of data. Further critical points concerning OLVF and OCDS are discussed below.
\paragraph{OLVF} It comprises two major components, namely, the instance classifier and the feature space classifier. The instance classifier is responsible for delivering a prediction, while the feature space classifier aids the instance classifier by compensating for the loss of information attributable to the varying feature space. The weight update of the feature space classifier given in eq. 7 and 8 of OLVF \cite{beyazit2019online} is as follows:
\begin{equation}
    \label{eqn:olvfweightupdate}
    \begin{gathered}
         \Bar{w}^e = \Bar{w}^e_t  \\
         \Bar{w}^s = \Bar{w}^s_t - 
                        \tau
                        \frac{
                            log(
                                e^{
                                    -{I(y_t, \hat{y}_t)}
                                    (\Bar{w}^s_t\mathbb{R}^{x^s_t} + \Bar{w}^n_t\mathbb{R}^{x^n_t})
                                    }
                                )
                            }
                            {
                            log(
                                1 + 
                                e^{
                                    -{I(y_t, \hat{y}_t)}
                                    (\Bar{w}^s_t\mathbb{R}^{x^s_t} + \Bar{w}^n_t\mathbb{R}^{x^n_t})
                                    }
                            )
                            } 
                        I(y_t, \hat{y}_t)
                        \mathbb{R}^{x^s_t} \\
         \Bar{w}^n = \tau
                        \frac{
                            log(
                                e^{
                                    -{I(y_t, \hat{y}_t)}
                                    (\Bar{w}^s_t\mathbb{R}^{x^s_t} + \Bar{w}^n_t\mathbb{R}^{x^n_t})
                                    }
                                )
                            }
                            {
                            log(
                                1 + 
                                e^{
                                    -{I(y_t, \hat{y}_t)}
                                    (\Bar{w}^s_t\mathbb{R}^{x^s_t} + \Bar{w}^n_t\mathbb{R}^{x^n_t})
                                    }
                            )
                            } 
                        I(y_t, \hat{y}_t)
                        \mathbb{R}^{x^n_t} \\
    \end{gathered}
\end{equation}
In the OLVF algorithm, the initialization of all weight parameters occurs at zero, as outlined in algorithm 1 of the OLVF paper. Consequently, the expression $\Bar{w}^s_t\mathbb{R}^{x^s_t} + \Bar{w}^n_t\mathbb{R}^{x^n_t}$ equals 0, which implies that $log(e^{-{I(y_t, \hat{y}_t)}(\Bar{w}^s_t\mathbb{R}^{x^s_t} + \Bar{w}^n_t\mathbb{R}^{x^n_t})}) = 0$. This results in all the weights of the feature classifier remaining at 0 consistently. Therefore, the feature space classifier does not contribute to the OLVF algorithm's predictions.

\paragraph{OCDS} The OCDS algorithm \cite{he2019online} updates weights according to the equation $w_{t+1} = w_t - \nabla_{w_t}\mathcal{F}$, where 
\begin{equation}
    \nabla_{w_t}\mathcal{F} = -2(y_t - w_t^\top \psi(x_t))\psi(x_t) + 
                                \beta_1 \partial||w_t||_1 + 
                                \beta_2 (L + L^\top)w_t.
\end{equation}
Here $\beta_1$ and $\beta_2$ are scale parameters to absorb the magnitudes of $\partial||w_t||_1$ and $(L + L^\top)w_t$, respectively. Similarly, we observe that the first term ($-2(y_t - w_t^\top \psi(x_t))\psi(x_t)$), frequently exceeds acceptable bounds for many datasets. To counteract this, we introduce an absorption scale hyperparameter ($\beta_0$) to moderate the value of the first term.

\subsection{Decision Stumps}
\label{sec:decision_stumps}

Decision stump \cite{iba1992induction} is a one-level decision tree characterized by a single root node leading to two leaves. This root node functions as the feature upon which the binary classification decision is predicated, directing an instance to one of the two classes (leaves) based on a specified threshold (refer to Figure \ref{fig:decisionstump}a). The decision stump's operation is given by
\begin{equation}
    \hat{y} = \text{sign}(x - t).
\end{equation}
Here, $x$ represents the feature value, and $t$ denotes the feature's threshold value. Accordingly, when the feature value equals or surpasses the threshold, the model predicts the positive class; otherwise, it predicts the negative class. Building upon the concept of decision stumps, Schreckenberger et al. proposed two methods: Dynamic Forest (DynFo) and Online Random Feature Forests for Feature Space Variabilities (ORF\textsuperscript{3}V).

DynFo \cite{schreckenberger2022dynamic} constitutes an ensemble approach that integrates multiple weak learners, each modeled on the decision stump, to achieve accurate instance classification. DynFo increases the weights of the well-performing weak learners within the ensemble while decreasing the weights of the underperforming weak learners, which may result in the elimination or retraining of the latter. Each weak learner operates on a subset of the cumulative feature space, termed accepted features, and bases its training on one feature from this subset. The method necessitates the maintenance of a buffer with a predetermined capacity to facilitate the training of newly introduced weak learners as well as the retraining of weak learners who perform poorly. The final output is based on the weighted prediction of all the weak learners. The weight of each learner is adjusted upward or downward depending on whether the learner predicts correctly, incorrectly, or is unable to make a prediction because the feature upon which the split is based is absent from the set of accepted features.

ORF\textsuperscript{3}V \cite{schreckenberger2023online}, similar to DynFo, considers an ensemble of decision stumps to produce a weighted prediction at each time instance. However, unlike DynFo, which trains each decision stump on multiple features, ORF\textsuperscript{3}V constructs each decision stump using only a single feature. Therefore, the arrival of a new feature prompts the creation of a new decision stump exclusively based on that feature. ORF\textsuperscript{3}V implements the Hoeffding bound \cite{hoeffding1994probability} to selectively prune decision stumps according to their statistical results and predictive capability. The updating of the weights for each decision stump in ORF\textsuperscript{3}V is conducted in a manner similar to that of DynFo.

Models based on decision stumps are inherently simple and facilitate straightforward application. Nevertheless, they fail to capitalize on the interrelationships between various features and necessitate a storage buffer for instance retention, which is essential for stump training. The number of weak learners can increase significantly with high-dimensional data. If a limit is imposed on the number of weak learners, then many features may not even participate in the prediction process. 

\underline{DynFo Extension Idea:} DynFo assigns an accepted feature set to each decision stump, which may encompass multiple features. However, for making predictions, it relies on a single feature from the accepted feature set. A diagram adapted from the original DynFo article is presented in Figure \ref{fig:decisionstump}b. A decision tree with one or more than one level can be employed instead of a decision stump (see Figure \ref{fig:decisionstump}c). This could potentially offer enhanced discriminative capability due to its greater complexity relative to the decision stump.

\subsection{Deep Learning}

Deep learning \cite{goodfellow2016deep} encompasses a subset of machine learning algorithms \cite{mahesh2020machine} that operate on the principles of neural networks. The foundational concept of deep learning involves the application of the back-propagation algorithm \cite{lecun1988theoretical} for the purpose of weight training. The prediction of the deep learning model is given by:
\begin{multline}
    \hat{y} = F_n(W_{n}*F_{n-1}(W_{n-1}*F_{n-2}(W_{n-2}* \\
    ...F_{1}(W_1*\mathbf{x} 
    + b_{1})
    + b_{n-2}) + b_{n-1}) + b_n),
\end{multline}
where $F_k$ denotes the feed-forward computation of $k^\text{th}$ layer. Based on the loss incurred, $l = L(y, \hat{y})$, where $L$ represents any selected loss metric, the parameters of the deep learning models are trained via back-propagation as follows:
\begin{equation}
    W = W - \eta\frac{\partial{l}}{\partial{W}},
\end{equation}
where $\eta$ is the learning rate. Deep learning models have been effectively employed in online learning scenarios, demonstrating good results \cite{sahoo2017online}. However, the architectures of deep learning models are notably inflexible, making their adaptation to haphazard inputs a challenge. Building upon the excellent predicting power of deep learning models, Agarwal et al. proposed an auxiliary network (Aux-Net) model and an auxiliary dropout (Aux-Drop) concept to address haphazard inputs.

Aux-Net \cite{agarwal2023auxiliary} operates under the premise that there exists a consistently available input feature, referred to as the base feature. The architecture of Aux-Net comprises four distinct modules: a base module dedicated to processing the base features, an auxiliary module designed to handle haphazard inputs, a middle module that integrates the information from both auxiliary and base modules, and an end module that performs further processing of the amalgamated information. The auxiliary module is characterized by its scalability and flexibility, enabling it to adapt its architecture in response to the variable nature of the haphazard inputs it receives.

Aux-Drop \cite{agarwal2023auxdrop} introduces a methodology that allows any online deep-learning algorithm to process haphazard inputs effectively.  Similar to Aux-Net, Aux-Drop also assumes the presence of a base feature. The concept involves applying dropout to a specific hidden layer (termed AuxLayer), which is partitioned into auxiliary and non-auxiliary nodes. There is a one-to-one connection between haphazard inputs and auxiliary nodes and in the absence of a haphazard input, the corresponding auxiliary node is selectively dropped. To meet the requirement of the number of dropped nodes, additional nodes are randomly selected for dropout in conjunction with the selectively dropped nodes.  Furthermore, as new features emerge, Aux-Drop incorporates new nodes into the set of auxiliary nodes. By leveraging dropout, Aux-Drop adeptly manages haphazard inputs within deep learning frameworks.

Aux-Net and Aux-Drop harness the predictive prowess of deep learning algorithms, rendering them particularly well-suited for large and complex datasets. Although Aux-Net and Aux-Drop show potential, the reliance on the assumption of base features constitutes a limitation. We discuss the possibilities of overcoming this assumption in Section \ref{sec:model_preparation}.

\section{Experiments}
\label{sec:experiments}

\subsection{Dataset Preparation}
\label{sec:dataset_preparation}

\input{Tables/Tab_Hyperparameters}

All the real datasets exhibit the characteristics of haphazard inputs thus they are used in their original form. However, this is not the case with synthetic datasets. To create haphazard input from synthetic datasets, we simulate the availability of each feature independently, employing a uniform distribution with a probability $p$. For example, if $p = 0.75$, $(1-p)*100 = 25\%$ of the values of each feature are simulated as unavailable independently of other features, following a uniform distribution. We consider three values of $p$: 0.25, 0.5, and 0.75. It is important to note that certain synthetic datasets (WPBC, australian, and WBC) already contain some missing values and are considered missing in addition to the aforementioned simulation (see Table \ref{tab:datasetdescription}).

\subsection{Model Preparation for no assumption}
\label{sec:model_preparation}
We observe that certain methods make assumptions like pre-training, storage buffer, and the availability of base features (see Table \ref{tab:modelcomparison}). We consider datasets to be entirely haphazard, necessitating models to adapt and manage haphazard inputs devoid of any preliminary assumptions. Consequently, we tailor the model or adjust the model's hyperparameters to ensure that the influence of such assumptions is minimized.

\paragraph{Pre-training} Online learning mandates that models start predictions from the outset. However, NB3 and FAE necessitate pre-training. To align with the principles of online learning, we pre-train these models using a single instance. 

\paragraph{Storage Buffer} OVFM, DynFo, and ORF3V incorporate storage buffers within their training process. These buffers store data instances to facilitate the computation of statistics necessary for model training. A quintessential feature of online learning is the model's exposure to each instance only once during training which is violated in these models. Despite this, the storage buffer remains an integral component of OVFM, DynFo, and ORF3V. To accommodate this with the concept of online learning, we maintain the storage buffer but limit its capacity to a modest size of 20 instances. 

\input{Tables/Tab_Results_HyperParameterSearch_2}

\paragraph{Base Features} Aux-Net and Aux-Drop operate on the premise that at least one base feature is present. To address this assumption, the following adaptations can be implemented:
\begin{itemize}
    \item Imputation: By designating one of the initial features received at the first time instance as the base feature, it can be imputed in subsequent steps when it is absent. Imputation strategies can be the use of a global mean, forward-filling, or more sophisticated methods. This adaption works for both Aux-Net and Aux-Drop. The potential drawback is the introduction of noise, however, it would affect only one feature. 
    \item Architectural change: Specifically for Aux-Drop, the initial layer can be reconfigured as an AuxLayer, thus allowing every feature to be haphazard input and effectively mitigating the base feature assumption.
    \item Dummy feature: The creation of a dummy feature as the base feature enables all actual features to be processed as haphazard inputs.
\end{itemize}
For Aux-Drop, we adopt the architectural change adaptation, while for Aux-Net, we implement forward-filling imputation for two base features.

\subsection{Implementation Details}
In adherence to the requirements of online learning tasks, we preserve the order of the data instances. We execute all non-deterministic models five times, reporting the mean $\pm$ standard deviation of their results. In contrast, we run the deterministic models --- NB3, FAE, and OLVF --- just once. The evaluation and comparison of models are based on all five metrics outlined in Section \ref{sec:metrics}, and we also document the time each model requires for execution.
We developed the majority of the models (refer to Table \ref{tab:modelcomparison}) using the PyTorch framework \footnote{\url{https://github.com/Rohit102497/HaphazardInputsReview}}. We conducted all experiments on an NVIDIA DGX A100 machine, exclusively employing CPUs to accommodate the serialized processing characteristic of online learning. Within Aux-Drop, we adopt the online deep-learning architecture specified by Sahoo et al. \cite{sahoo2017online} as the foundational model. Next, we discuss the hyperparameters of each model.

\input{Tables/Tables_Result_Small_Bacc_Time_Bold}
\input{Tables/Tables_Result_Medium_Large_Bacc_Time}

\paragraph{Hyperparameters} Table \ref{tab:hyperparameter_description} details the definitions of each model's hyperparameters. We conducted a hyperparameter search for all applicable models, with the search values also presented in Table \ref{tab:hyperparameter_description}. We established the hyperparameter values and their search ranges based on the guidelines provided in each model's originating paper. Due to the substantial computational demands of certain models, we fixed some hyperparameter values in accordance with the original papers rather than conducting a search. The hyperparameter optimization was executed at a feature availability probability ($p$) of 0.5 for all synthetic datasets, and the optimized parameters were then applied uniformly to $p$ = 0.25 and $p$ = 0.75. Given the prevalence of imbalanced datasets, we selected the optimal hyperparameters based on balanced accuracy. Table \ref{tab:hyperparameters_noassumption} lists the best hyperparameters identified for each model at $p$ = 0.5. Table \ref{tab:hyperparameters_noassumption} includes only those hyperparameters subjected to the search process, while the fixed hyperparameter values are available in Table \ref{tab:hyperparameter_description}.

\subsection{Results}
\label{sec:results}

In the main manuscript of our survey, we compare all the models based on balanced accuracy and provide comprehensive results across other metrics in Supplementary Section I. The rationale for selecting balanced accuracy is outlined in Q.1. of Section \ref{sec:observations}. We further include execution times to assess the computational complexity of each model. Table \ref{tab:results_small} displays the results of all models on small datasets, encompassing both real and synthetic datasets with $p$ = 0.25, 0.5, and 0.75. The outcomes for medium and large datasets are detailed in Table \ref{tab:results_medium_large}.

Given the volume of these results, direct comparison of all models proves challenging. Hence, we introduce 5 additional metrics for a comprehensive evaluation: performance, prediction consistency, data scalability, speed, and feature scalability, as detailed in Table \ref{tab:results_inference}. We assess efficacy across different dataset groups: small, medium, and large. Furthermore, we average the metric value across these dataset categories to facilitate a single comparison score. Table \ref{tab:results_inference} provides this quantitative comparison. Additionally, a qualitative analysis is conducted on the aforementioned metrics, with models rated on a scale from one to five stars ($*$ to $*****$), indicating the worst to the best model respectively, as shown in Table \ref{tab:modelcomparison}.

\paragraph{Performance} It is defined as the average of the mean balanced accuracy of each model on the specified group of datasets. DynFo, OVFM, and Aux-Drop deliver the best performance (perf) for the `Small', `Medium', and `Large' dataset groups, respectively. OVFM leads overall with a balanced accuracy of 63.1, followed by Aux-Drop at 59.49. We distribute the models into five performance bins with a step size of (best-worst)/5 = (63.1-50.7)/5 = 2.48. The performance categorization is as follows: models fall into \{$*$; $*$ $*$; $***$; $***$ $*$; $*****$\} if their perf\textsubscript{Average} $\in$ \{[50.7, 53.18); [53.18, 55.66); [55.66, 58.14); [58.14, 60.62); [60.62, 63.10]\}. 

\paragraph{Data Scalability} It refers to the percentage improvement in performance as dataset size increases from small to medium to large. We calculate this as a percentage change in performance between consecutive dataset groups: ((perf\textsubscript{group2} - perf\textsubscript{group1})/perf\textsubscript{group1})*100, where (group1, group2) can be (small, medium) or (medium, large). A model exhibiting good data scalability should show a substantial performance increase, indicating that the data scalability measure is directly proportional to performance gains. However, as the volume of data increases, the model should incur minimal performance degradation. To account for this, the data scalability metric is inversely proportional to the square of absolute performance decrease. Given the squaring of performance losses, the coordinate system's origin is shifted to accurately reflect values within the interval (0, 1).  Thus, we mathematically define data scalability measure as $\sum_{p \in Inc}{(1 + p)}/\sum_{n \in Dec}{(1 + {|n|}^2)}$ where $Inc$ and $Dec$ denotes set of increased and decreased performance, respectively, and $|\cdot|$ represents the absolute value operator. Aux-Drop is highly data scalable compared to other models. Categorization for data scalability uses a step size of 0.076 and is segmented as \{$*$; $*$ $*$; $***$; $***$ $*$; $*****$\} for Data Scalability\textsubscript{Measure} $\in$ \{[0.0, 0.08); [0.08, 0.15); [0.15, 0.23); [0.23, 0.3); [0.3, 0.38]\}.

\paragraph{Prediction Consistency} It is the average of the standard deviation of balanced accuracy for each model across specified dataset groups. Models with lower standard deviations yield more consistent results. Prediction consistency is not applicable to deterministic models such as NB3, FAE, and OLVF. DynFo exhibits consistency in `Small' and `Large' datasets, while OVFM and Aux-Net demonstrate high consistency in `Medium' datasets. Overall, DynFo gives the highest consistent prediction. For categorization, we employ a step size of 0.22, assigning \{$*****$; $***$ $*$; $***$; $*$ $*$; $*$\} to Prediction Consistency\textsubscript{Average} $\in$ \{[0.21, 0.43); [0.43, 0.65); [0.65, 0.87); [0.87, 1.09); [1.09, 1.31]\}.

\paragraph{Speed} is defined as the inverse of the average of the mean time taken by each model on the specified group of datasets. As evident from Table \ref{tab:results_inference}, OLVF is the fastest on `Small' and `Medium' datasets and NB3 is fastest on the `Large' datasets.  Overall, OLVF is the fastest model, whereas DynFo is the slowest model. The categorization is done as follows: \{$*****$; $***$ $*$; $***$; $*$ $*$; $*$\} if $log$(Time\textsubscript{Average}) $\in$ \{(1, 2]; (2, 3]; (3, 4]; (4, 5]; $\ge$ 5\}.

\paragraph{Feature Scalability} Feature Scalability refers to the model's ability to handle an increasing number of features for a fixed number of instances with minimal impact on running time. The SUSY and HIGGS datasets serve as a basis for comparison, sharing the same number of instances but differing in feature count; SUSY has 8 features, while HIGGS has 21, making HIGGS 2.625 times higher in feature size. For a consistent comparison, we examine the $p = 0.5$ experiment where a hyperparameter search is conducted. We consider the best hyperparameters found for the SUSY dataset and consider them for HIGGS data wherever possible for fair comparison. Aux-Drop, with a ratio of 1, demonstrates high feature scalability by requiring similar running times for both datasets. Categorization of feature scalability uses a step size of 0.84, assigning \{$*****$; $***$ $*$; $***$; $*$ $*$; $*$\} to Feature Scalability\textsubscript{Ratio} $\in$ \{[1, 1.84); [1.84, 2.68); [2.68, 3.52); [3.52, 4.36); [4.36, 5.20]\}.

\input{Tables/Tab_Result_Inference}

\section{Observations}
\label{sec:observations}

In this section, we provide observations derived from the results produced by all models, as outlined in Section \ref{sec:results}.

\textbf{Q.1. Are current metrics sufficient in the field of haphazard inputs?} The haphazard input field employs the number of errors and accuracy for model comparison and both the metrics are interrelated. The accuracy of NB3 on the WPBC dataset with $p$ = 0.25 is 73.23, which surpasses the accuracy of OLVF (70.71). However, the balanced accuracy of NB3 is 49.48, which falls significantly short of OLVF's balanced accuracy of 58.08 (see Table \ref{tab:results_small}). With an imbalance ratio of 23.74\% for WPBC (refer to Table \ref{tab:datasetdescription}), NB3 may have correctly predicted more instances, but potentially only from the majority class, thus inflating its accuracy relative to OLVF. OLVF, however, emerges as a superior classifier, having more effectively identified instances from both classes. This discrepancy indicates that existing metrics are insufficient for the haphazard inputs domain. Consequently, we introduce three additional metrics--AUROC, AUPRC, and balanced accuracy--to enable more accurate model comparisons.

\textbf{Q.2. Do complex models help in addressing haphazard inputs?} The top two models in terms of performance are OVFM and Aux-Drop, whereas, the bottom two models are ORF\textsuperscript{3}V and NB3 (see Table \ref{tab:results_inference}). This observation suggests that complexity in models like OVFM and Aux-Drop does indeed contribute to effectively addressing haphazard inputs. Additionally, it is noted that execution time does not necessarily correlate with model complexity. The simple decision stump-based ORF\textsuperscript{3}V model requires a total execution time of 10073.08 seconds, which is substantially greater than the 4164.42 seconds required by the more complex deep learning-based Aux-Drop model, as detailed in Table \ref{tab:results_inference}.

\textbf{Q.3. What is the environmental impact of the field of haphazard inputs and how to reduce it?}
Given the issue of climate change, evaluating the environmental impact of research activities is essential. In our study, we measure the carbon footprint of each model during benchmarking and hyperparameter optimization on our DGX A100 machine, utilizing the green algorithms methodology\footnote{\url{https://calculator.green-algorithms.org/}} \cite{lannelongue2021green}. The parameters for these calculations are as follows: core type - CPU, number of cores - 128, model - Any (Dual AMD Rome 7742 not supported), available memory - 1 TB, platform - Local Server, and location - Europe (Norway). The carbon footprint and energy consumption for each model, based on its execution time, are recorded in Table \ref{tab:Carbon_Footporint}. It is important to note that these values are not for model comparison, as they depend on the number of hyperparameter searches; rather, they serve to illustrate the carbon footprint of our research. Our research has resulted in a total carbon emission of 84.26 kg, corresponding to an energy consumption of 11058.3 kWh. \\
\underline{Motivation:} Our motivation for reporting the carbon footprint is twofold. Firstly, we aim to support the United Nations' Sustainable Development Goals (SDGs) by assessing the environmental impact of our research to promote sustainable practices. Secondly, we have recognized that previous research in the haphazard input field is often not reproducible due to unavailable code, leading to redundant benchmarking efforts. By providing open-source reproducible code and documenting the carbon footprint of each model, we hope to encourage the research community to build upon our benchmarking results. \\
\underline{Model Comparison:} Furthermore, we offer a concise comparison of the carbon footprint for each model. We select two datasets: SUSY ($p$ = 0.5) and imdb, chosen for their large instance count and feature size, respectively, representing two critical dimensions. The cumulative carbon footprint for a single run of both datasets for each model is shown in Figure \ref{fig:carbon_emission}. OLVF demonstrates the lowest carbon emissions, followed by NB3, while DynFo exhibits the highest carbon emissions.

\input{Tables/Tab_Carbon_Footprint}
\input{FigureTex/Fig_CarbonEmission}
\input{FigureTex/Fig_RealSynthetic}

\textbf{Q.4. How does the model fare in real and synthetic datasets?} In Figure \ref{fig:real_vs_synthetic}, we illustrate the mean performance of each model on both real and synthetic datasets by aggregating the mean balanced accuracy for each real dataset. For synthetic datasets, we perform a similar aggregation across all $p$ values. OLVF and OVFM are the best-performing models in real and synthetic datasets, respectively. We observe that simpler methods like naive Bayes, yield strong performance in real datasets, whereas complex models like linear classifiers and deep learning, are more effective in synthetic datasets. The diminished performance of complex models on real datasets may be attributed to the curse of dimensionality \cite{verleysen2005curse}, as these datasets often feature high dimensionality.

\textbf{Q.5. Why do all the models perform poorly in diabetes\_us dataset?} This dataset presents a significant challenge, with all models yielding a balanced accuracy near 50\%. Notably, except for linear classifiers, models predominantly predict the majority class. Linear classifiers, particularly the OVFM, exhibit attempts to recognize both classes, as reflected in their error rates (see Supplementary) and balanced accuracy. The difficulty with the diabetes\_us dataset may stem from its abundance of categorical features, which could explain the relative success of the OVFM model 
in classifying this dataset.

\textbf{Q.6. Does execution time increase with increasing $p$ values?}  As models process an increasing volume of data with the rise of the $p$ value from 0.25 to 0.75, one would intuitively expect a corresponding increase in the execution time required by each model. This trend is evident in NB3, DynFo, ORF\textsuperscript{3}V (except in large datasets for DynFo and ORF\textsuperscript{3}V), and Aux-Net (see Table \ref{tab:results_small} and \ref{tab:results_medium_large}). Conversely, FAE, OLVF, OCDS, OVFM, and Aux-Drop exhibit no specific pattern in execution time across varying $p$ values. This can be attributed to the nature of these models. Specifically, FAE's dynamic adjustment of the number of classifiers means its computational demand is contingent on the data characteristics rather than volume, resulting in the absence of specific execution time patterns. Similarly, the feature reconstruction processes in OCDS and OVFM are data-dependent, therefore, the volume of data does not directly influence the execution time. OLVF and Aux-Drop have a fixed amount of computation, leading to comparable execution time across different $p$ values.



\section{Related Developments}

In this section, we explore the field of haphazard inputs to find architectural ideas for modeling such inputs, delineate the subfields within this domain, and identify possible applications related to the field of haphazard inputs.

\subsection{Ideas on adapting ISTS-based model to handle Haphazard Inputs}
\label{sec:adaptablemodels}

\input{FigureTex/Fig_AdaptedSeFT}

Irregularly Sampled Time Series (ISTS) handles time series data where the sampling rate of each feature is different \cite{weerakody2021review, 8013802}. Thus, at a specific time instance, only a few features may be available. The ISTS data can also be seen as haphazard inputs, however, it differs fundamentally as it pertains to an offline learning context \cite{ben1997online}, devoid of characteristics like obsolete features, sudden features, and an unknown number of total features. Nevertheless, certain models within the ISTS domain can be adapted to handle haphazard inputs. Below we list those models and their possible adaptations.

\paragraph{Set Functions for Time Series (SeFT)} SeFT \cite{horn2020set} models ISTS data by transforming each feature value into a set comprising the measured value, a feature indicator, and the time difference from the last observation of the identical feature. SeFT employs a single deep neural network to generate a k-dimension vector for each set. These vectors are aggregated via an operator before being fed into a final neural network to get a prediction. The diagram of SeFT is shown in Figure \ref{fig:adaptedSeFT}. SeFT may be used for haphazard inputs by considering each feature value as a set. All the sets at each time instance are passed to a neural network to obtain a k-dimensional vector, then aggregated and finally used to predict an output from the final neural network. This possible adaptation of the SeFT method is depicted in Figure \ref{fig:adaptedSeFT}.

\paragraph{Switch LSTM Aggregate Network (SLAN)} SLAN \cite{agarwal2023modelling},  leverages multiple LSTM blocks \cite{8840975} to process ISTS data, with each feature assigned to a dedicated LSTM block. A switch layer enables SLAN to adapt its structure in response to unavailable features, activating LSTM blocks corresponding to observed features. Subsequently, the long-term hidden states from the active LSTM blocks are aggregated after each time instance and shared as a global long-term hidden state for the next time instance. Finally, the decayed short-term hidden states of all LSTM along with the global long-term hidden state are concatenated to produce a final output. SLAN can be adapted to handle haphazard inputs by allowing each LSTM block to process one feature and aggregating both short-term and long-term states. The aggregated short-term and long-term states can be further passed through a neural network to produce a final output. However, this adaptation of SLAN may result in a significantly increased model size.

\subsection{Subfields of Haphazard Inputs}
\label{sec:subfield}

\input{FigureTex/Fig_SubfieldHaphazard}

In this survey, we identified four fields that handle streaming data with varying input dimensions. However, these fields assume some form of structure in their data and hence can be considered a subfield of haphazard inputs. These fields are trapezoidal data streams \cite{OLSFtrapezoidal}, feature evolvable streams \cite{hou2017learning}, unpredictable feature evolution \cite{hou2021prediction}, and incremental and decremental features \cite{hou2017one}. Figure \ref{fig:subfieldhaphazard} depicts the evolution of the feature space across these four fields. Next, we briefly discuss these fields and their relation to haphazard inputs.

\paragraph{Trapezoidal data streams} It deals with streaming data where the feature space expands progressively over time \cite{OLSFtrapezoidal, liu2022online}. Here, the features received at time $t$ are guaranteed to be present in subsequent time instances, with the possibility of receiving additional new features. Mathematically, this is represented as $d_t <= d_{t+1}$, where $d_t$ denotes the feature count at time $t$. Unlike haphazard inputs, trapezoidal data streams do not accommodate the missing data, missing features, and obsolete features.
    
\paragraph{Feature evolvable streams} It addresses streaming data where old features vanish and new features occur with an assumption of an overlap period between this transition \cite{hou2017learning, zhang2020learning}. It also assumes that the data comes in batches. Therefore, if $d_{T_1}$ and $d_{T_2}$ represent the feature space during time period $T_1$ and $T_2$ and $B$ is the overlap period, then the feature space evolves as $d_{T_1}$, $d_{T_1}$ + $d_{T_2}$ and $d_{T_2}$ during time intervals: [1, $T_1$], [$T_1+1$, $T_1+B$] and [$T_1+B+1$, $T_1+B+T_2$], respectively. Feature evolvable streams work on the assumption of overlap period and batch and hence cannot handle any characteristics of haphazard inputs.
    
\paragraph{Unpredictable feature evolution} It extends the scope of feature evolvable streams by relaxing the assumption on overlapping periods \cite{hou2021prediction}. The $d_{T_1}$ feature space in the time period $T_1+1$ to $T_1+B$ is allowed to contain missing values. Similar to feature evolvable streams, it works on the assumption of overlap and batch. Thus, it fails to accommodate haphazard inputs and their characteristics.
    
\paragraph{Incremental and decremental features} This domain contemplates streaming data where certain features vanish, others survive, and new ones are augmented \cite{hou2017one, dong2021evolving}. Data arrives in batches, with the initial batch encompassing both vanishing and surviving features and subsequent batches comprising surviving and newly augmented features. This pattern persists across the whole data. Although the incremental and decremental nature of features bears resemblance to haphazard inputs, the presence of batch fundamentally limits the ability to address the distinctive characteristics of haphazard inputs.

\subsection{Applications and their molding into haphazard inputs}
\label{sec:applications}

The biggest challenge in the field of haphazard inputs lies in identifying and securing datasets from real-life applications. Here, we provide two applications, anticipating that the research community will curate appropriate datasets to advance the study of haphazard inputs.

\paragraph{Sub-cellular organism}
\label{sec:mitochondria}
Mitochondria, sub-cellular structures inside living cells, play a crucial role in cell metabolism, and energy production and delivery at different functional zones of the cells. Hence, modeling mitochondria and other organelles may inform the behavior of cells, thereby aiding new drug discoveries \cite{paul2021artificial}. During the drug discovery process, mitochondria morphology undergoes some changes, with the number of fusion, fission, and kiss-and-run events \cite{sekh2020simulation} increasing or decreasing depending on the administered drugs. Quantifying these events is invaluable to biologists, but discerning them via the naked eye is impractical because there are hundreds of mitochondria in a cell. By integrating haphazard inputs with a segmentation model \cite{minaee2021image}, these mitochondrial dynamics can be effectively modeled. The phenomena of mitochondrial fusion and fission introduce obsolete and sudden features, while kiss-and-run events can lead to sudden, obsolete, and missing features. Collaborative, interdisciplinary research with life scientists holds promise for applying haphazard input models to address this challenge.

\paragraph{Autonomous car} Technological advancements and the advent of sophisticated cars have led to the development of autonomous cars that can sense their environment, navigate, and make real-time decisions \cite{8457076}. The software enabling these cars integrates multiple specialized models for distinct tasks. One may envision that a perfect self-driving car would require an end-to-end pipeline that can correctly process data from the car sensors, encapsulating both the car's status and its operational environment. However, this would create haphazard inputs. Therefore, the creation of a dedicated haphazard dataset for autonomous cars could facilitate the development of such an end-to-end pipeline, potentially yielding more reliable and safer autonomous cars \cite{guo2019safe}.

\section{Future Research Direction}
\label{sec:future_directions}

Future research must concentrate on developing sophisticated models and curating comprehensive real datasets pertinent to haphazard inputs. The subsequent section substantiates this assertion from multiple perspectives.

\paragraph{Datasets}
\label{sec:data_pov}
This survey examines 20 datasets, encompassing both real and synthetic datasets. Despite this, the current collection of datasets is insufficient. We identified only 3 medium and 3 large datasets within our survey, with just 1 real dataset in each size category, which is inadequate to meet research demands. Consequently, there is a pressing need for specialized medium and large datasets that address significant applications, which are detailed in Section \ref{sec:future_directions}(b). Notably, all 20 datasets we reviewed are tabular. Therefore, it would also be of interest to introduce image and video datasets.  The study of mitochondria, as discussed in Section \ref{sec:applications}(a), might be a potential application for image and video datasets.

\paragraph{Applications}
\label{sec:application_pov} 
Current applications of haphazard inputs include government restriction analysis, spam email classification, movie sentiment classification, and patient readmission identification. These applications not only serve as a direction but also establish a foundation for future applications. Section \ref{sec:applications} introduces two prospective application directions for the research community, focusing on sub-cellular organisms and autonomous cars. Furthermore, the concept of haphazard inputs could be extended to model applications in aircraft health modeling \cite{jigajinni2021health}, customer profiling in e-commerce \cite{wiedmann2002customer}, autonomous ship modeling \cite{9989588}, and smart city monitoring \cite{luckey2021artificial}.

\paragraph{Architectures} 
The field of haphazard inputs is in a nascent stage, with merely nine models equipped to manage such data. Certainly, these models are not sufficient, necessitating the development of more sophisticated models for accurate predictions. Table \ref{tab:results_inference} underscores the need for advanced models, like the deep learning-based Aux-Drop for large datasets and OVFM for medium datasets, to enhance performance. This necessity is further corroborated by Tables \ref{tab:results_small} and \ref{tab:results_medium_large}, which show sub-optimal model performance across numerous datasets, indicating significant potential for improvement. This survey presents novel architectural ideas for decision stumps and deep learning in Sections \ref{sec:decision_stumps} and \ref{sec:adaptablemodels}, respectively. Moreover, the recent progress in Large Language Models (LLMs), such as TabLLM \cite{hegselmann2023tabllm}, which can perform few-shot classification of tabular data, offers a promising direction for haphazard inputs. Given the discussion on the need for image and video datasets in Section \ref{sec:future_directions}(a), consequently, there would also be a need for computer vision models \cite{prince2012computer}. Additionally, the application of graph neural networks \cite{ben2022graph} to model haphazard inputs presents multiple new avenues for exploration in this field.

\paragraph{Online Learning}

The main characteristic of haphazard inputs is the nature of its data, \emph{i.e.}, streaming data. Consequently, online learning becomes an integral part of haphazard input models. Therefore, the challenges associated with online learning --- like concept drift \cite{lu2018learning}, stability-plasticity dilemma \cite{9349197}, imbalanced data \cite{8443399}, and normalization \cite{ross2013normalized} --- are also inherited by haphazard inputs. These challenges remain unexplored in existing literature. Future research must address these issues to develop improved and efficient models for haphazard inputs. These specific challenges are outlined here:

\begin{itemize}
    \item \textit{Concept Drift}: The streaming data is usually non-stationary \cite{7489383}, \emph{i.e.}, the distribution and statistical properties of the data change with time. Thus, it induces a concept drift and leads to poor predictions. Therefore, the model needs to adapt to these temporal changes.
    \item \textit{Stability-Plasticity Dilemma}: The model needs to learn new knowledge (plasticity) without forgetting the previously learned knowledge (stability). However, a trade-off is often observed, leading to a dilemma, where enhancing plasticity may compromise stability and vice-versa \cite{9892055}.
    \item \textit{Imbalanced data}: Imbalanced datasets, characterized by disproportionate class distributions, may bias models to favor the majority class in predictions. Thus, it is necessary to devise solutions to mitigate imbalanced data problems. Study shows that techniques like weighted learning can mitigate the effects of data imbalance \cite{8443399}.
    \item \textit{Normalization}: Normalization refers to transforming data to a preferred coordinate system such that each data value belongs to a consistent range. However, in online learning, the data arrives sequentially, therefore, the transformation operation is difficult and unclear. Research must explore and innovate upon normalization techniques, drawing from works like \cite{ross2013normalized} and \cite{gupta2019adaptive}, to suit the needs of haphazard inputs.
\end{itemize}

\section{Conclusion}
\label{sec:conclusion}

We have witnessed recent growth in online learning under haphazard input conditions, yet the field remains nascent with noticeable gaps. Recognizing these gaps, we offer a comprehensive review of currently available models. Our taxonomy divides them into four types: naive Bayes, linear classifiers, decision stumps, and deep learning. We also classify datasets into small, medium, and large categories. We underscore the limitation of current comparison metrics and introduce three additional metrics --- AUROC, AUPRC, and balanced accuracy --- accompanied by five novel metrics derived from balanced accuracy and execution time. Our analysis confirms the value of complex models in managing haphazard inputs. We introduce an assessment of the environmental impact of our research, promoting sustainable practices in line with the UN's SDGs to reduce the carbon footprint of computational research. This reflection not only pushes the boundaries of scientific and technological advancements. It also champions a sustainable research ethos that respects environmental limitations. We provide open-source reproducible code for each model along with their carbon footprints, encouraging the research community to build upon our benchmarking results. In our efforts to advance the field of the haphazard inputs field, we outline future research directions, encompassing potential applications and corresponding datasets, prospective model architectures, and online learning challenges.

\section*{Acknowledgments}
This work was supported by the Research Council of Norway Project (nanoAI, Project ID: 325741), H2020 Project (OrganVision, Project ID: 964800), HORIZON-ERC-Stg Project(3D-nanoMorph, Project ID:894233) and VirtualStain (UiT, Cristin Project ID: 2061348)

\bibliographystyle{ieeetr}
\bibliography{main}  


\section{Biography Section}

\vspace{-33pt}
\begin{IEEEbiography}[{\includegraphics[width=1in,height=1.25in,clip,keepaspectratio]{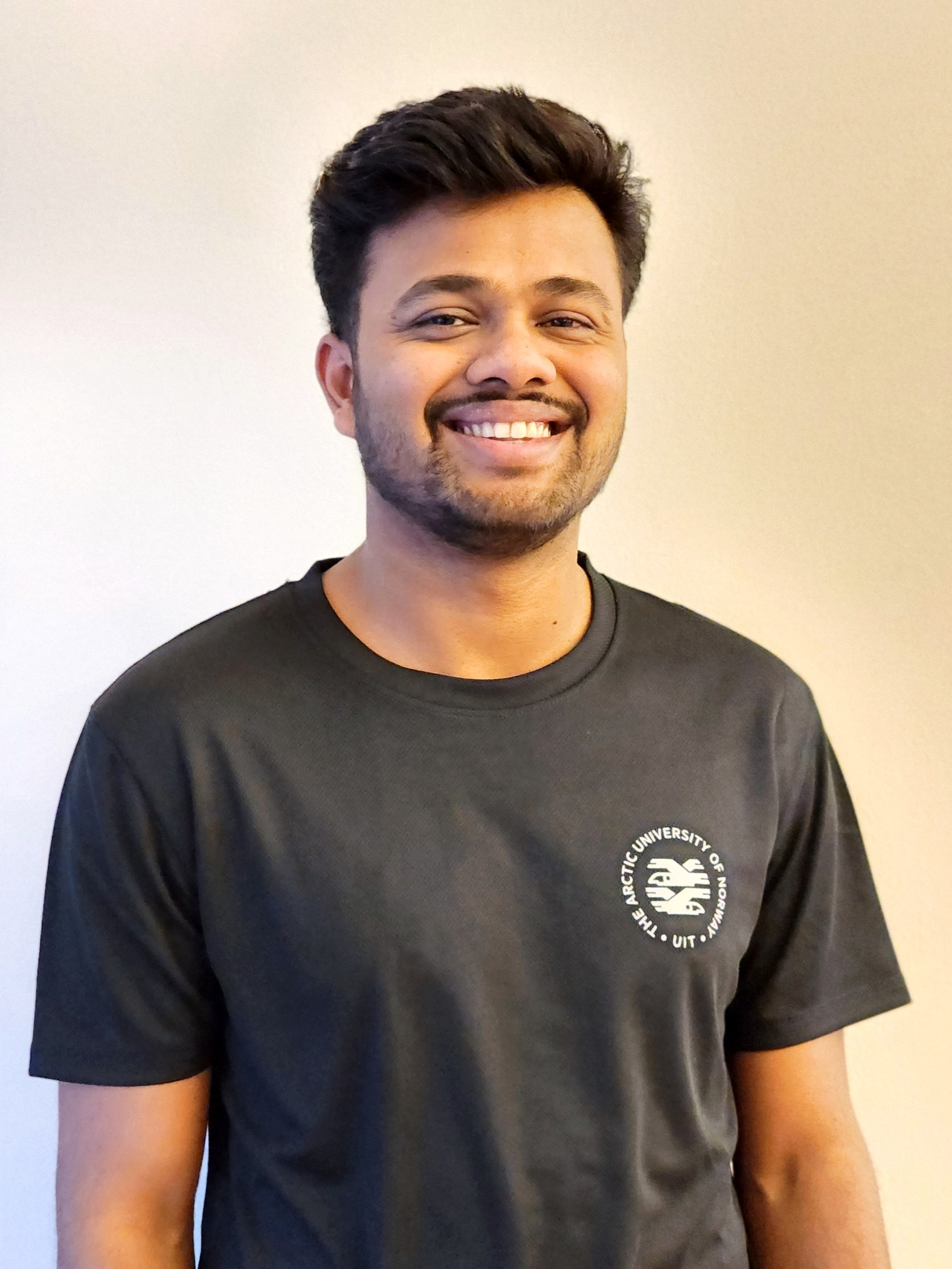}}]{Rohit Agarwal} is currently pursuing his Ph.D. degree from UiT The Arctic University of Norway. His research focuses on haphazard inputs and irregularly sampled time series. Prior to this, he was a Cloud Engineer at Adobe Inc. from 2020 to 2021. He received the gold medal in his Integrated Master's degree from the Indian Institute of Technology (Indian School of Mines), Dhanbad, India in the year 2020 for his excellent academic performance.

\end{IEEEbiography}


\vspace{-33pt}
\begin{IEEEbiography}[{\includegraphics[width=1in,height=1.25in,clip,keepaspectratio]{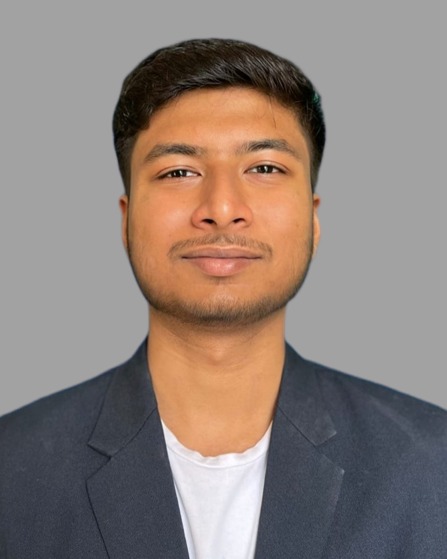}}]{Arijit Das} is pursuing his Bachelors in Electronics and Communication Engineering along with Masters in Computer Science and Engineering from Indian Institute of Technology (Indian School of Mines), Dhanbad, India. He is expected to graduate in the year 2026. His research interest lies in the field of Deep Learning and Haphazard Inputs for Neural Networks. He has participated in Inter IIT Tech meets and has also received several accolades in various data science and machine learning hackathons.

\end{IEEEbiography}

\vspace{-33pt}
\begin{IEEEbiography}[{\includegraphics[width=1in,height=1.25in,clip,keepaspectratio]{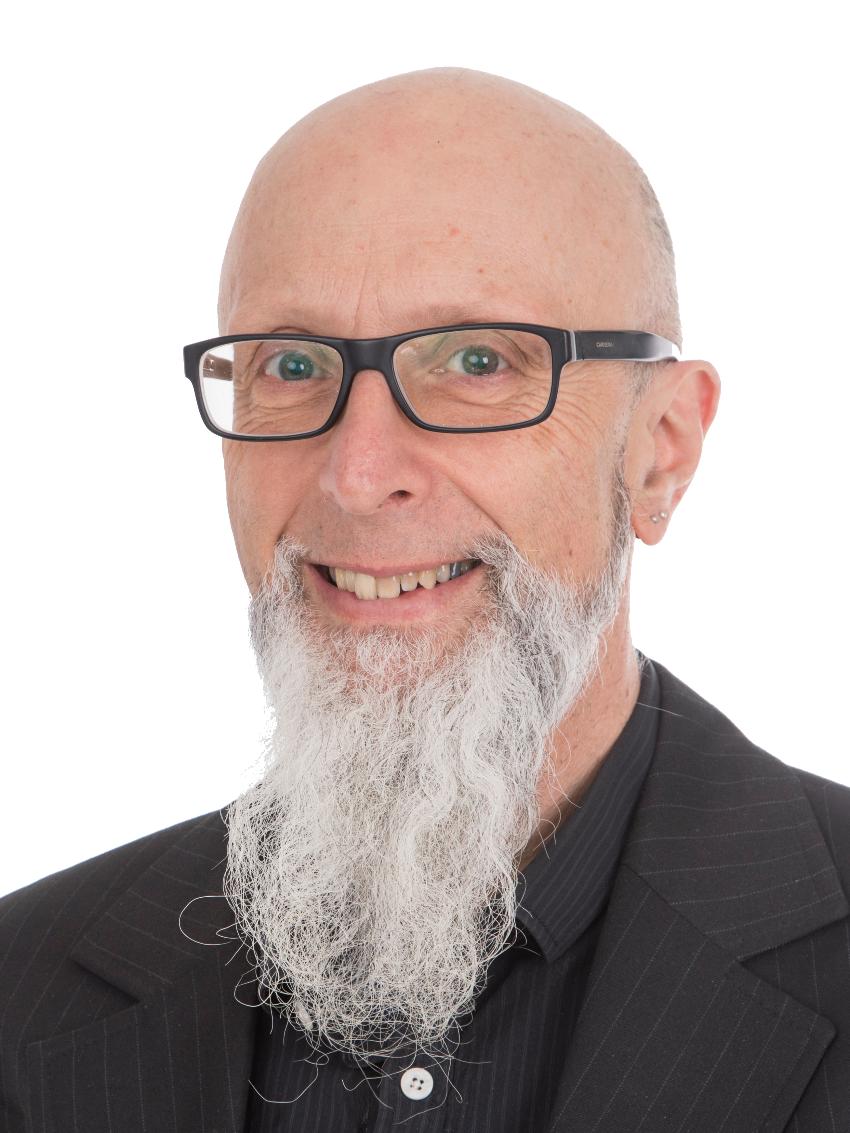}}]{Alexander Horsch} received a Ph.D. in computer science in 1989 from the Technical University Munich (TUM), Munich, Germany. He is a professor at UiT The Arctic University of Norway. His main scientific contributions are in the creation of systems for biomedical data analytics. Early work focused on computer-aided diagnosis of cancer. Later he worked in sensor-based physical activity research on large cohorts. Currently, his main research focus is on explainable deep learning with use cases from medicine, biology, and public health.
\end{IEEEbiography}

\vspace{-33pt}
\begin{IEEEbiography}[{\includegraphics[width=1in,height=1.25in,clip,keepaspectratio]{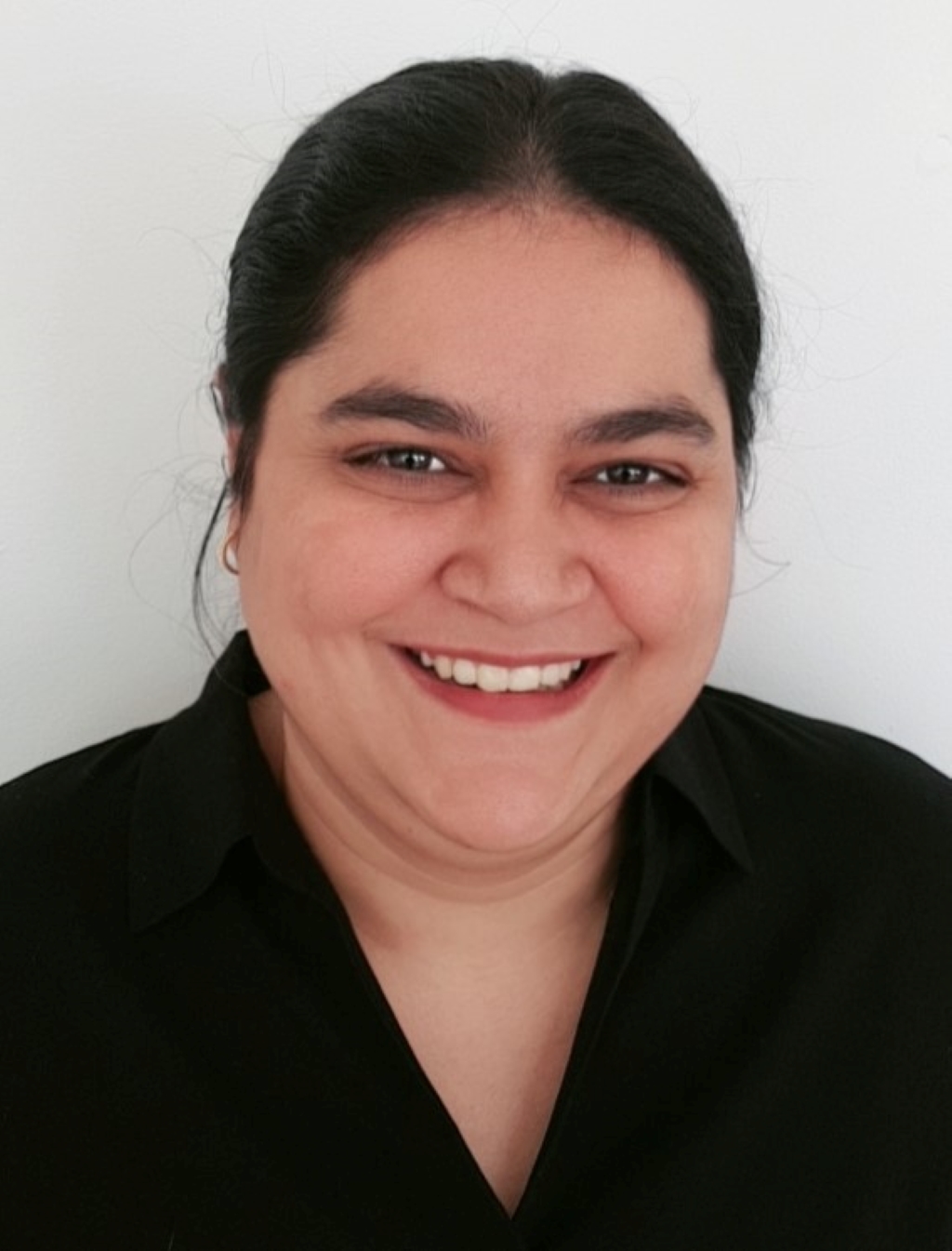}}]{Krishna Agarwal} (OSA Senior Member, URSI Senior Member) received her Ph.D. from the National University of Singapore in the year 2011 and her B.Tech. degree from the Indian Institute of Technology (Indian School of Mines), Dhanbad, India, in 2003. She has received the "Distinguished Alumnus Award" from the Indian Institute of Technology (Indian School of Mines), Dhanbad, India for the year 2020. She was awarded the Marie Skłodowska-Curie Actions Individual Fellowship for the year 2017-2019 and is currently a Professor at UiT The Arctic University of Norway. Her current research interests are computational nanoscopy, super-resolution imaging, and inverse problems.

\end{IEEEbiography}

\vspace{-33pt}
\begin{IEEEbiography}[{\includegraphics[width=1in,height=1.25in,clip,keepaspectratio]{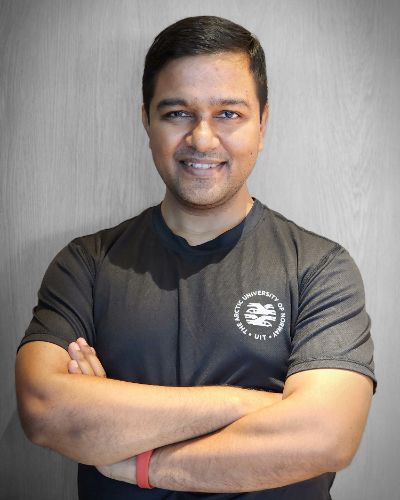}}]{Dilip K. Prasad}  is an associate professor at the Department of Computer Science, UiT The Arctic University of Norway. He received his Ph.D. and B.Tech degree in Computer Science and Engineering from Nanyang Technological University, Singapore, and the Indian Institute of Technology (Indian School of Mines), Dhanbad, India in 2013 and 2003 respectively.  Prior to Ph.D., he worked for 5 years with IBM, Infosys, Mediatek, and Philips. He was selected as a fellow for the Kauffman Global Scholarship in 2011. He is a co-author of book titled "Interpretability in Deep Learning", Springer, 2023.

\end{IEEEbiography}


\vfill

\clearpage

\twocolumn[%
   \begin{center}
     {\huge\sffamily Supplementary - Online Learning under Haphazard Input Conditions: A Comprehensive Review and Analysis}\\
   \end{center}
   \vspace{2ex}%
]

\title{Supplementary - Online Learning under Haphazard Input Conditions: A Comprehensive Review and Analysis}

\maketitle

\section*{I. Benchmarking}

We report the outcomes of our experiments using four metrics --- number of errors, accuracy, AUROC, and AUPRC --- not included in the main manuscript. To facilitate readability and comparison, we additionally provide balanced accuracy and execution time, which are reported in the main manuscript. Tables \ref{tab:results_synthetic_p25_small}, \ref{tab:results_synthetic_p50_small}, and \ref{tab:results_synthetic_p75_small} display the results of all models on small synthetic datasets at $p$ values of 0.25, 0.5, and 0.75, respectively. The outcomes for medium and large synthetic datasets at these $p$ values are detailed in Tables \ref{tab:results_synthetic_medium} and \ref{tab:results_synthetic_large}, respectively. Finally, Table \ref{tab:results_real_data} details the models' performance on real datasets across small, medium, and large categories.

\section*{II. Poor Performance of ORF\textsuperscript{3}V}

Tables \ref{tab:results_synthetic_p25_small}, \ref{tab:results_synthetic_p50_small}, \ref{tab:results_synthetic_p75_small}, \ref{tab:results_synthetic_medium}, and \ref{tab:results_synthetic_large} demonstrate that ORF\textsuperscript{3}V underperforms significantly. In contrast, the performance outcomes presented in its foundational article surpass our ORF\textsuperscript{3}V benchmarks \cite{schreckenberger2023online}. This discrepancy stems from the assumption and size of the storage buffer discussed in Section VI-B of the main manuscript. The original study utilizes a 500-instance storage buffer, significantly deviating from the online learning setting's principles. Conversely, our benchmark constrains the storage buffer to 20 instances, resulting in a notably reduced performance for ORF\textsuperscript{3}V.

\section*{III. PRISMA search: 8th March 2024}

The final literature search was performed on March 8th, 2024, to identify any new methods that emerged in the last few months in addition to the 9 methods reported in the main manuscript. We found three articles: DCDF2M \cite{sajedi2024data}, OIL \cite{lee2023study}, and OLCF \cite{zhou2023online}. We chose to exclude these methods from our analysis. Below we provide a brief overview of these methods and why they were excluded.

\paragraph{DCDF2M} Data stream classification in dynamic feature space using feature mapping (DCDF2M) presents a general algorithm for haphazard input modeling independent of any specific classifier, adaptable to the classifier most appropriate for the target application. 
DCDF2M is somewhat similar to OCDS\cite{he2019online}, utilizing the graph-based approach. However, DCDF2M employs a threshold to determine if the relation between two features is significant enough, dropping insignificant relations and allowing feature independence. It also characterizes feature relations using a polynomial of degree $D$. DCDF2M provides a pruning mechanism that activates when two features remain independent over a certain number of pre-determined instances. We did not include DCDF2M in our benchmarking due to code unavailability, the model's complexity, and reproducibility issues.

\paragraph{OIL} Online Imputation-based Learning (OIL) is a two-stage process to model haphazard inputs. Initially, OIL utilizes an imputation method --- such as mean, mode, kNN, Mice \cite{van2011mice}, or MissForest \cite{stekhoven2012missforest} --- to construct a complete feature space, and subsequently employs an adaptive random forest \cite{gomes2017adaptive} for making predictions. However, OIL operates under the significant assumption of maintaining a storage buffer for 500 instances, which contradicts the core principles of online learning. Even when limiting this assumption to a buffer of 20 instances, the method by which OIL manages sudden features after a substantial interval, such as 1000 instances, or addresses obsolete features, remains uncertain. Due to these considerations, we have omitted OIL from our main manuscript's comparative analysis.

\paragraph{OLCF} Online subspace Learning method under Capricious Features (OLCF) introduces a subspace estimator that maps heterogeneous haphazard inputs feature space to a low-dimensional subspace. A classifier is learned in this latent subspace for final predictions. Although OLCF is intended for haphazard inputs, it assumes that dimensionality changes ($d_{t-1} \neq d_t$) only because of sudden or obsolete features. However, in the context of haphazard inputs, missing features and missing values also result in dimensionality changes. Therefore, OLCF, in its current form, cannot fully address the complexities of haphazard inputs. Additionally, the code for OLCF is unavailable, leading us to exclude it from our benchmarking process.

\section*{IV. Datasets}

The reference and the link of each dataset corresponding to its category are presented in this section.
The Small datasets include: WPBC\footnote{\url{https://archive.ics.uci.edu/dataset/16/breast+cancer+wisconsin+prognostic}} \cite{mangasarian1995breast}, ionosphere\footnote{\url{https://archive.ics.uci.edu/ml/datasets/ionosphere}} \cite{sigillito1989classification}, WDBC\footnote{\url{https://archive.ics.uci.edu/dataset/17/breast+cancer+wisconsin+diagnostic}} \cite{mangasarian1995breast}, australian\footnote{\url{https://archive.ics.uci.edu/dataset/143/statlog+australian+credit+approval}} \cite{quinlan1987simplifying}, WBC\footnote{\url{https://archive.ics.uci.edu/dataset/15/breast+cancer+wisconsin+original}} \cite{mangasarian1990cancer}, diabetes-f\footnote{\url{https://www.kaggle.com/datasets/uciml/pima-indians-diabetes-database}} \cite{smith1988using}, crowdsense(c3) and crowdsense(c5)\cite{camprodon2019smart}, german\footnote{\url{https://archive.ics.uci.edu/dataset/144/statlog+german+credit+data}} \cite{misc_statlog_(german_credit_data)_144}, Italy Power Demand\footnote{\url{https://www.timeseriesclassification.com/description.php?Dataset=ItalyPowerDemand}} (IPD) \cite{keogh2006intelligent}, svmguide3\footnote{\url{https://www.csie.ntu.edu.tw/~cjlin/libsvmtools/datasets/binary.html}} \cite{hsu2003practical}, kr-vs-kp\footnote{\url{https://archive.ics.uci.edu/dataset/22/chess+king+rook+vs+king+pawn}} \cite{shapiro1984role}, spambase\footnote{\url{https://archive.ics.uci.edu/dataset/94/spambase}} \cite{misc_spambase_94}, and spamassasin \footnote{\url{https://spamassassin.apache.org/old/publiccorpus/}} \cite{katakis2005utility}. The datasets classified as Medium include: magic04\footnote{\url{https://archive.ics.uci.edu/dataset/159/magic+gamma+telescope}} \cite{bock2004methods}, imdb\footnote{\url{https://ai.stanford.edu/~amaas/data/sentiment/}} \cite{maas2011learning}, and a8a\footnote{\url{https://www.csie.ntu.edu.tw/~cjlin/libsvmtools/datasets/binary.html}} \cite{kohavi1996scaling}. The Large datasets are comprised of diabetes\_us\footnote{supplementary materials at \url{https://www.hindawi.com/journals/bmri/2014/781670/}} \cite{strack2014impact}, SUSY\footnote{\url{https://archive.ics.uci.edu/dataset/279/susy}} \cite{baldi2014searching}, and HIGGS\footnote{\url{https://archive.ics.uci.edu/dataset/280/higgs}} \cite{baldi2014searching}.

\input{Tables/Tab_Result_Synthetic_p25_small}
\input{Tables/Tab_Result_Synthetic_p50_small}
\input{Tables/Tab_Result_Synthetic_p75_small}
\input{Tables/Tab_Result_Synthetic_Medium}
\input{Tables/Tab_Result_Synthetic_Large}
\input{Tables/Tab_Result_Real}

\end{document}

%% file: FigureTex/Fig_OnlineLearning.tex
\begin{figure}[t]
    \begin{center}
    \includegraphics[width=0.85\columnwidth, trim={0.0cm 0.0cm 0.0cm 0.0cm}, clip]{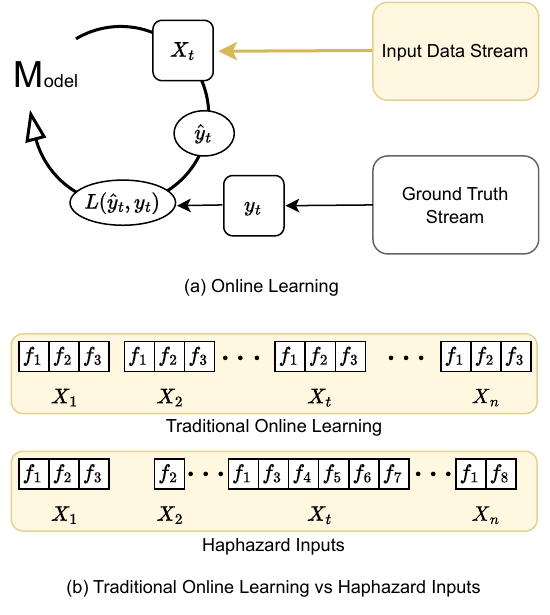} 
    \end{center}
    \caption{(a) The working of online learning method. (b) The fixed input feature space of traditional online learning versus the variable input feature space of haphazard inputs.}
    \label{fig:onlinelearning}
\end{figure}

%% file: FigureTex/Fig_PrismaGuideline.tex
\begin{figure*}[t]
    \begin{center}
    \includegraphics[width=\textwidth, trim={0.0cm 0.0cm 0.0cm 0.0cm}, clip]{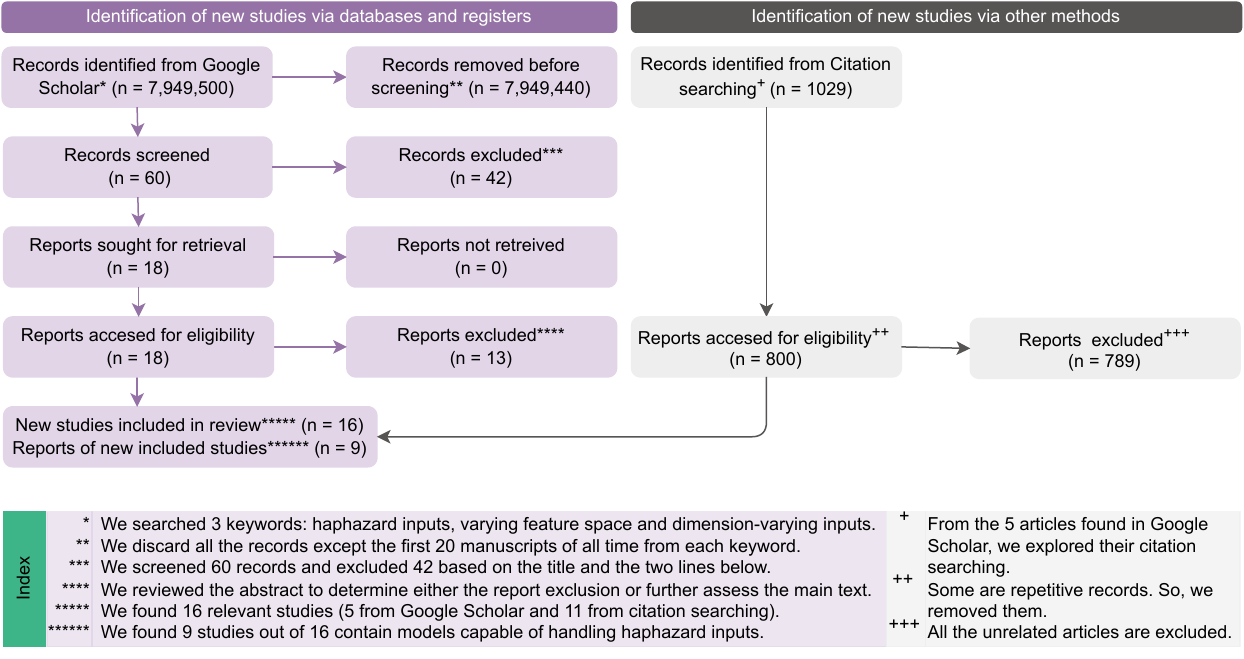} 
    \end{center}
    \caption{PRISMA \cite{page2021prisma} flowchart of our systematic review.}
    \label{fig:prisma}
\end{figure*}

%% file: FigureTex/Fig_CharacteristicsExample.tex
\begin{figure}[t]
    \begin{center}
    \includegraphics[width=\columnwidth, trim={0.0cm 0.0cm 0.0cm 0.0cm}, clip]{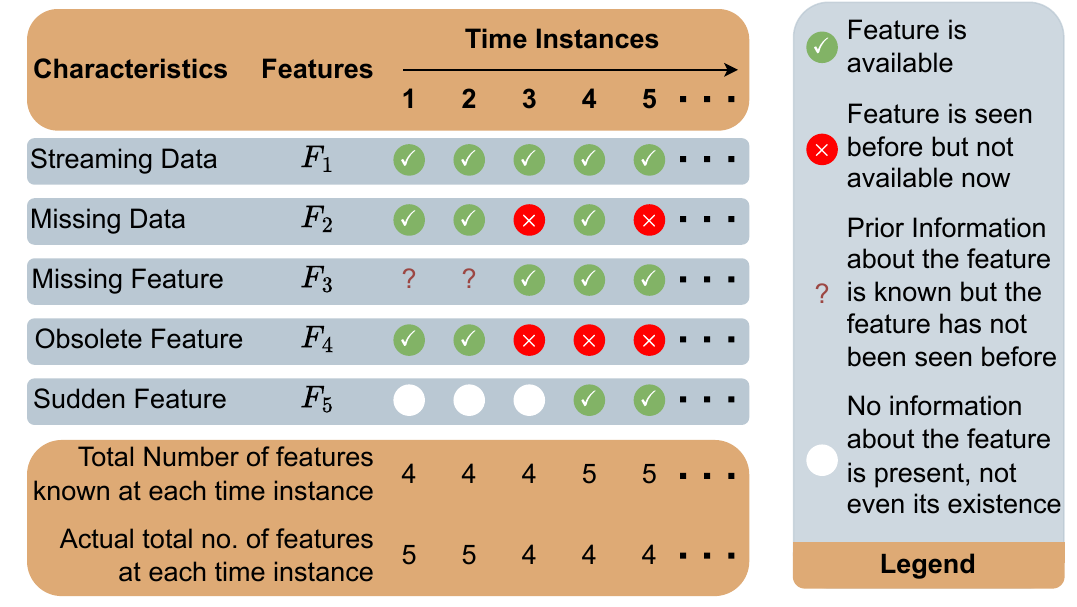} 
    \end{center}
    \caption{An example of haphazard inputs showcasing its characteristics.}
    \label{fig:CharacteristicsExample}
\end{figure}

%% file: Tables/Tab_CharacterisitcsOfHaphazardInputs.tex
\begin{table*}[t]
  \caption{The inability of online learning methods to handle haphazard inputs even with additional techniques like imputation, extrapolation, priori information, and Gaussian noise. This table is adapted from the Aux-Drop article \cite{agarwal2023auxdrop}.}
  \label{tab:comparisonOfmodels}
  \centering
  \begin{tabular}{lcccccc}
  \toprule
  Characteristics & Models  & Online & ODL & ODL  & ODL & ODL\\
  & for & Learning & + & + & + & +\\
  & Haphazard & Methods - & Online Data & Extrapolation & Prior & Gaussian\\
  & Inputs & ODL \cite{sahoo2017online} & Imputation &  & Information & Noise\\
  \midrule
  Streaming data & $\checkmark$ & $\checkmark$ & $\checkmark$ &  $\checkmark$ & $\checkmark$ & $\checkmark$ \\
  Missing data & $\checkmark$ & $\times$ & $\checkmark$ & $\times$ &  $\checkmark$ & $\checkmark$\\
  Missing features & $\checkmark$ & $\times$ & $\times$ & $\times$ &  $\times$ & $\checkmark$\\
  Obsolete features & $\checkmark$ & $\times$ & $\times$ &  $\checkmark$ & $\times$ & $\checkmark$ \\
  Sudden features & $\checkmark$ & $\times$ & $\times$ &  $\times$ & $\times$ & $\times$\\
  Unknown no. of features & $\checkmark$ & $\times$ & $\times$ & $\times$ & $\times$ & $\times$ \\
  \bottomrule
  \end{tabular}
\end{table*}

%% file: Tables/Tab_DatasetDescription4.tex
\begin{table*}[t]
  \centering
  \caption{The description of all the datasets is given here. The meaning of each notation in the table is as follows: IPD - Italy Power Demand, \#Inst. - number of instances, \#Feat. - number of features, Imbalance Ratio - \% of positive class instances, Real Data - $\times$ and $\checkmark$ means the real and synthetic data, respectively, Data Type - Num. and Cat. means numerical and categorical data type, respectively, \#Missing Values - number of missing values and $\times$ means zero missing values. }
  \label{tab:datasetdescription}
  \begin{adjustbox}{width=\linewidth,center}
  \begin{tabular}{r@{\hskip 4mm}ccccccrll}
  \toprule
  \textit{Dataset} &  \#Inst. & \#Feat. & Imbalance & Year & Real & Data & \#Missing & Field & Positive\\
  &  & & Ratio & & Data & Type & Values & of Data & Class\\
  \midrule
  
  \rowcolor{lightgray} \multicolumn{10}{c}{Small} \vspace{1mm}\\
  \textit{WPBC}\cite{mangasarian1995breast} & 198 & 33 & 23.74\% & 1995 & $\times$ & Num. & 4 & Breast cancer & Recurrence before 24 months \\
  \textit{ionosphere}\cite{sigillito1989classification} & 351 & 34 & 64.10\% & 1989 & $\times$ & Num. & $\times$ &  Radar & Shows evidence of ionosphere\\
  \textit{WDBC}\cite{mangasarian1995breast} & 569 & 30 & 37.26\% & 1995 & $\times$ & Num. & $\times$ & Breast Cancer & Malignant Tumor\\
  \textit{australian}\cite{quinlan1987simplifying} & 690 & 14 & 44.49\% & 1987 & $\times$ & Mix & 67 & Credit Card & Customer credit approved\\
  \textit{WBC}\cite{mangasarian1990cancer}  & 699 & 9 & 34.48\% & 1990 & $\times$ & Num. & 16 & Breast Cancer & Malignant Tumor\\
  \textit{diabetes-f}\cite{smith1988using} & 768 & 8 & 34.90\% & 1988 & $\times$ & Num. & $\times$ & Diabetes & Have diabetes\\
  \textit{crowdsense(c3)}\cite{camprodon2019smart} & 786 & 954 & 91.35\%& 2019 & $\checkmark$ & Num. & 597046 & Government’s restriction & High restriction severity\\
  \textit{crowdsense(c5)}\cite{camprodon2019smart} & 786 & 954 & 10.94\% & 2019 & $\checkmark$ & Num. & 597046 & Government’s restriction & High restriction severity \\
  \textit{german}\cite{misc_statlog_(german_credit_data)_144} & 959 & 24 & 71.32\% & 1994 & $\times$ & Num. & $\times$ & Credit Risk & Good\\
  \textit{$\text{IPD}^*$}\cite{keogh2006intelligent} & 1096 & 24 & 49.91\% & 2006 & $\times$ & Num. & $\times$ & Electric Power & Demand: October to March\\
  \textit{svmguide3}\cite{hsu2003practical} & 1243 & 21 & 76.19\% & 2003 & $\times$ & Num. & $\times$ & Vehicle & Traffic lights \\
  \textit{kr-vs-kp}\cite{shapiro1984role} & 3196 & 36 & 52.22\% & 1984 & $\times$ & Cat. & $\times$ & Chess & Won\\
  \textit{spambase}\cite{misc_spambase_94} & 4601 & 57 & 39.40\% & 1999 & $\times$ & Num. & $\times$ & Email Classification & Spam  \\
  \textit{spamassasin}  \cite{katakis2005utility} & 6047 & 7500 & 31.00\% & 2005 & $\checkmark$ & Text & 44536078 & Email Classification & Spam \vspace{1mm}\\
  
  \rowcolor{lightgray} \multicolumn{10}{c}{Medium} \vspace{1mm} \\
  \textit{magic04}\cite{bock2004methods} & 19020 & 10 & 64.84\% & 2004 & $\times$ & Num. & $\times$ & Gamma particles & Photons caused by gammas\\
  \textit{imdb}\cite{maas2011learning} & 25000 & 7500 & 50\% & 2011 & $\checkmark$ & Num. & 184415419 & Movie sentiment & Positive\\
  \textit{a8a}\cite{kohavi1996scaling} & 32561 & 123 & 75.92\% & 1994 & $\times$ & Cat. & $\times$ & Income & greater than \$50K/yr \vspace{1mm}\\
  
  \rowcolor{lightgray} \multicolumn{10}{c}{Large} \vspace{1mm} \\ 
  \textit{diabetes\_us}\cite{strack2014impact}  & 101766 & 47 & 11.16\% & 2014 & $\checkmark$ & Mix & 374017 & Diabetes & Readmitted before 30 days\\ 
  \textit{SUSY}\cite{baldi2014searching}  & 1M & 8 & 45.79\% & 2014 & $\times$ & Num. & $\times$ & Monte-Carlo Simulation & Susy are produced\\ 
  \textit{HIGGS}\cite{baldi2014searching} & 1M & 21 & 52.97\% & 2014 & $\times$& Num. & $\times$ & Monte-Carlo Simulation & Higgs boson are produced\\
  
  \bottomrule
  \end{tabular}
  \end{adjustbox}
 \end{table*}

%% file: FigureTex/Fig_ModelEmergence.tex
\begin{figure*}[t]
    \begin{center}
    \includegraphics[width=.8\textwidth, trim={0.0cm 0.0cm 0.0cm 0.0cm}, clip]{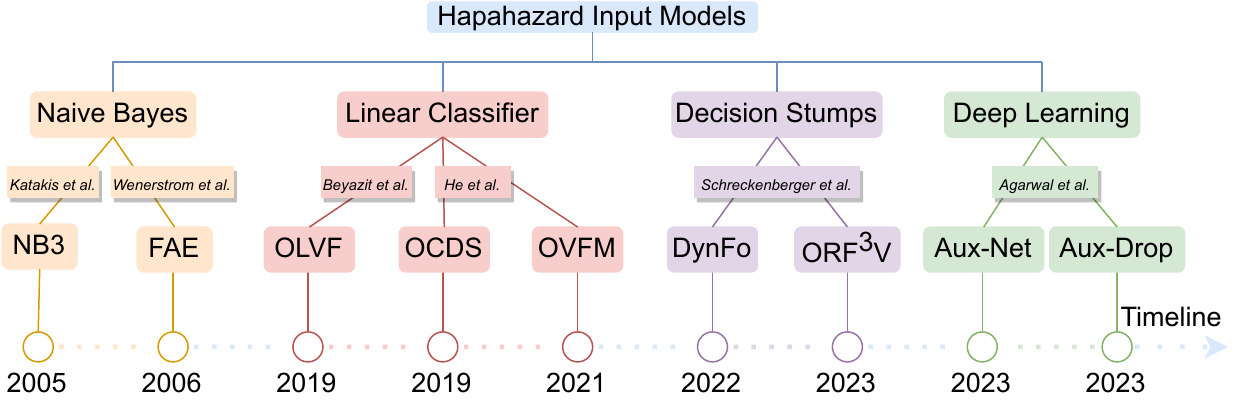} 
    \end{center}
    \caption{Timeline of all the models capable of handling haphazard inputs.}
    \label{fig:timeline}
\end{figure*}

%% file: Tables/Tab_classificationbasedonassumption_results_2.tex

\begin{table*}[t]
  \caption{Qualitative comparison of haphazard inputs models. The full form of the acronym used in the Assumption column are as follows: FI - Feature Independence, FR - Feature Relatedness, PT - Pretraining, SB - Storage Buffer, BF - Base Feature. The meaning of the symbols used in the code column is: O - we implemented these methods, ? - we modified the code, $\checkmark$ - we used the available implementation as it is. The more number of $*$ denotes the best model.}
  \label{tab:modelcomparison}
  \centering
  \begin{tabular}{rrccccccccc}
  \toprule
  \multirow{2}{*}{Category} & \multirow{2}{*}{Model} & \multirow{2}{*}{Assumption} & Feature & \multirow{2}{*}{Performance} & Data & Prediction & \multirow{2}{*}{Speed} & Feature & Determ- & \multirow{2}{*}{Code}\\
  & & & Reconstruct & & Scalability & Consistency & & Scalability & inistic &\\
  \midrule
  Naive & NB3 \cite{katakis2005utility} & FI, PT & $\times$ & $*$ & $*$ & - & $*****$ & $***$ $*$ & $\checkmark$ & O \\
  Bayes & FAE \cite{wenerstrom2006temporal} &  FI, PT & $\times$ & $*$ & $*$ & - & $*$ $*$ & $***$ $*$ & $\checkmark$ & O \\
  \midrule
  
  \multirow{2}{*}{Linear} & OLVF \cite{beyazit2019online}& $\times$ & $\times$ & $***$ $*$ & $*$ & - & $*****$ & $*****$ & $\checkmark$ & O \\
  \multirow{2}{*}{Classifier} & OCDS \cite{he2019online} & FR & $\checkmark$ & $*$ $*$ & $*$ & $***$ & $***$ & $*****$ & $\times$ & O \\
  & OVFM \cite{he2021online} & FR, SB & $\checkmark$ & $*****$ & $*$ & $***$ $*$& $***$ & $***$ $*$ & $\times$ & ? \\
  \midrule

  Decision & DynFo \cite{schreckenberger2022dynamic} & SB & $\times$ & $***$ $*$ & $*$ & $*****$ & $*$ & $*$ & $\times$ & O \\
  Stumps & ORF\textsuperscript{3}V \cite{schreckenberger2023online} &  SB & $\times$ & $*$  & $*$ $*$ & $*****$ & $*$ $*$ & $*****$ & $\times$ & O \\
  \midrule

  Deep & Aux-Net \cite{agarwal2023auxiliary} & BF & $\times$ & $*$ $*$ & $*$ & $*$ & $*$ $*$ & $***$ $*$ & $\times$ & O\\
  Learning & Aux-Drop \cite{agarwal2023auxdrop} &  BF & $\times$ & $***$ $*$ & $*****$ & $***$ & $***$ & $*****$ & $\times$ & $\checkmark$\\
  \bottomrule
  \end{tabular}
\end{table*}

%% file: FigureTex/Fig_DecisionStump.tex
\begin{figure*}[!t]
    \begin{center}
    \includegraphics[width=.8\textwidth, trim={0.0cm 0.0cm 0.0cm 0.0cm}, clip]{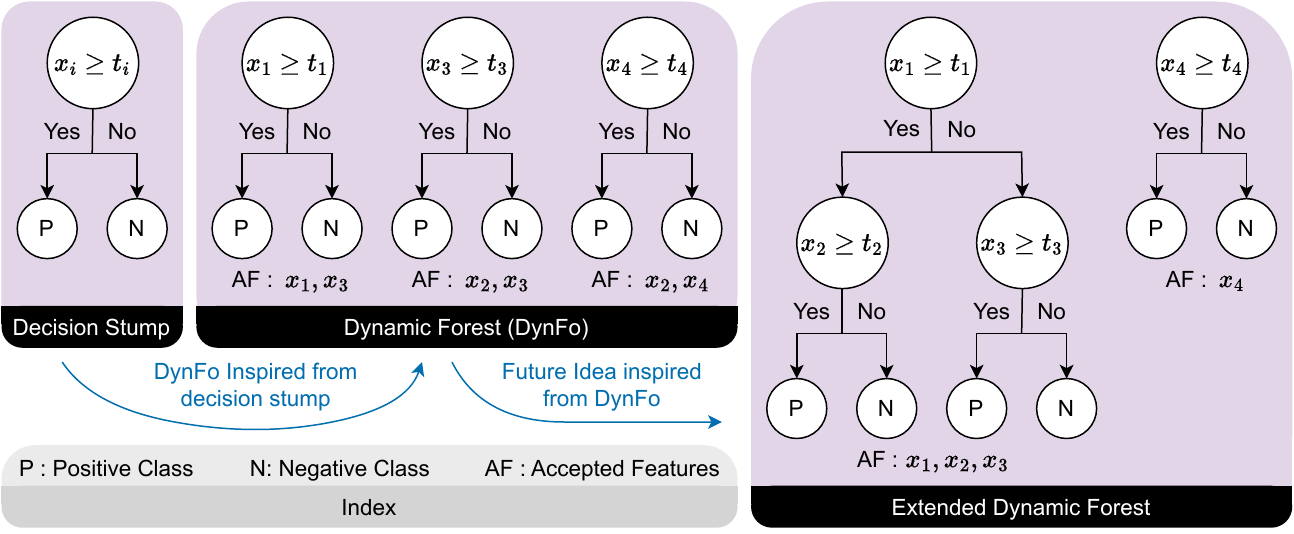} 
    \end{center}
    \caption{Decision stump and possible ideas. The central figure is adapted from the DynFo article \cite{schreckenberger2022dynamic}.}
    \label{fig:decisionstump}
\end{figure*}

%% file: Tables/Tab_Hyperparameters.tex
\begin{table*}[!t]
  \caption{Description of all the hyperparameters used in each model and their search values.}
  \label{tab:hyperparameter_description}
  \centering
\begin{tabular}{ clp{8.3cm}r } 
\toprule

\textbf{Model} & \textbf{Hyperparameter(s)} & \multicolumn{1}{c}{\textbf{Description}} & \multicolumn{1}{r}{\textbf{Search}} \\
\midrule

NB3 & n & proportion of total features to consider as the number of top features & [.2, .4, .6, .8, 1] \\
\midrule

\multirow{8}{*}{FAE} & m & number of instances after which learner's output is considered & 5\\
& \multirow{2}{*}{f} & threshold on the difference of youngest learner's feature set and current top M features & \multirow{2}{*}{0.15}\\
& \multirow{2}{*}{p} & number of consecutive instances a learner is under the threshold before being removed & \multirow{2}{*}{3}\\
& r & minimum number of instances between the 2 consecutive learners & 10\\
& N & number of instances to calculate accuracy over & 50\\
& M & proportion of total features to train new feature forest on & [.2, .4, .6, .8, 1] \\
\midrule

\multirow{4}{*}{OLVF} & C & loss bounding parameter for instance classifier & [.0001, .01, 1]\\
& $\bar{C}$ & loss bounding parameter for feature classifier & [.0001, .01, 1]\\
& B & proportion of selected features for sparsity & [.01, .1, .3, .5, .7, .9, 1]\\
& $\lambda$ & regularization parameter & [.0001, .01, 1]\\
\midrule

\multirow{6}{*}{OCDS} & T & number of instances after which 'p' (weighing factor) is updated & [8, 16] \\ 
& $\alpha$ & absorption scale parameter used in equation 10 of \cite{he2019online} & [.0001, .001, .01, .1, 1]\\
& $\beta_0$ & absorption scale introduced by us for the 1st term in eq. 9 of \cite{he2019online} & [.0001, .001, .01, .1, 1]\\
& $\beta_1$ & tradeoff parameter used in equation 9 of \cite{he2019online} & [.0001, .001, .01, .1, 1]\\
& $\beta_2$ & tradeoff parameter used in equation 9 of \cite{he2019online} & [.0001, .001, .01, .1, 1]\\
\midrule

\multirow{6}{*}{OVFM} & decay\_choice (dc) & decay update rules choices (see original code \cite{he2021online}) & [0, 1, 2, 3, 4] \\
& contribute\_error\_rate (ce) & used in the original code implementation of classifiers in \cite{he2021online} & [.001, .005, .01, .02, .05]\\
& window\_size (ws) & a (buffer-like) window to store data instances & 20\\ 
& decay\_coef\_change (dc) & set 'True' for learning rate decay, 'False' otherwise & [True, False]\\
& batch\_size\_denominator (bs) & used in update step in case of learning rate decay & [8, 10, 20] \\
& batch\_c (bc) & added to the denominator for stability in learning rate decay & 8 \\
\midrule

\multirow{9}{*}{DynFo} & $\alpha$ & impact on weight update & [.1, .5] \\
& $\beta$ & probability to keep the weak learner in ensemble & [.5, .8]\\
& $\delta$ & fraction of features to consider (bagging parameter) & [.001, .01]\\
& $\epsilon$ & penalty if the split of decision stump is not in the current instance & [.001, .01]\\
& $\gamma$ & threshold for error rate & [.5, .8]\\
& M & number of learners in the ensemble & [500, 1000]\\
& N & buffer size of instances & 20\\
& $\theta{1}$ & lower bounds for the update strategy & .05 \\
& $\theta{2}$ & upper bounds for the update strategy & [.6, .75] \\
\midrule

\multirow{7}{*}{ORF3V} & forestSize (fS) & number of stumps for every feature & [3, 5, 10]\\
& replacementInterval (rI) & instances after which stumps might get replaced & [5, 10]\\
& updateStrategy (uS) & strategy to replace stumps & ["oldest", "random"]\\
& replacementChance (rC) & probability to not replace stump for "random" update strategy & .7 \\
& windowsize (ws) & buffer storage size on which to determine feature statistics & 20\\
& $\alpha$ & weight update parameter & [.01, .1, .3, .5, .9]\\
& $\delta$ & pruning parameter & .001 \\
\midrule

\multirow{6}{*}{Aux-Net} & n\_base\_layer (nb) & number of base layers & 5 \\
& n\_end\_layers (ne) & number of end layers & 5 \\
& n\_nodes\_layers (n) & number of nodes in each layer & 50 \\
& lr & learning rate & [.001, .005, .01, .05, .1, .3, .5] \\
& b & discount rate & .99\\
& s & smoothing parameter & .2\\
\midrule

\multirow{7}{*}{Aux-Drop} & max\_num\_hidden\_layers (nl) & number of hidden layers & 6\\
& neuron\_per\_hidden\_layer (n) & number of nodes in each hidden layer except the AuxLayer & 50 \\
& n\_neuron\_aux\_layer (na) & total number of neurons in the AuxLayer & $\sim$5$\times$num\_feat\\
& b & discount rate & .99\\
& s & smoothing parameter & .2\\
& lr & learning rate & [.001, .005, .01, .05, .1, .3, .5] \\
& dropout\_p (d) & dropout rate in the AuxLayer & [.3, .5] \\
\bottomrule

\end{tabular}

\end{table*}

%% file: Tables/Tab_Results_HyperParameterSearch_2.tex
\begin{table*}[!t]
  \caption{Best hyperparameters of the models based on balanced accuracy. In synthetic datasets, we consider $p = 0.5$ for hyperparameter search. The best value of $\lambda$ for OLVF is always found to be .0001 for all the datasets. Experiments marked with \textsuperscript{$\S$} employed heuristically set hyperparameters, without a search process, due to the high computational demands.}
  \label{tab:hyperparameters_noassumption}
  \centering
  \begin{adjustbox}{width=\linewidth,center}
  \begin{tabular}{r@{\hskip 4mm}rrrrrrrrr} 
  \toprule
  
  \multirow{2}{*}{\textit{Dataset}} & NB3 & FAE  & \multicolumn{1}{c}{OLVF} & \multicolumn{1}{c}{OCDS} & \multicolumn{1}{c}{OVFM} & \multicolumn{1}{c}{DynFo} & \multicolumn{1}{c}{ORF\textsuperscript{3}V} & Aux-Net & \multicolumn{1}{c}{Aux-Drop} \\
  
  & (n) & (M) & (C, $\bar{C}$, B) & (T, $\alpha$, $\beta_0$, $\beta_1$, $\beta_2$) & (dc, ce, dc, bs) & ($\alpha$, $\beta$, $\delta$, $\epsilon$, $\gamma$, M, $\theta{2}$) & (fS, rI, uS, $\alpha$) & (lr) & (na, lr, d)\\
  \midrule

  \rowcolor{lightgray} \multicolumn{10}{c}{Small} \vspace{1mm}\\
  \textit{WPBC} & 1.0 & 0.4 & 1, 1, 1& 16, .001, .0001, .0001, .01 & 2, .01, F, 20 & .5, .8, .001, .001, .5, 500, .75 & 10, 5, 'oldest', .01 & 0.001 & 200, .01, .5 \vspace{.5mm}\\

  \textit{ionosphere} & 0.2 & 0.2 & 1, 1, 1& 8, 1, 1, .0001, .01 & 4, .02, F, 20 & .1, .8, .001, .001, .5, 500, .75 & 5, 10, 'oldest', .9 & 0.001 & 200, .5, .3 \vspace{.5mm}\\

  \textit{WDBC} & 0.6 & 0.2 & .0001, .0001, 1& 8, .01, .0001, 1, .0001 & 0, .02, F, 8 & .1, .8, .001, .001, .8, 500, .75 & 10, 10, 'oldest', .9 & .01 & 200, .01, .5 \vspace{.5mm}\\

  \textit{australian} & 0.8 & 1.0 & .01, .0001, 1& 16, .0001, .0001, .01, .01 & 4, .01, F, 10 & .1, .8, .001, .001, .8, 500, .75 & 10, 10, 'oldest', .9 & .01 & 100, .005, .5 \vspace{.5mm}\\

  \textit{WBC} & 0.2 & 0.8 & .01, .0001, 1& 8, .0001, .0001, 1, .01 & 2, .02, F, 8 & .1, .5, .001, .001, .5, 500, .75 & 5, 5, 'oldest', .9 & .001 & 100, .1, .3 \vspace{.5mm}\\

  \textit{diabetes-f} & 1.0 & 1.0 & .01, .0001, .3 & 8, .0001, .01, .0001, .0001 & 2, .05, F, 8 & .5, .8, .001, .01, .5, 500, .75 & 3, 5, 'oldest', .3 & .05 & 100, .001, .5 \vspace{.5mm}\\

  \textit{crowdsense(c3)} & 0.6 & 0.2 & .0001, .0001, 1 & 8, .0001, .01, .0001, .01 & 4, .01, F, 20 & .5, .5, .01, .01, .5, 500, .75 & 10, 5, 'oldest', .01 & .001\textsuperscript{$\S$} & 5000, .001, .5 \vspace{.5mm}\\

  \textit{crowdsense(c5)} & 0.8 & 0.8 & .0001, .0001, 1 & 8, .01, .01, .01, .01 & 4, .05, F, 20 & .5, .5, .01, .001, .5, 500, .75 & 10, 5, 'oldest', .01 & .001\textsuperscript{$\S$} & 5000, .001, .3 \vspace{.5mm}\\

  \textit{german} & 1.0 & 0.2 & .01, .0001, .01 & 8, .01, .0001, 1, .0001 & 3, .005, F, 8 & .5, .5, .001, .01, .5, 500, .75 & 5, 5, 'oldest', .3 & .001 & 100, .001, .5 \vspace{.5mm}\\

  \textit{IPD} & 0.6 & 0.8 & 1, .01, 1 & 16, 1, 1, .0001, .01 & 4, .001, F, 20 & .1, .8, .001, .001, .5, 500, .75 & 10, 10, 'random', .1 & .001 & 100, .3, .3 \vspace{.5mm}\\

  \textit{svmguide3} & 0.2 & 0.4 & 1, 1, 1 & 16, 1, 1, .01, .0001 & 0, .001, F, 20 & .5, .5, .001, .01, .5, 500, .75 &  5, 10, 'random', .3 & .1 & 100, .001, .3 \vspace{.5mm}\\

  \textit{kr-vs-kp} & 1.0 & 1.0 & 1, 1, 1 & 16, .0001, .0001, .01, .01 & 4, .005, F, 20 & .1, .8, .001, .001, .5, 500, .75 & 5, 5, 'random', .1 & .001 & 200, .1, .3 \vspace{.5mm}\\

  \textit{spambase} & 1.0 & 0.2 & .01, .0001, 1& 16, .01, .01, .0001, .0001 & 4, .001, F, 20 & .1, .5, .001, .001, .8, 500, .75 & 10, 10, 'oldest', .01 & .005 & 300, .005, .5 \vspace{.5mm}\\

  \textit{spamassasin} & 0.2 & 0.2 & 1, .0001, 1 & 16, .001, .001, .001, .001\textsuperscript{$\S$} & 4, .001, F, 20\textsuperscript{$\S$} & .5, .5, .001, .001, .7, 1000, .6\textsuperscript{$\S$} & 10, 5, 'oldest', .01 & .01\textsuperscript{$\S$} & 30000, .01, .3\textsuperscript{$\S$} \vspace{.5mm}\\

  \rowcolor{lightgray} \multicolumn{10}{c}{Medium} \vspace{1mm}\\
  \textit{magic04} & 0.6 & 1.0 & .0001, .0001, 1 & 16, .01, .01, .0001, .0001 & 3, .005, F, 20 & .5, .5, .1, .001, .7, 1000, .6 & 10, 5, 'random', .01 & .5 & 100, .01, .3 \vspace{.5mm}\\

  \textit{imdb} & 0.4 & 0.8 & .01, .0001, 1 & 16, 1, .0001, .01, .01\textsuperscript{$\S$} & 4, .001, F, 20\textsuperscript{$\S$} & .5, .8, .001, .001, .7, 1000, .6\textsuperscript{$\S$} & 10, 5, 'oldest', .01\textsuperscript{$\S$} & .01\textsuperscript{$\S$} & 30000, .01, .3\textsuperscript{$\S$} \vspace{.5mm}\\

  \textit{a8a} & 0.2 & 0.2 & 1, .0001, 1 & 16, 1, 1, .0001, .0001 & 4, .001, F, 20\textsuperscript{$\S$} & .5, .5, .03, .001, .7, 1000, .6 & 10, 10, 'oldest', .1 & .01\textsuperscript{$\S$} & 500, .01, .3 \vspace{.5mm}\\
  
  \rowcolor{lightgray} \multicolumn{10}{c}{Large} \vspace{1mm}\\
  \textit{diabetes\_us} & 0.8 & 0.4 & .0001, .0001, 1& 16, .01, .01, 1, .0001 & 4, .001, F, 20\textsuperscript{$\S$} & .5, .5, .1, .001, .7, 1000, .6\textsuperscript{$\S$} & 10, 5, 'oldest', .01 & 0.05\textsuperscript{$\S$} & 200, .05, .3 \vspace{.5mm}\\

  \textit{SUSY} & 1.0 & 0.6 & .01, .01, 1 & 16, .0001, .01, .0001, .0001 & 4, .001, F, 20\textsuperscript{$\S$} & .5, .5, .4, .001, .7, 1000, .6\textsuperscript{$\S$} & 5, 10, 'random', .1\textsuperscript{$\S$} & 0.05\textsuperscript{$\S$} & 100, .05, .3\textsuperscript{$\S$} \vspace{.5mm}\\

  \textit{HIGGS} & 0.2 & 0.4 & .01, .0001, 1 & 8, .0001, .0001, .01, .0001 & 4, .001, F, 20\textsuperscript{$\S$} & .5, .5, .2, .001, .7, 1000, .6\textsuperscript{$\S$} & 5, 10, 'random', .1\textsuperscript{$\S$} & 0.05\textsuperscript{$\S$} & 100, .05, .3\textsuperscript{$\S$} \\
  \bottomrule
  
  \end{tabular}
  \end{adjustbox}
\end{table*}

%% file: Tables/Tables_Result_Small_Bacc_Time_Bold.tex
\begin{table*}[!th]
  \caption{Comparison of models on Small Datasets. The deterministic models --- NB3, FAE, and OLVF --- underwent a single execution. In contrast, the non-deterministic models were executed 5 times, with the mean $\pm$ standard deviation reported. A \textsuperscript{$\ddagger$} symbol indicates non-deterministic models that were run only once on specific datasets due to substantial time constraints. The bAcc stands for balanced Accuracy, Time is reported in seconds, and \textsuperscript{$\dagger$} denotes the real datasets. The win count for each model in ($p$ = .25)/($p$ = .5)/($p$ = .75)/real dataset is shown in the last row.}
  \label{tab:results_small}
  \centering
  \resizebox{.9\linewidth}{!}{%
  \begin{tabular}{l|c|cccccccccc}
  \toprule
  Dataset & $p$ & Metric & NB3 & FAE & OLVF & OCDS & OVFM & DynFo & ORF\textsuperscript{3}V & Aux-Net & Aux-Drop\\
  \midrule

  \multirow{6}{*}{WPBC} & 
  \multirow{2}{*}{0.25} 
  & bAcc & 49.48 & 48.93 & \textbf{58.08} & 49.88$\pm$1.24 & 46.73$\pm$1.01 & 56.58$\pm$1.40 & 50.54$\pm$0.49 & 52.08$\pm$0.95	& 49.89$\pm$2.45 \\
  & & Time & 0.05 & 0.08 & 0.02  & 0.04$\pm$0.00  & 2.21$\pm$0.01  & 2.57$\pm$0.01 & 0.35$\pm$0.00 & 7.86$\pm$0.38 & 1.07$\pm$0.04 \\
  \cmidrule{2-12}

  & 
  \multirow{2}{*}{0.5}
  & bAcc & 49.67 & 51.96 & \textbf{53.76} & 49.67$\pm$0.00 & 51.32$\pm$1.04 & 51.59$\pm$0.54 & 52.60$\pm$0.00	& 50.66$\pm$0.77 & 50.78$\pm$2.39	\\
  & & Time & 0.05 & 0.08 & 0.03 & 0.06$\pm$0.00 & 2.11$\pm$0.13 & 3.46$\pm$0.09 & 0.66$\pm$0.00 & 15.81$\pm$0.78 & 1.27$\pm$0.07\\
  \cmidrule{2-12}
    
  & 
  \multirow{2}{*}{0.75}  
  & bAcc & 49.67 & 47.42	& 48.37	& 49.34$\pm$0.00 & 46.18$\pm$2.05	& 49.64$\pm$0.91 & \textbf{50.71$\pm$0.00} & 50.25$\pm$0.81 & 50.51$\pm$0.65	\\
  & & Time & 0.07  & 0.12   & 0.03  & 0.05$\pm$0.00  & 1.28$\pm$0.00 & 3.43$\pm$0.09 & 1.00$\pm$0.05 & 22.73$\pm$0.58 & 1.10$\pm$0.06 \\
  \midrule

  \multirow{6}{*}{ionosphere} & 
  \multirow{2}{*}{0.25}
  & bAcc & 45.44	& 49.78 & \textbf{59.98} & 58.88$\pm$1.05 & 56.10$\pm$0.74 & 57.30$\pm$0.96 & 42.26$\pm$1.83 & 49.86$\pm$0.26 & 49.91$\pm$0.12	\\
  & & Time & 0.08 & 0.12 & 0.04 & 0.08$\pm$0.00 & 4.10$\pm$0.06 & 4.56$\pm$0.11 & 0.39$\pm$0.01 & 14.45$\pm$0.27 & 1.86$\pm$0.00 \\
  \cmidrule{2-12}
  & \multirow{2}{*}{0.5} 
  & bAcc & 49.41 & 49.78 & \textbf{66}	& 60.42$\pm$1.97 & 61.15$\pm$0.42 & 63.37$\pm$1.03 & 43.51$\pm$1.75	& 49.93$\pm$0.24 &  54.92$\pm$1.69	\\
  & & Time & 0.08 & 0.12 & 0.06 & 0.11$\pm$0.00 & 6.35$\pm$0.96 & 6.09$\pm$0.13 & 0.75$\pm$0.01 & 26.51$\pm$1.53 &  1.97$\pm$0.02\\
  \cmidrule{2-12}  
  & \multirow{2}{*}{0.75} 
  & bAcc & 49.78 &  49.78	& 65.6	& 57.96$\pm$0.75 & \textbf{70.25$\pm$0.82}	& 66.26$\pm$0.76 & 43.40$\pm$1.02 & 50.32$\pm$1.44	& 61.36$\pm$0.89	\\
  & & Time & 0.08 &   0.17  & 0.05  & 0.08$\pm$0.00  & 2.69$\pm$0.01 & 6.06$\pm$0.06 & 1.09$\pm$0.01 & 40.80$\pm$0.35 & 1.87$\pm$0.00\\
  \midrule

  \multirow{6}{*}{WDBC} & 
  \multirow{2}{*}{0.25} 
  & bAcc & 49.4 & 48.72 & 49.41 & 49.01$\pm$0.78	& 80.35$\pm$1.05 & \textbf{83.00$\pm$1.45} & 41.32$\pm$1.76 & 49.58$\pm$0.35 & 49.48$\pm$0.41	\\
  & & Time & 0.11 & 0.21 & 0.07   & 0.12$\pm$0.00  & 5.73$\pm$0.13  & 7.36$\pm$0.12 & 1.07$\pm$0.07 & 20.90$\pm$0.90 & 3.08$\pm$0.08\\
  \cmidrule{2-12}
  & \multirow{2}{*}{0.5} 
  & bAcc & 50.01 & 49.23 & 53.25 & 52.40$\pm$0.75 & 87.78$\pm$0.53 & \textbf{90.47$\pm$0.51} & 41.32$\pm$1.21 & 50.52$\pm$1.26 &  50.53$\pm$1.72	\\
  & & Time & 0.12 & 0.3 & 0.08 & 0.16$\pm$0.00 & 3.90$\pm$0.01 & 9.19$\pm$0.26 & 2.15$\pm$0.09 & 36.17$\pm$0.30 &  3.08$\pm$0.02\\
  \cmidrule{2-12}  
  & \multirow{2}{*}{0.75} 
  & bAcc & 50.1  & 49.95	& 50.43	& 50.27$\pm$1.43 & 50.53$\pm$11.97 & \textbf{90.64$\pm$0.34}	& 41.90$\pm$1.32 & 50.04$\pm$0.24 & 50.85$\pm$0.86 \\
  & & Time & 0.14  & 0.39   & 0.08  & 0.13$\pm$0.00  & 3.39$\pm$0.01 & 9.73$\pm$0.10 & 3.14$\pm$0.06 & 56.44$\pm$0.54 & 3.06$\pm$0.04 \\
  \midrule

  \multirow{6}{*}{australian} & 
  \multirow{2}{*}{0.25} 
  & bAcc & 49.89 & 59.69 & 53.09 & 49.79$\pm$0.40	& \textbf{65.37$\pm$0.42} & 50.06$\pm$0.42	& 45.11$\pm$0.86 & 49.88$\pm$0.36 & 51.51$\pm$2.00	\\
  &  & Time & 0.07 & 0.34 &  0.08  & 0.12$\pm$0.00  & 5.07$\pm$0.07  & 0.97$\pm$0.02 & 0.46$\pm$0.00 & 15.31$\pm$0.27 & 3.63$\pm$0.01 \\
  \cmidrule{2-12}
  & \multirow{2}{*}{0.5} 
  & bAcc & 53.62 & 57.01 & 56.22 & 50.76$\pm$1.18 & 74.96$\pm$0.53 & \textbf{75.88$\pm$0.23} & 44.11$\pm$0.65 & 50.61$\pm$1.32 & 55.77$\pm$1.88	 \\
  & & Time & 0.08 & 0.53 & 0.1 & 0.17$\pm$0.00 & 6.67$\pm$0.89 & 10.68$\pm$0.23 & 0.82$\pm$0.01 & 20.36$\pm$0.12 & 4.33$\pm$0.11 \\
  \cmidrule{2-12}  
  & \multirow{2}{*}{0.75} 
  & bAcc & 54.74 & 57	 & 57.11	& 50.30$\pm$0.72 & 53.45$\pm$7.34	& \textbf{78.66$\pm$0.36} & 42.86$\pm$1.20 & 49.95$\pm$0.07	& 59.22$\pm$3.86	\\
  & & Time & 0.09  & 0.76   & 0.08  & 0.13$\pm$0.00  & 5.24$\pm$0.02 & 11.72$\pm$0.09 & 1.22$\pm$0.08 & 27.54$\pm$0.20 & 3.68$\pm$0.08 \\
  \midrule

  \multirow{6}{*}{wbc} & 
  \multirow{2}{*}{0.25} 
  & bAcc & 50.66 & 53.4 & 52.02 & 50.71$\pm$2.12	 & \textbf{81.70$\pm$0.00} & 67.20$\pm$1.92 & 29.38$\pm$1.13 & 49.90$\pm$0.24 & 77.07$\pm$1.32	\\
  & & Time & 0.05 &  0.38 & 0.07  & 0.12$\pm$0.00   & 4.20$\pm$0.07  & 1.04$\pm$0.03 & 0.23$\pm$0.00 & 13.36$\pm$0.21 & 3.76$\pm$0.08 \\
  \cmidrule{2-12}
  & \multirow{2}{*}{0.5} 
  & bAcc & 49.78 & 52.64 & 55.38 & 51.06$\pm$0.70 & 87.75$\pm$0.00 & \textbf{93.05$\pm$0.26} & 31.82$\pm$2.22 & 58.49$\pm$8.23 & 82.25$\pm$1.72	\\
  & & Time & 0.05 & 0.40 & 0.09 & 0.16$\pm$0.00 & 4.26$\pm$0.04 & 28.10$\pm$0.53 & 0.37$\pm$0.00 & 16.21$\pm$0.28 &  4.23$\pm$0.13 \\
  \cmidrule{2-12}  
  & \multirow{2}{*}{0.75} 
  & bAcc & 49.88 & 53.32	& 66.46	& 50.35$\pm$3.07 & 88.37$\pm$0.00	& \textbf{93.91$\pm$0.35} & 31.18$\pm$1.18 & 78.46$\pm$4.57	& 83.79$\pm$2.17	\\
  & & Time & 0.06  & 0.55   & 0.08  & 0.12$\pm$0.00  & 5.06$\pm$0.03 & 26.52$\pm$0.13 & 0.50$\pm$0.00 & 19.72$\pm$0.21 & 3.72$\pm$0.04 \\
  \midrule

  \multirow{6}{*}{diabetes-f} & 
  \multirow{2}{*}{0.25} 
  & bAcc & 48.7 & 45.53 & 49.5  & 51.05$\pm$1.08	& \textbf{58.24$\pm$0.00} & 56.03$\pm$0.60	& 45.74$\pm$0.31 & 49.96$\pm$0.05 & 49.51$\pm$0.70	\\
  &  & Time & 0.06 & 0.48  & 0.11  & 0.17$\pm$0.00 & 2.44$\pm$0.07 & 1.67$\pm$0.25 & 0.36$\pm$0.00 & 14.26$\pm$0.30 & 4.09$\pm$0.04 \\
  \cmidrule{2-12}
  & \multirow{2}{*}{0.5} 
  & bAcc & 50.73 & 47.05 & 50.19 & 50.79$\pm$0.73 & \textbf{59.20$\pm$0.00} & 53.82$\pm$0.42 & 49.16$\pm$1.16 & 50.08$\pm$0.35 & 51.20$\pm$0.90	  \\
  & & Time & 0.06 & 0.67 & 0.13 & 0.18$\pm$0.00 & 1.82$\pm$0.02 & 10.48$\pm$0.28 & 0.58$\pm$0.01 & 17.36$\pm$0.17 & 5.41$\pm$0.28 \\
  \cmidrule{2-12}  
  & \multirow{2}{*}{0.75} 
  & bAcc & 49.6  & 49.28	& 49.17	& 50.34$\pm$0.73 & 55.66$\pm$8.93	& \textbf{57.34$\pm$0.63} & 46.70$\pm$1.20 & 49.94$\pm$0.05 & 50.37$\pm$0.85	\\
  & & Time & 0.07  & 0.68   & 0.11  & 0.13$\pm$0.00  & 1.55$\pm$0.01 & 12.88$\pm$0.19 & 0.81$\pm$0.01 & 20.33$\pm$0.27 & 4.11$\pm$0.08 \\
  \midrule

  \multirow{2}{*}{crowdsense(c3)\textsuperscript{$\dagger$}} & 
  & bAcc & 99.45 & \textbf{99.51} & \textbf{99.51} & 99.11$\pm$0.03 & 78.23$\pm$1.97 &  95.79$\pm$1.11 & 99.40$\pm$0.04	& 87.87$\pm$3.76 & 93.15$\pm$2.16\\
  & & Time & 4.14 & 9.66 & 0.28 & 39.84$\pm$0.47 & 354.31$\pm$4.99 & 69.89$\pm$0.62 & 33.59$\pm$0.06 & 4909.36$\pm$752.39 &  464.23$\pm$0.92\\
  \midrule
  
  \multirow{2}{*}{crowdsense(c5)\textsuperscript{$\dagger$}} & 
  & bAcc & 94.85 & 93.64 & \textbf{97.9} & 94.10$\pm$0.35	& 61.90$\pm$6.16 & 94.15$\pm$0.27 & 95.62$\pm$0.41 & 49.93$\pm$0.07	& 83.72$\pm$5.48\\
  & & Time & 4.02 & 58.52 & 0.27 & 43.73$\pm$0.61 & 354.56$\pm$5.32 & 68.67$\pm$0.39 & 33.80$\pm$0.09 & 5563.30$\pm$498.90 & 35.72$\pm$4.56\\
  \midrule

  \multirow{6}{*}{german} & 
  \multirow{2}{*}{0.25} 
  & bAcc & 51.37 & 49.82 &  \textbf{52.58}	& 49.96$\pm$1.11 & 52.38$\pm$0.00 & 50.58$\pm$0.27 & 49.68$\pm$0.21	& 50.32$\pm$0.45 & 50.14$\pm$0.67	\\
  & & Time & 0.16 &  0.3   &  0.15  & 0.19$\pm$0.00  & 13.17$\pm$0.06 & 29.36$\pm$0.48 & 0.79$\pm$0.00 & 30.12$\pm$0.76 & 5.15$\pm$0.09 \\
  \cmidrule{2-12}
  & \multirow{2}{*}{0.5} 
  & bAcc & 50.71 & 49.85 & 53.11 & 51.11$\pm$1.84 & \textbf{57.09$\pm$0.00} & 51.66$\pm$0.23 & 50.12$\pm$0.10 & 50.14$\pm$0.36 & 51.48$\pm$1.07 \\
  & & Time & 0.17 & 0.35 & 0.18 & 0.26$\pm$0.00 & 17.07$\pm$0.41 & 34.33$\pm$0.87 & 1.34$\pm$0.07 & 45.73$\pm$0.78 & 6.14$\pm$0.16 \\
  \cmidrule{2-12}  
  & \multirow{2}{*}{0.75} 
  & bAcc & 49.89 & 49.78 & \textbf{53.13}	& 51.88$\pm$0.95 & 52.56$\pm$0.62	& 50.71$\pm$0.14 & 50.18$\pm$0.15	& 49.98$\pm$0.05 & 51.24$\pm$1.08	\\
  & & Time & 0.19  & 0.55  & 0.16   & 0.20$\pm$0.00  & 15.31$\pm$0.04 & 36.00$\pm$0.26 & 1.78$\pm$0.01 & 72.57$\pm$2.86 & 5.30$\pm$0.33 \\
  \midrule

  \multirow{6}{*}{IPD} & 
  \multirow{2}{*}{0.25} 
  & bAcc & 49.82 & 50.09 & 64.97 & 58.07$\pm$1.11 & \textbf{76.00$\pm$0.46} & 75.57$\pm$0.75 & 47.93$\pm$0.31	& 60.24$\pm$1.00 & 60.63$\pm$0.77	\\
  & & Time & 0.18 & 1.14   & 0.15  & 0.22$\pm$0.00  & 9.93$\pm$0.59  & 14.25$\pm$0.07 & 8.25$\pm$0.09   & 33.88$\pm$0.77 & 5.85$\pm$0.08 \\
  \cmidrule{2-12}
  & \multirow{2}{*}{0.5} 
  & bAcc & 51.26 & 50.09 & 78.38 & 67.03$\pm$0.89 & \textbf{88.10$\pm$0.31} & 84.83$\pm$0.54 & 49.13$\pm$0.25 & 72.52$\pm$2.33 & 70.32$\pm$1.29	\\
  & & Time & 0.19 & 1.45 & 0.16 & 0.29$\pm$0.00 & 14.38$\pm$3.28 & 16.86$\pm$0.22 & 16.22$\pm$0.20 & 53.76$\pm$1.76 &  6.92$\pm$0.12\\
  \cmidrule{2-12}  
  & \multirow{2}{*}{0.75} 
  & bAcc & 50.06 & 49.82	& 87.87	&  72.35$\pm$0.69 & 88.59$\pm$0.98 & \textbf{90.00$\pm$0.20} & 48.72$\pm$0.65 & 79.94$\pm$1.70	& 76.42$\pm$1.73 \\
  & & Time & 0.23  & 1.73   & 0.14  &  0.23$\pm$0.00  & 5.43$\pm$0.01 & 18.73$\pm$0.09 & 24.29$\pm$0.06 & 82.54$\pm$1.82 & 6.41$\pm$0.48 \\
  \midrule
  
  \multirow{6}{*}{svmguide3} & 
  \multirow{2}{*}{0.25} 
  & bAcc & 49.75 & 49.95 &  52.09	& 50.84$\pm$0.37 & \textbf{54.26$\pm$0.72} & 50.79$\pm$0.33 & 49.54$\pm$0.43	& 49.99$\pm$0.02 & 50.21$\pm$0.40	\\
  & & Time & 0.17 & 0.33   &  0.14  & 0.24$\pm$0.00  & 11.71$\pm$0.58 & 37.24$\pm$0.20 & 4.17$\pm$0.06  & 35.78$\pm$1.10 & 6.90$\pm$0.36 \\
  \cmidrule{2-12}
  & \multirow{2}{*}{0.5} 
  & bAcc & 49.95 & 49.95 & 53.46 & 51.90$\pm$0.84 & \textbf{58.13$\pm$0.41} & 51.02$\pm$0.41 & 49.71$\pm$0.24 & 49.97$\pm$0.03 &  50.61$\pm$0.45	\\
  & & Time & 0.18 & 0.33 & 0.19 & 0.33$\pm$0.00 & 7.96$\pm$0.04 & 42.81$\pm$1.82 & 7.93$\pm$0.14 & 52.31$\pm$0.20 &  7.44$\pm$0.08 \\
  \cmidrule{2-12}  
  & \multirow{2}{*}{0.75} 
  & bAcc & 49.95 & 49.95	& 53.64	& 50.36$\pm$0.46 & \textbf{59.70$\pm$0.30}	& 50.60$\pm$0.09 & 49.86$\pm$0.10 & 49.97$\pm$0.03 & 50.68$\pm$0.68	\\
  & & Time & 0.18  & 0.4    & 0.16  & 0.25$\pm$0.00  & 7.74$\pm$0.04 & 46.43$\pm$0.20 & 10.71$\pm$0.16 & 81.69$\pm$3.73 & 7.49$\pm$0.27 \\
  \midrule

  \multirow{6}{*}{kr-vs-kp} & 
  \multirow{2}{*}{0.25} 
  & bAcc & 50.35 & 49.68 & 52.47 & 50.21$\pm$0.52 & \textbf{62.13$\pm$0.00} & 54.42$\pm$0.66 & 44.16$\pm$0.18	& 50.83$\pm$0.76 & 53.12$\pm$0.38	\\
  & & Time & 0.79 &  12.26 & 0.38  & 0.70$\pm$0.00  & 105.13$\pm$13.74 & 72.43$\pm$3.16 & 5.27$\pm$0.09 & 127.93$\pm$3.05 & 17.35$\pm$0.17 \\
  \cmidrule{2-12}
  & \multirow{2}{*}{0.5} 
  & bAcc & 50.7 & 50.29 & 57.44 & 51.50$\pm$0.55	& \textbf{71.50$\pm$0.00} &	57.97$\pm$0.19	& 44.18$\pm$0.38 & 58.00$\pm$1.11 & 59.35$\pm$0.75	\\
  & & Time & 0.86 & 16.99 & 0.46 & 1.00$\pm$0.08 & 68.52$\pm$0.02 & 75.84$\pm$1.07 & 7.54$\pm$0.08 & 237.75$\pm$3.50 & 22.58$\pm$1.53\\
  \cmidrule{2-12}  
  & \multirow{2}{*}{0.75} 
  & bAcc & 49.87 & 50.08	& 65.74	& 50.67$\pm$0.75 & \textbf{80.53$\pm$0.00}	& 58.72$\pm$0.14 & 43.96$\pm$0.23 & 63.66$\pm$2.23 & 64.22$\pm$0.31 \\
  & & Time & 1.04  & 16.42  & 0.43  & 0.74$\pm$0.00  & 335.12$\pm$123.47 & 83.40$\pm$1.08 & 9.74$\pm$0.08 & 390.14$\pm$2.77 & 18.10$\pm$0.82 \\
  \midrule
  
  \multirow{6}{*}{spambase} & 
  \multirow{2}{*}{0.25} 
  & bAcc & 50.54 &51.17 & 64.34 & 53.63$\pm$0.04	& 64.26$\pm$0.17 & \textbf{66.05$\pm$0.25}	& 48.94$\pm$0.09 & 49.92$\pm$0.06 & 55.57$\pm$0.99 \\
  & & Time & 1.56 & 25.36 &  0.56 & 1.68$\pm$0.00 & 134.27$\pm$14.13 & 138.10$\pm$0.30 & 19.54$\pm$0.14 & 280.91$\pm$1.43 & 26.58$\pm$1.28 \\
  \cmidrule{2-12}
  & \multirow{2}{*}{0.5} 
  & bAcc & 51.11 & 50.06 & 61.56 & 51.69$\pm$0.44 & \textbf{73.38$\pm$0.30} & 71.00$\pm$0.22 & 49.73$\pm$0.09 & 52.85$\pm$2.60 & 61.16$\pm$1.20	\\
  & & Time & 1.7 & 31.20 & 0.76 & 1.81$\pm$0.07 & 117.10$\pm$12.53 & 180.98$\pm$0.99 & 32.14$\pm$0.11 & 627.64$\pm$6.53 & 29.88$\pm$1.78 \\
  \cmidrule{2-12}  
  & \multirow{2}{*}{0.75} 
  & bAcc & 49.81 & 49.94	& 59.8	& 49.72$\pm$0.68 & 51.64$\pm$3.01	& \textbf{72.59$\pm$0.21} & 49.69$\pm$0.09 & 61.23$\pm$1.66	& 62.75$\pm$1.18	\\
  & & Time & 1.85  & 39.06  & 0.65  & 1.35$\pm$0.00  & 490.74$\pm$349.53 & 176.00$\pm$0.77 & 42.81$\pm$0.10 & 1071.04$\pm$9.10 & 27.26$\pm$2.21 \\
  \midrule

  \multirow{2}{*}{spamassasin\textsuperscript{$\dagger$}} &
  & bAcc & 92.96 & 95.99 & \textbf{99.6} & 96.39$\pm$0.00	& 99.28$\pm$0.09 & 97.61$\pm$0.14 & 94.05$\pm$0.04 & 97.28\textsuperscript{$\ddagger$} & 62.08$\pm$0.39\\
  & & Time & 254.32 & 5880.89 & 4.96 & 8809.23$\pm$856.44 & 32739.41$\pm$3369.92 & 2176.94$\pm$22.49 & 1741.31$\pm$22.76 & 67934.42\textsuperscript{$\ddagger$} & 18071.34$\pm$685.19\\
  \midrule

  \multicolumn{2}{c|}{\textit{\textbf{Win Count} (11/11/11/3)}} & \textit{bAcc} & \textit{0/0/0/0} & \textit{0/0/0/1} & \textit{3/2/1/3} & \textit{0/0/0/0} & \textit{6/6/3/0} &  \textit{2/3/6/0} & \textit{0/0/1/0} & \textit{0/0/0/0} & \textit{0/0/0/0} \\

  
  \bottomrule
  \end{tabular}  }
\end{table*}

%% file: Tables/Tables_Result_Medium_Large_Bacc_Time.tex
\begin{table*}[!th]
  \caption{Comparison of models on medium and large datasets. The deterministic models --- NB3, FAE, and OLVF --- underwent a single execution. In contrast, the non-deterministic models were executed 5 times, with the mean $\pm$ standard deviation reported. A \textsuperscript{$\ddagger$} symbol indicates non-deterministic models that were run only once on specific datasets due to substantial time constraints. The bAcc stands for balanced Accuracy, Time is reported in seconds, and \textsuperscript{$\dagger$} denotes the real datasets. The win count for each model in ($p$ = .25)/($p$ = .5)/($p$ = .75)/real dataset is shown in the last row.}
  \label{tab:results_medium_large}
  \centering
  \resizebox{.93\linewidth}{!}{%
  \begin{tabular}{l|c|cccccccccc}
  \toprule
  Dataset & $p$ & Metric & NB3 & FAE & OLVF & OCDS & OVFM & DynFo & ORF\textsuperscript{3}V & Aux-Net & Aux-Drop\\
  \midrule

  \rowcolor{lightgray} \multicolumn{12}{c}{Medium} \vspace{1mm}\\

  \multirow{6}{*}{magic04} & 
  \multirow{2}{*}{0.25} 
  & bAcc & 50.01 & 50.01 & 53.18 & 51.89$\pm$0.10 & 55.76$\pm$0.00 & 52.75$\pm$0.30	& 47.94$\pm$0.22 & 50.09$\pm$0.07 & \textbf{56.04$\pm$0.53} \\
  & & Time & 1.46 & 72.7 &  2.09 & 4.30$\pm$0.01 & 129.82$\pm$14.06 & 46.59$\pm$5.63 & 952.28$\pm$23.98 & 376.94$\pm$3.42 & 111.51$\pm$7.59 \\
  \cmidrule{2-12}
  & \multirow{2}{*}{0.5} 
  & bAcc & 50.02 & 50.00 & 54.6 & 53.40$\pm$0.45	& \textbf{61.08$\pm$0.06} &	55.12$\pm$0.06	& 48.56$\pm$0.11 & 50.09$\pm$0.03 & 59.29$\pm$0.48	 \\
  & & Time & 1.6 & 72.16 & 2.69 & 5.44$\pm$0.65 & 87.05$\pm$3.38 & 1110.78$\pm$38.39 & 1109.44$\pm$17.12 & 478.10$\pm$30.42 & 116.15$\pm$1.19 \\
  \cmidrule{2-12}  
  & \multirow{2}{*}{0.75} 
  & bAcc & 49.99 & 50	& 56.19	& 53.76$\pm$1.07 & 60.78$\pm$0.12	& 56.75$\pm$0.02 & 49.32$\pm$0.04 & 50.05$\pm$0.07 & \textbf{63.18$\pm$0.61}	\\
  & & Time & 1.78  & 51.8  & 2.16  & 3.20$\pm$0.00  & 129.95$\pm$23.78 & 1134.59$\pm$13.96 & 1259.35$\pm$21.46 & 693.89$\pm$19.41 & 122.99$\pm$2.65 \\
  \midrule

  \multirow{2}{*}{imdb\textsuperscript{$\dagger$}} &
  & bAcc & 81.56 & \textbf{82.18} & 80.08 & 50.14$\pm$0.02 & 77.42\textsuperscript{$\ddagger$} & 57.98$\pm$0.29 & 76.47$\pm$0.11 & 67.41\textsuperscript{$\ddagger$} & 73.10$\pm$0.19\\
  & & Time & 1257.01 & 4117.33 & 22.17 & 47818.07$\pm$2783.21 & 134556.64\textsuperscript{$\ddagger$} & 4072.86$\pm$138.29 & 2768.24$\pm$23.26 & 223699.04\textsuperscript{$\ddagger$} & 46437.42$\pm$18114.58\\
  \midrule

  \multirow{6}{*}{a8a} & 
  \multirow{2}{*}{0.25} 
  & bAcc & 50.01 & 50	& 60.67 & 54.75$\pm$0.87 & \textbf{61.58$\pm$0.00} & 50.01$\pm$0.03	& 49.99$\pm$0.00 & 50.00$\pm$0.00 & 50.00$\pm$0.01 \\
  & & Time & 22.94 & 25.7  & 4.96  & 19.17$\pm$0.02 & 2207.11$\pm$111.43 & 4394.51$\pm$72.96 & 207.23$\pm$4.25 & 5484.63$\pm$54.75 & 421.91$\pm$119.74 \\
  \cmidrule{2-12}
  & \multirow{2}{*}{0.5} 
  & bAcc & 50.01 & 50.00 & 66.46 & 64.04$\pm$1.01 & \textbf{68.57$\pm$0.00} & 50.11$\pm$0.01 & 50.01$\pm$0.00 & 50.00$\pm$0.00 & 55.33$\pm$1.99\\
  & & Time & 23.52 & 26.18 & 6.59 & 51.67$\pm$3.29 & 4120.49$\pm$415.07 & 4858.85$\pm$133.63 & 395.44$\pm$0.80 & 17639.23$\pm$2014.52 & 235.54$\pm$8.39\\
  \cmidrule{2-12}  
  & \multirow{2}{*}{0.75} 
  & bAcc & 50.01 & 50	& 70.6 & 68.81$\pm$1.10 & \textbf{72.70$\pm$0.00}	& 50.13$\pm$0.01 & 49.99$\pm$0.00	& 50.00$\pm$0.00 & 62.87$\pm$0.93	\\
  & & Time & 28.16 & 37.84  & 6.37  & 32.28$\pm$3.32 & 2929.01$\pm$1001.51 & 5263.24$\pm$92.75 & 569.20$\pm$11.54 & 28776.93$\pm$2045.19 & 577.31$\pm$768.95 \\
  \midrule

  \multicolumn{2}{c|}{\textit{\textbf{Win Count}} \textit{(2/2/2/1)}} & \textit{bAcc} & \textit{0/0/0/0} & \textit{0/0/0/1} & \textit{0/0/0/0} & \textit{0/0/0/0} & \textit{1/2/1/0} &  \textit{0/0/0/0} & \textit{0/0/0/0} & \textit{0/0/0/0} & \textit{1/0/1/0} \\
  \midrule

  \rowcolor{lightgray} \multicolumn{12}{c}{Large} \vspace{1mm}\\

  \multirow{2}{*}{diabetes\_us\textsuperscript{$\dagger$}} &
  & bAcc & 50	& 50.08 & 50.11 & 50.27$\pm$0.14	& \textbf{50.35$\pm$0.74} & 50.00$\pm$0.00 & 50.00$\pm$0.00	& 50.00$\pm$0.00 & 50.00$\pm$0.00\\
  & & Time & 41.55 & 32727.48 & 16.47 & 47.96$\pm$0.24 & 4251.94$\pm$210.06 & 487.49$\pm$1.34 & 491.84$\pm$11.12 & 25760.59$\pm$105.48 & 567.92$\pm$3.34\\
  \midrule

  \multirow{6}{*}{SUSY} & 
  \multirow{2}{*}{0.25} 
  & bAcc & 50 &   49.9    & 51.12	 & 52.11$\pm$0.19 & 59.72$\pm$0.00 & 54.68$\pm$0.01 & 49.37$\pm$0.01	& 50.53$\pm$1.17 & \textbf{61.98$\pm$0.10} \\
  & & Time & 68.97 &12100.93& 105.85 & 163.96$\pm$1.52& 3309.30$\pm$184.51 & 654852.7$\pm$8956.39 & 24252.71$\pm$37.64 & 16455.42$\pm$167.84 & 5773.05$\pm$129.01 \\
  \cmidrule{2-12}
  & \multirow{2}{*}{0.5} 
  & bAcc & 50	& 50.01 & 53.21	& 54.03$\pm$0.28 & 65.60$\pm$0.00 & 58.27$\pm$0.00 & 48.33$\pm$0.02 & 57.89$\pm$7.19 & \textbf{68.79$\pm$0.14} \\
  & & Time & 75.86 & 10731.91 & 135.96 & 255.58$\pm$0.50 & 4210.01$\pm$453.61 & 343724.64$\pm$5176.04 & 23928.85$\pm$169.87 & 19612.72$\pm$567.55 & 6054.62$\pm$660.84\\
  \cmidrule{2-12}  
  & \multirow{2}{*}{0.75} 
  & bAcc & 50 & 50.12	& 55.98	& 54.84$\pm$0.48 & 68.52$\pm$0.00	& 60.94$\pm$0.01 & 47.53$\pm$0.03 & 53.67$\pm$8.13 & \textbf{73.55$\pm$0.11}	\\
  & & Time & 83.59 & 12324.28 & 111.37 & 166.07$\pm$0.20 & 2201.85$\pm$12.91 & 114179.9$\pm$2433.89 & 25690.16$\pm$79.67 & 23987.22$\pm$1326.00 & 5787.78$\pm$123.95 \\
  \midrule

  \multirow{6}{*}{HIGGS} & 
  \multirow{2}{*}{0.25} 
  & bAcc & 50 &   50.16 & 50.57 & 49.97$\pm$0.07	& 50.97$\pm$0.00 & 50.18\textsuperscript{$\ddagger$} & 49.86$\pm$0.03 & 49.99$\pm$0.00 & \textbf{51.17$\pm$0.05} \\
  & & Time & 136.12 & 17777.05 & 112.34 & 201.20$\pm$12.85 & 8675.39$\pm$335.33 & 1845662.94\textsuperscript{$\ddagger$} & 42814.64$\pm$94.35 & 28363.61$\pm$153.32 & 5784.12$\pm$159.85 \\
  \cmidrule{2-12}
  & \multirow{2}{*}{0.5} 
  & bAcc & 50 & 50.01 & 51.21	& 50.06$\pm$0.06 & 51.83$\pm$0.00 & 50.21\textsuperscript{$\ddagger$} & 49.82$\pm$0.02 & 49.99$\pm$0.01	& \textbf{53.09$\pm$0.05} \\
  & & Time & 142.97 & 7885.78 & 145.83 & 267.35$\pm$0.54 & 10342.77$\pm$1686.50 & 1788308.41\textsuperscript{$\ddagger$} & 42623.97$\pm$97.07 & 44123.61$\pm$283.06 & 6039.45$\pm$565.24 \\
  \cmidrule{2-12}  
  & \multirow{2}{*}{0.75} 
  & bAcc & 50	& 50.55	& 51.98	& 49.97$\pm$0.05 & 52.66$\pm$0.00	& 50.16\textsuperscript{$\ddagger$} & 49.75$\pm$0.03 & 49.98\textsuperscript{$\ddagger$} & \textbf{55.55$\pm$0.11}	\\
  & & Time & 153.05 & 549606.27 & 125.45 &   200.58$\pm$2.95 & 7407.39$\pm$295.04 & 801655.55\textsuperscript{$\ddagger$} & 44079.06$\pm$134.05 & 65500.41\textsuperscript{$\ddagger$} & 5762.40$\pm$170.11 \\
  \midrule

  \multicolumn{2}{c|}{\textit{\textbf{Win Count}} \textit{(2/2/2/1)}} & \textit{bAcc} & \textit{0/0/0/0} & \textit{0/0/0/0} & \textit{0/0/0/0} & \textit{0/0/0/0} & \textit{0/0/0/1} &  \textit{0/0/0/0} & \textit{0/0/0/0} & \textit{0/0/0/0} & \textit{2/2/2/0} \\

  \bottomrule
  \end{tabular}  }
\end{table*}

%% file: Tables/Tab_Result_Inference.tex
\begin{table*}[!th]
  \caption{Model comparison based on Performance, Data Scalability, Prediction Consistency, Speed, and Feature Scalability.}
  \label{tab:results_inference}
  \centering
  \begin{tabular}{lrccccccccc}
  \toprule
  Metric & Groups & NB3 & FAE & OLVF & OCDS & OVFM & DynFo & ORF\textsuperscript{3}V & Aux-Net & Aux-Drop\\
  \midrule

  \multirow{4}{*}{Performance} 
  & Small & 53.95 & 54.47 & 61.16 & 56.49 & 66.80 & \textbf{68.30} & 49.67 & 56.25 & 59.33 \\
  & Medium & 54.52 & 54.60 & 63.12 & 56.68 & \textbf{65.41} & 53.26 & 53.18 & 52.52 & 59.97 \\
  & Large & 50.00 & 50.12 & 52.03 & 51.61 & 57.09 & 53.49 & 49.24 & 51.72 & \textbf{59.16}\\
   & \textit{\textbf{Average}} & 52.82  &  53.06  &  58.77  &  54.93  &  \textbf{63.10}  &  58.35  &  50.70  & 53.50 & 59.49\\
  \midrule

  \multirow{2}{*}{Data} & S $\rightarrow$ M & 1.06 & 0.24 & 3.20 & 0.34 & -2.08 & -22.02 & \textbf{7.07} & -6.63 & 1.08 \\
  \multirow{2}{*}{Scalability} & M $\rightarrow$ L & -8.29 & -8.21 & -17.57 & -8.94 & -12.72 & \textbf{0.43} & -7.41 & -1.52 & -1.35\\
   &  \textbf{\textit{Measure}} & 0.02  &  0.01  &  0.01  &  0.01  &  0.00  &  0.00  &  0.11  & 0.00 & \textbf{0.38}\\
  \midrule

  & Small & - & - & - & 1.19 & 1.45 & \textbf{0.53} & 0.65 & 1.14 & 1.32 \\
  Prediction & Medium & - & - & - & 0.66 & \textbf{0.03} & 0.10 & 0.07 & \textbf{0.03} & 0.68 \\
  Consistency & Large & - & - & - & 0.18 & 0.11 & \textbf{0.00} & 0.02 & 2.75 & 0.08 \\
   & \textbf{\textit{Average}} & - & - & - & 0.68  &  0.53  &  \textbf{0.21}  &  0.25  & 1.31 & 0.69\\
  \midrule

  & Small & 7.59  &  169.54  &  \textbf{0.32}  &  247.34  &  968.61  &  96.52  &  56.03  & 2278.47 & 522.94\\
  Time & Medium & 190.92  &  629.10  &  6.72  &  6847.73  &  20594.30 &  2983.06  &  1037.31  & 39592.68 & 6860.40 \\
  (= 1/Speed) & Large & \textbf{100.30}  &  91879.10  &  107.61  &  186.10  &  5771.24  & 792695.95 &  29125.89  & 31971.94 & 5109.91\\
   & \textbf{\textit{Average}} & 99.6  &  30892.58  &  \textbf{38.22}  &  2427.06  &  9111.38  &  265258.51  &  10073.08  & 24614.36 &   4164.42 \\
  \midrule

  Feature  & SUSY (8) & 75.86 & 10731.91 & 135.96 & 255.58 & 4210.01 & 343724.64 & 23928.85 & 19612.72 & 6054.62 \\
  Scalability & HIGGS (21) & 166.78 & 25253.06 & 146.08 & 292.24 & 10342.77 & 1788308.41 & 42623.97 & 44123.61 & 6039.45 \\
  (8 $\rightarrow$ 21) & \textbf{\textit{Ratio}} (2.625) & 2.20  &  2.35  &  1.07  &  1.14  & 2.46 &  5.20  &  1.78  & 2.25 & \textbf{1.00}\\
  

  \bottomrule
  \end{tabular}
\end{table*}

%% file: Tables/Tab_Carbon_Footprint.tex
\begin{table*}[!t]
  \caption{Carbon footprint (in kg) of each model in this survey. Time is in seconds and energy is in kwh.}
  \label{tab:Carbon_Footporint}
  \centering
  \begin{tabular}{c@{\hskip 4mm}cccccccccc} 
  \toprule
  
   & NB3 & FAE  & OLVF & OCDS & OVFM & DynFo & ORF\textsuperscript{3}V & Aux-Net & Aux-Drop & \textbf{Total}\\
  \midrule

  Time & 9579.05 & 952541.15 & 91699.98 & 537130.53 & 701807.78 & 7937630.13 & 1708852.33 & 1210730.09 & 551530.98 & \textbf{13701502.02}\\
  Carbon & 0.06	& 6.43	& 0.62	& 3.62	& 4.73	& 53.55	& 11.53	 & 8.17 & 3.72	& \textbf{92.43}\\
  Energy & 8.5 & 843.33 & 81.22 & 475.58 & 621.34 & 7030 & 1510	& 1070 & 488.33 & \textbf{12128.3} \\
  \bottomrule
  
  \end{tabular}
\end{table*}

%% file: FigureTex/Fig_CarbonEmission.tex
\begin{figure}[t]
    \begin{center}
    \includegraphics[width=.9\columnwidth, trim={0.2cm 0.3cm 0.3cm 0.2cm}, clip]{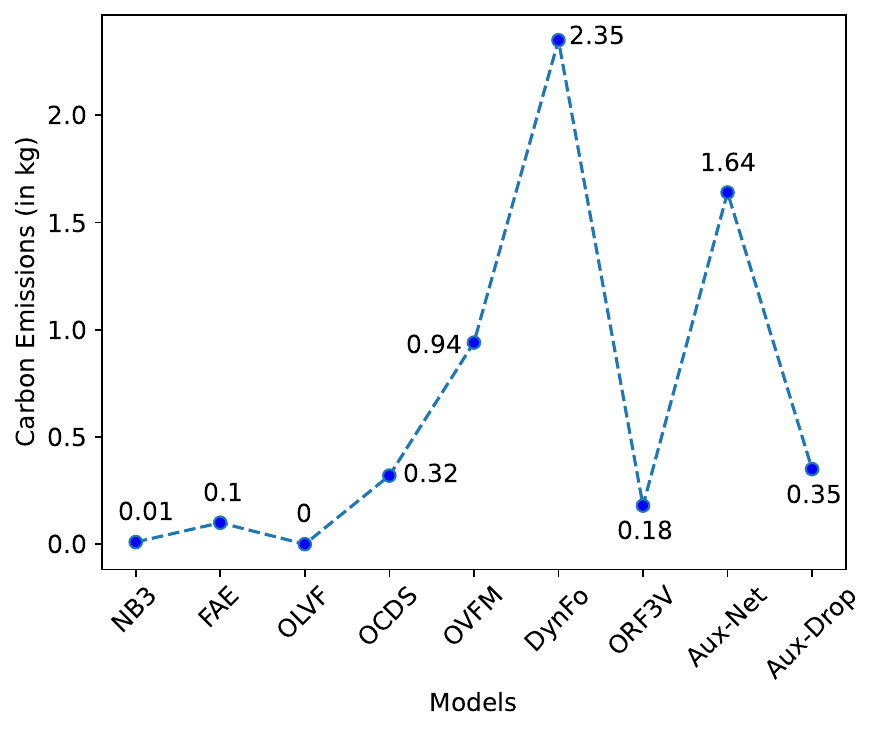} 
    \end{center}
    \caption{Comparison of models on carbon footprint: The sum of carbon emission on one run of the SUSY dataset at $p$ = 0.5 and the imdb dataset for each model is reported here.}
    \label{fig:carbon_emission}
\end{figure}

%% file: FigureTex/Fig_RealSynthetic.tex
\begin{figure}[t]
    \begin{center}
    \includegraphics[width=.9\columnwidth, trim={0.2cm 0.3cm 0.3cm 0.2cm}, clip]{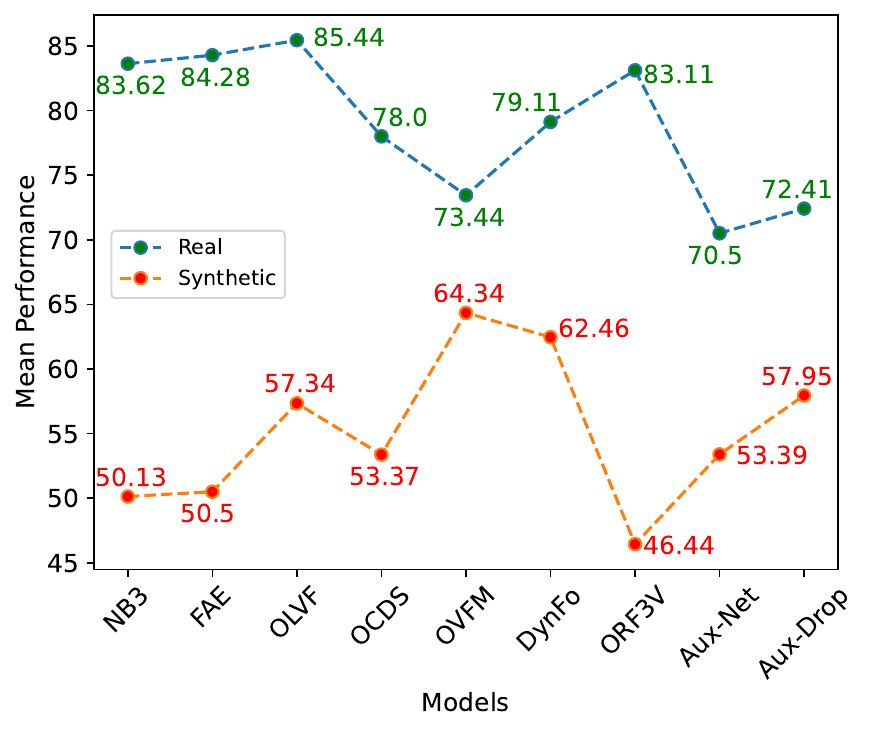} 
    \end{center}
    \caption{Mean model performance in real and synthetic datasets: The average of the mean balanced accuracy of all the real dataset and synthetic datasets considering all $p$ values are shown here.}
    \label{fig:real_vs_synthetic}
\end{figure}

%% file: FigureTex/Fig_AdaptedSeFT.tex
\begin{figure*}[th]
    \begin{center}
    \includegraphics[width=\textwidth, trim={1.5cm 0.0cm 0.0cm 0.0cm}, clip]{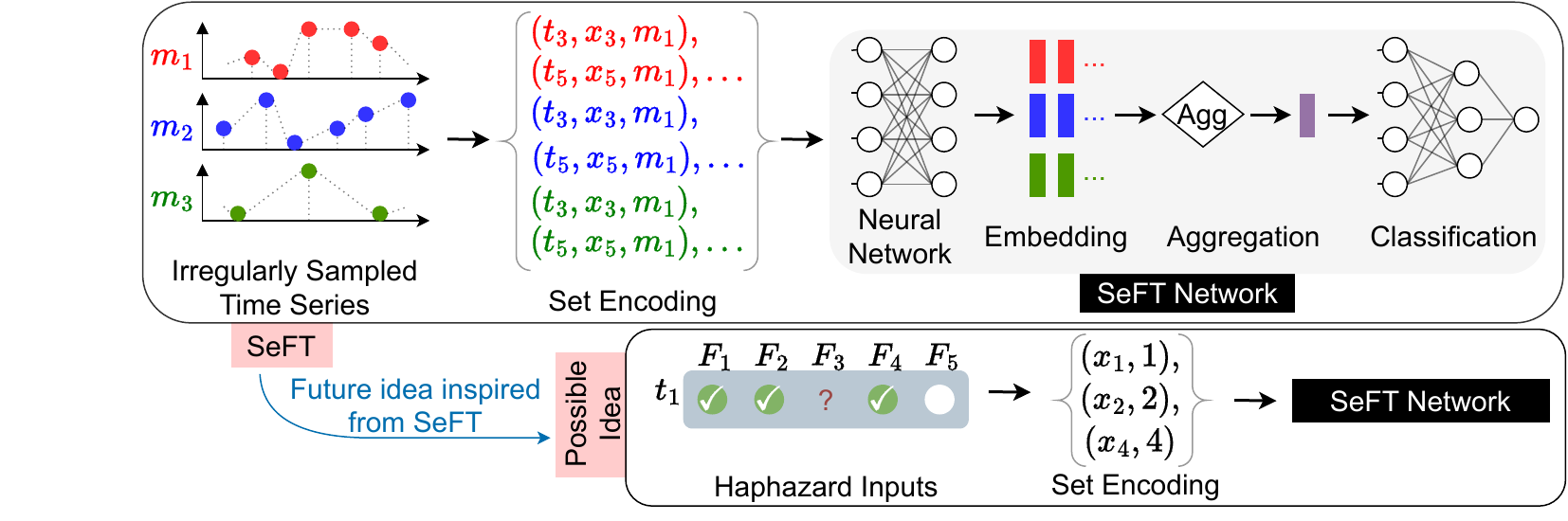} 
    \end{center}
    \caption{Set function for time series (SeFT) and the possible idea of adapting SeFT to handle haphazard inputs.}
    \label{fig:adaptedSeFT}
\end{figure*}

%% file: FigureTex/Fig_SubfieldHaphazard.tex
\begin{figure*}[t]
    \begin{center}
    \includegraphics[width=\textwidth, trim={0.0cm 0.0cm 0.0cm 0.0cm}, clip]{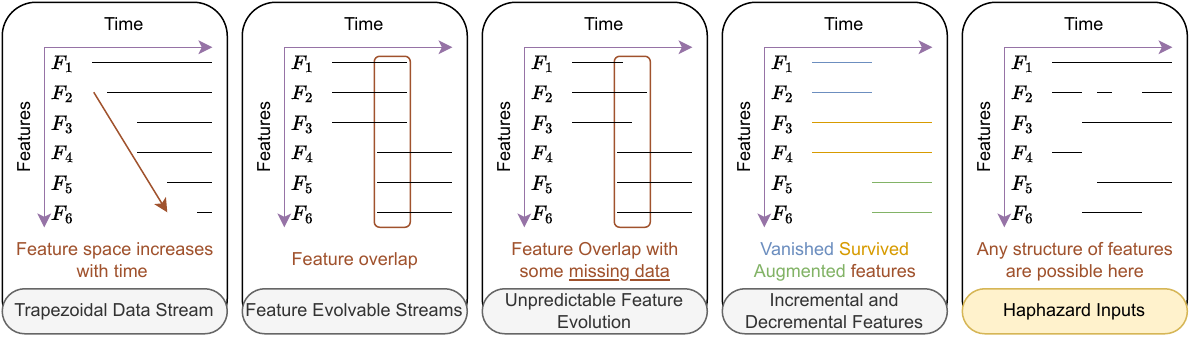} 
    \end{center}
    \caption{Feature space in the subfield of haphazard inputs. \textit{Note: The time is shown continuously here for ease of understanding. However, it can be discrete as well as continuous. In this review, the datasets are discrete.}}
    \label{fig:subfieldhaphazard}
\end{figure*}

%% file: Tables/Tab_Result_Synthetic_p25_small.tex
\begin{table*}[!th]
  \caption{Comparison of models on Small Synthetic Datasets (\underline{$p$ = 0.25 }): The deterministic models --- NB3, FAE, and OLVF --- underwent a single execution. In contrast, the non-deterministic models were executed 5 times, with the mean $\pm$ standard deviation reported. Here, Err, Acc, ROC, PRC, bAcc, and Time stand for the number of errors, accuracy, AUROC, AUPRC, balanced accuracy, and execution time, respectively.}
  
  \label{tab:results_synthetic_p25_small}
  \centering
  \resizebox{\linewidth}{4.4in}{%
  \begin{tabular}{lrccccccccc}
  \toprule
  Dataset & Metric & NB3 & FAE & OLVF & OCDS & OVFM & DynFo & ORF\textsuperscript{3}V & Aux-Net & Aux-Drop\\
  \midrule

  \multirow{6}{*}{WPBC} & Err &  53 & 79 & 58	& 51.80$\pm$1.92 & 72.80$\pm$2.17	& 49.40$\pm$2.07 & 46.20$\pm$0.45 & 47.80$\pm$2.59 	& 54.40$\pm$5.59 \\
  & Acc & 73.23 & 60.1 & 70.71 & 73.84$\pm$0.97 & 63.23$\pm$1.09 & 74.92$\pm$1.05 & 76.55$\pm$0.23 & 75.86$\pm$1.31  & 72.53$\pm$2.83 \\
  & ROC &	 54.33 & 54.8 &50.26 & 52.28$\pm$1.74 & 49.11$\pm$1.36 & 44.55$\pm$1.13 & 49.65$\pm$3.59 & 48.29$\pm$1.69  & 51.96$\pm$1.76 \\
  & PRC &	29.13 & 24.36 & 23.65 & 24.49$\pm$2.40 & 24.25$\pm$0.79 & 21.41$\pm$0.35 & 21.94$\pm$1.60 & 28.59$\pm$2.81  & 23.88$\pm$1.78 \\
  & bAcc & 49.48 & 48.93 & 58.08 & 49.88$\pm$1.24 & 46.73$\pm$1.01 & 56.58$\pm$1.40 & 50.54$\pm$0.49 & 52.08$\pm$0.95  & 49.89$\pm$2.45 \\
  & Time & 0.05 & 0.08 & 0.02  & 0.04$\pm$0.00  & 2.21$\pm$0.01  & 2.57$\pm$0.01 & 0.35$\pm$0.00 & 7.86$\pm$0.38  & 1.07$\pm$0.04 \\
  \midrule
  
  \multirow{6}{*}{ionosphere} & Err &  152 & 127	& 118 & 119.20$\pm$2.86 & 130.60$\pm$1.95	& 113.20$\pm$2.95 & 193.20$\pm$6.22	& 127.40$\pm$1.67  & 126.40$\pm$0.55 \\
  & Acc & 56.7 & 63.82 & 66.38 & 66.04$\pm$0.82 & 62.79$\pm$0.56 & 67.66$\pm$0.84 & 44.80$\pm$1.78 & 63.70$\pm$0.48  & 63.99$\pm$0.16 \\
  & ROC &	48.37 & 50.67 & 51.56 & 63.23$\pm$0.63 & 58.00$\pm$0.28 & 54.80$\pm$0.79	& 44.25$\pm$1.88 & 46.99$\pm$0.71  & 48.30$\pm$1.18 \\
  & PRC &	67.33 & 65.76 & 64.08 & 70.77$\pm$0.67 & 66.35$\pm$0.23 & 68.47$\pm$0.47	& 61.03$\pm$1.01 & 64.89$\pm$1.80  & 61.69$\pm$0.83 \\
  & bAcc & 45.44	& 49.78 & 59.98 & 58.88$\pm$1.05 & 56.10$\pm$0.74 & 57.30$\pm$0.96 & 42.26$\pm$1.83 & 49.86$\pm$0.26  & 49.91$\pm$0.12	\\
  & Time & 0.08 & 0.12 & 0.04 & 0.08$\pm$0.00 & 4.10$\pm$0.06 & 4.56$\pm$0.11 & 0.39$\pm$0.01 & 14.45$\pm$0.27  & 1.86$\pm$0.00 \\
  \midrule
  
  \multirow{6}{*}{WDBC} & Err & 219 & 267	& 247 & 267.60$\pm$2.61 & 103.40$\pm$5.86 & 92.60$\pm$7.13 & 316.60$\pm$10.71 & 277.80$\pm$72.77 	& 221.20$\pm$6.83 \\
  & Acc & 61.51 & 53.08 & 56.59 & 52.97$\pm$0.46 & 81.83$\pm$1.03 & 83.70$\pm$1.25	& 44.26$\pm$1.89 & 51.18$\pm$12.79 	& 61.12$\pm$1.20 \\
  & ROC &	49.85 & 58.66 & 64.68 & 48.25$\pm$1.26	& 88.31$\pm$0.73 & 42.12$\pm$0.36	& 34.78$\pm$1.16 & 49.96$\pm$2.49 	& 45.88$\pm$1.40	\\
  & PRC &	36.08 & 45.14 & 60.36 & 41.63$\pm$1.64	& 83.50$\pm$0.83 & 32.63$\pm$0.37	& 28.65$\pm$0.47 & 37.69$\pm$2.37 	& 34.81$\pm$0.71	\\
  & bAcc & 49.4 & 48.72 & 49.41 & 49.01$\pm$0.78	& 80.35$\pm$1.05 & 83.00$\pm$1.45	& 41.32$\pm$1.76 & 49.58$\pm$0.35 	& 49.48$\pm$0.41	\\
  & Time & 0.11 & 0.21 & 0.07 & 0.12$\pm$0.00  & 5.73$\pm$0.13  & 7.36$\pm$0.12 & 1.07$\pm$0.07 & 20.90$\pm$0.90  & 3.08$\pm$0.08\\
  \midrule
  
  \multirow{6}{*}{australian} & Err &  344 & 284 & 314 & 368.60$\pm$3.51	& 233.00$\pm$2.92 & 323.80$\pm$4.15 & 387.00$\pm$5.43 & 313.20$\pm$3.27  & 320.80$\pm$18.47	\\
  & Acc &	50.14 & 58.84 & 54.49 & 46.58$\pm$0.51 & 66.23$\pm$0.42 & 53.00$\pm$0.60	& 43.83$\pm$0.79 & 54.61$\pm$0.47 	& 53.51$\pm$2.68	\\
  & ROC &	50.72 & 58.18 & 60.3  & 52.62$\pm$0.95	& 73.24$\pm$0.15 & 53.28$\pm$0.59	& 49.23$\pm$0.23 & 56.50$\pm$1.54  & 50.59$\pm$3.71	\\
  & PRC &	46.25 & 56.5 & 55.66 & 50.19$\pm$1.18 & 67.09$\pm$0.30 & 48.05$\pm$0.36	& 44.71$\pm$0.23 & 47.19$\pm$0.66  & 47.15$\pm$2.48	\\
  & bAcc & 49.89 & 59.69 & 53.09 & 49.79$\pm$0.40	& 65.37$\pm$0.42 & 50.06$\pm$0.42	& 45.11$\pm$0.86 & 49.88$\pm$0.36  & 51.51$\pm$2.00	\\
  & Time & 0.07 & 0.34 &  0.08  & 0.12$\pm$0.00  & 5.07$\pm$0.07  & 0.97$\pm$0.02 & 0.46$\pm$0.00 & 15.31$\pm$0.27  & 3.63$\pm$0.01 \\
  \midrule
  
  \multirow{6}{*}{wbc} & Err & 244 & 317 & 363 & 383.60$\pm$38.58 & 110.00$\pm$0.00 & 168.60$\pm$8.68 & 527.20$\pm$5.63 & 242.80$\pm$2.49  & 125.20$\pm$6.46	\\
  & Acc & 65.09 & 54.65 & 48.07 & 45.12$\pm$5.52	& 84.26$\pm$0.00 & 75.85$\pm$1.24	& 24.47$\pm$0.81 & 65.26$\pm$0.36 	& 82.09$\pm$0.92	\\
  & ROC &	49.08 & 70.09 & 80.25 & 62.71$\pm$13.73 & 89.65$\pm$0.00 & 51.57$\pm$1.49 & 41.21$\pm$0.91 & 54.91$\pm$3.66  & 85.02$\pm$0.58	\\
  & PRC & 34.81 & 63.19 & 74.48 & 58.40$\pm$10.89 & 85.73$\pm$0.00 & 35.67$\pm$0.71 & 28.99$\pm$0.65 & 36.54$\pm$1.95  & 80.57$\pm$1.02	\\
  & bAcc & 50.66 & 53.4 & 52.02 & 50.71$\pm$2.12	& 81.70$\pm$0.00 & 67.20$\pm$1.92 & 29.38$\pm$1.13 & 49.90$\pm$0.24  & 77.07$\pm$1.32	\\
  & Time & 0.05 &  0.38 & 0.07  & 0.12$\pm$0.00   & 4.20$\pm$0.07  & 1.04$\pm$0.03 & 0.23$\pm$0.00 & 13.36$\pm$0.21  & 3.76$\pm$0.08 \\
  \midrule
  
  \multirow{6}{*}{diabetes-f} & Err & 294	& 382 & 273 & 404.80$\pm$10.38	& 260.00$\pm$0.00 & 311.60$\pm$7.09 & 387.20$\pm$4.32 & 268.40$\pm$0.55  & 339.00$\pm$64.72 \\
  & Acc & 61.72 & 50.26 & 64.45 & 47.29$\pm$1.35	& 66.15$\pm$0.00 & 59.37$\pm$0.92	& 49.52$\pm$0.56 & 65.05$\pm$0.07  & 55.86$\pm$8.43	\\
  & ROC &	49.76 & 52.98 & 55.49 & 53.29$\pm$0.60	& 62.53$\pm$0.00 & 49.18$\pm$0.45	& 50.62$\pm$0.87 & 51.00$\pm$2.04 	& 49.15$\pm$1.22	\\
  & PRC &	34.62 & 38.29 & 39.27 & 41.06$\pm$0.72	& 50.05$\pm$0.00 & 35.52$\pm$0.21	& 35.04$\pm$0.49 & 36.45$\pm$1.88 	& 35.56$\pm$1.10	\\
  & bAcc & 48.7 & 45.53 & 49.5  & 51.05$\pm$1.08	& 58.24$\pm$0.00 & 56.03$\pm$0.60	& 45.74$\pm$0.31 & 49.96$\pm$0.05 	& 49.51$\pm$0.70	\\
  & Time & 0.06 & 0.48  & 0.11  & 0.17$\pm$0.00 & 2.44$\pm$0.07 & 1.67$\pm$0.25 & 0.36$\pm$0.00 & 14.26$\pm$0.30  & 4.09$\pm$0.04 \\
  \midrule
  
  \multirow{6}{*}{german} & Err & 289	& 279 & 503 & 563.80$\pm$11.01	& 302.00$\pm$0.00 & 304.80$\pm$2.39 & 288.80$\pm$1.30 & 278.00$\pm$4.12 	& 396.20$\pm$80.79	\\
  & Acc & 69.86 &  70.91 &  47.55 & 41.21$\pm$1.15 & 68.51$\pm$0.00	& 68.18$\pm$0.25 & 69.85$\pm$0.14	& 71.01$\pm$0.43  & 58.69$\pm$8.42	\\
  & ROC &	48.43 &  52.93 &  49.51	& 49.88$\pm$1.35 & 55.96$\pm$0.00	& 47.37$\pm$0.48 & 46.08$\pm$0.56	& 54.18$\pm$0.28  & 50.82$\pm$1.61	\\
  & PRC &	68.73 &  72.56 &  71.93	& 71.08$\pm$0.85 & 73.57$\pm$0.00 & 70.16$\pm$0.38 & 68.79$\pm$0.34	& 73.93$\pm$0.15  & 71.53$\pm$1.10	\\
  & bAcc & 51.37 & 49.82 &  52.58	& 49.96$\pm$1.11 & 52.38$\pm$0.00 & 50.58$\pm$0.27 & 49.68$\pm$0.21	& 50.32$\pm$0.45  & 50.14$\pm$0.67	\\
  & Time & 0.16 &  0.3   &  0.15  & 0.19$\pm$0.00  & 13.17$\pm$0.06 & 29.36$\pm$0.48 & 0.79$\pm$0.00 & 30.12$\pm$0.76  & 5.15$\pm$0.09 \\
  \midrule
  
  \multirow{6}{*}{IPD} & Err & 550 & 548 & 384 & 459.60$\pm$12.22 & 263.00$\pm$5.00 & 267.60$\pm$8.17 & 569.00$\pm$3.39	& 436.20$\pm$11.08 	&  431.60$\pm$8.44 \\
  & Acc & 49.82 & 50     & 64.96 & 58.07$\pm$1.11 & 76.00$\pm$0.46 & 75.56$\pm$0.75 & 48.04$\pm$0.31	& 60.20$\pm$1.01  & 60.62$\pm$0.77	\\
  & ROC &	50.8 & 49.4	   & 53.14 & 61.45$\pm$1.11 & 84.39$\pm$0.39 & 48.83$\pm$0.66 & 40.96$\pm$0.62	& 65.73$\pm$1.41  & 65.02$\pm$1.28	\\
  & PRC &	51.39 & 49.98  & 52.66 & 59.47$\pm$0.78 & 82.57$\pm$0.37 & 49.38$\pm$0.50 & 45.50$\pm$0.33	& 63.36$\pm$1.90  & 63.67$\pm$1.86	\\
  & bAcc & 49.82 & 50.09 & 64.97 & 58.07$\pm$1.11 & 76.00$\pm$0.46 & 75.57$\pm$0.75 & 47.93$\pm$0.31	& 60.24$\pm$1.00  & 60.63$\pm$0.77	\\
  & Time & 0.18 & 1.14   & 0.15  & 0.22$\pm$0.00  & 9.93$\pm$0.59  & 14.25$\pm$0.07 & 8.25$\pm$0.09 & 33.88$\pm$0.77  & 5.85$\pm$0.08 \\
  \midrule
  
  \multirow{6}{*}{svmguide3} & Err &  303 & 297 & 351 & 357.00$\pm$4.36 & 381.20$\pm$4.76 & 335.00$\pm$2.00 & 338.60$\pm$2.97	& 296.20$\pm$0.45 	& 319.80$\pm$18.66	\\
  & Acc &	75.62 & 76.11  &  71.76	& 71.28$\pm$0.35 & 69.33$\pm$0.38	& 73.03$\pm$0.16 & 72.74$\pm$0.24	& 76.17$\pm$0.04  & 74.27$\pm$1.50	\\
  & ROC & 48.29 & 52.84  &  53.45	& 50.08$\pm$0.53 & 56.97$\pm$0.53 & 50.13$\pm$0.35 & 48.06$\pm$0.14	& 49.70$\pm$0.70  & 48.54$\pm$0.65	\\
  & PRC &	75.05 & 77.16  &  77.23	& 75.74$\pm$0.24 & 79.00$\pm$0.16 & 76.81$\pm$0.18 & 75.08$\pm$0.12	& 76.73$\pm$1.01  & 74.97$\pm$0.70	\\
  & bAcc & 49.75 & 49.95 &  52.09	& 50.84$\pm$0.37 & 54.26$\pm$0.72 & 50.79$\pm$0.33 & 49.54$\pm$0.43	& 49.99$\pm$0.02  & 50.21$\pm$0.40	\\
  & Time & 0.17 & 0.33   &  0.14  & 0.24$\pm$0.00  & 11.71$\pm$0.58 & 37.24$\pm$0.20 & 4.17$\pm$0.06  & 35.78$\pm$1.10  & 6.90$\pm$0.36 \\
  \midrule
  
  \multirow{6}{*}{kr-vs-kp} & Err & 1563 & 1540 & 1513 & 1586.80$\pm$18.07 & 1203.00$\pm$0.00 & 1436.20$\pm$20.52 & 1802.60$\pm$6.07	& 1522.60$\pm$25.05  & 1466.40$\pm$11.30 \\
  & Acc &	51.1 &   51.81 & 52.66 & 50.35$\pm$0.57 & 62.36$\pm$0.00 & 55.05$\pm$0.64 & 43.58$\pm$0.19	& 52.36$\pm$0.78  & 54.12$\pm$0.35	\\
  & ROC &	49.13 &  47.36 & 47.82 & 50.81$\pm$0.47 & 66.03$\pm$0.00 & 50.58$\pm$0.17 & 47.07$\pm$0.14	& 51.99$\pm$1.61  & 54.47$\pm$0.18	\\
  & PRC &	51.6 &   49.91 & 50.5  & 52.65$\pm$0.32 & 65.06$\pm$0.00 & 52.41$\pm$0.20 & 49.60$\pm$0.06	& 53.63$\pm$1.08  & 55.91$\pm$0.37	\\
  & bAcc & 50.35 & 49.68 & 52.47 & 50.21$\pm$0.52 & 62.13$\pm$0.00 & 54.42$\pm$0.66 & 44.16$\pm$0.18	& 50.83$\pm$0.76  & 53.12$\pm$0.38	\\
  & Time & 0.79 &  12.26 & 0.38  & 0.70$\pm$0.00  & 105.13$\pm$13.74 & 72.43$\pm$3.16 & 5.27$\pm$0.09 & 127.93$\pm$3.05  & 17.35$\pm$0.17 \\
  \midrule
  
  \multirow{6}{*}{spambase} & Err & 1834 & 1829 & 1516 & 2534.20$\pm$1.30 & 1530.00$\pm$8.51 & 1410.80$\pm$11.30 & 1901.20$\pm$5.54	& 1820.80$\pm$2.05 	& 1734.00$\pm$71.99 \\
  & Acc &	60.14 & 60.25 & 67.05 & 44.92$\pm$0.03	& 66.75$\pm$0.19 & 69.33$\pm$0.25	& 58.67$\pm$0.12 & 60.43$\pm$0.04 	& 62.31$\pm$1.56 \\
  & ROC &	47.58 & 50.56 & 58.47 & 58.48$\pm$0.04	& 71.54$\pm$0.30 & 47.35$\pm$0.33	& 46.70$\pm$0.26 & 66.94$\pm$8.02 	& 57.14$\pm$1.21 \\
  & PRC &	39.59 & 40.57 & 48.46 & 50.21$\pm$0.07	& 60.88$\pm$0.41 & 37.12$\pm$0.29	& 37.04$\pm$0.25 & 50.55$\pm$5.26 	& 50.16$\pm$1.29 \\
  & bAcc & 50.54 &51.17 & 64.34 & 53.63$\pm$0.04	& 64.26$\pm$0.17 & 66.05$\pm$0.25	& 48.94$\pm$0.09 & 49.92$\pm$0.06 	& 55.57$\pm$0.99 \\
  & Time & 1.56 & 25.36 &  0.56 & 1.68$\pm$0.00 & 134.27$\pm$14.13 & 138.10$\pm$0.30 & 19.54$\pm$0.14 & 280.91$\pm$1.43  & 26.58$\pm$1.28 \\

\bottomrule
  \end{tabular}
  }
\end{table*}

%% file: Tables/Tab_Result_Synthetic_p50_small.tex
\begin{table*}[!th]
  \caption{Comparison of models on Small Synthetic Datasets (\underline{$p$ = 0.5}): The deterministic models --- NB3, FAE, and OLVF --- underwent a single execution. In contrast, the non-deterministic models were executed 5 times, with the mean $\pm$ standard deviation reported. Here, Err, Acc, ROC, PRC, bAcc, and Time stand for the number of errors, accuracy, AUROC, AUPRC, balanced accuracy, and execution time, respectively.}
  \label{tab:results_synthetic_p50_small}
  \centering
  \resizebox{\linewidth}{4.3in}{%
  \begin{tabular}{lrccccccccc}
  \toprule
  Dataset & Metric & NB3 & FAE & OLVF & OCDS & OVFM & DynFo & ORF\textsuperscript{3}V & Aux-Net & Aux-Drop\\
  \midrule

  \multirow{6}{*}{WPBC} & Err & 48 & 61 & 60 & 48.00$\pm$0.00 & 61.60$\pm$2.07 & 50.80$\pm$0.84 & 45.00$\pm$0.00 & 49.00$\pm$1.22 & 52.60$\pm$3.78 \\
  & Acc & 75.76 &	69.19 & 69.7 & 75.76$\pm$0.00 & 68.89$\pm$1.05 & 74.21$\pm$0.42 & 77.16$\pm$0.00 & 75.25$\pm$0.62 & 73.43$\pm$1.91 \\
  & ROC &	50.11 & 44.19 & 44.24 & 46.89$\pm$0.26 & 51.37$\pm$0.72	& 39.39$\pm$1.06 & 42.04$\pm$3.46	& 45.62$\pm$2.16 & 49.36$\pm$2.19 \\
  & PRC &	24.36 &	22.82 & 23.06 & 21.95$\pm$0.04 & 24.03$\pm$0.32	& 19.38$\pm$0.38 & 20.80$\pm$1.05	& 24.62$\pm$1.41 & 24.17$\pm$1.77	\\
  & bAcc & 49.67 & 51.96 & 53.76 & 49.67$\pm$0.00 & 51.32$\pm$1.04	& 51.59$\pm$0.54 & 52.60$\pm$0.00	& 50.66$\pm$0.77 & 50.78$\pm$2.39	\\
  & Time & 0.05 & 0.08 & 0.03 & 0.06$\pm$0.00 & 2.11$\pm$0.13 & 3.46$\pm$0.09 & 0.66$\pm$0.00 & 15.81$\pm$0.78 & 1.27$\pm$0.07\\
  \midrule
  
  \multirow{6}{*}{ionosphere } & Err & 131 & 127 & 98 & 111.00$\pm$7.38 & 113.40$\pm$1.52 & 95.80$\pm$2.59 & 181.20$\pm$5.22 & 127.40$\pm$1.34 &  115.80$\pm$3.96 \\
  & Acc & 62.68 &	63.82 & 72.08 & 68.38$\pm$2.10	& 67.69$\pm$0.43 & 72.63$\pm$0.74	& 48.23$\pm$1.49 & 63.70$\pm$0.38 &  67.01$\pm$1.13	\\
  & ROC &	57.62 &	46.64 &	55.13 & 66.99$\pm$2.06	& 61.60$\pm$0.33 & 59.80$\pm$0.75	& 43.65$\pm$0.72 & 46.88$\pm$1.61 & 64.31$\pm$1.47	\\
  & PRC &	72.48 &	64.02 & 64.39 & 73.26$\pm$1.81	& 66.90$\pm$0.20 & 73.05$\pm$0.56	& 60.88$\pm$0.55 & 65.12$\pm$1.97 &  73.69$\pm$1.47	\\
  & bAcc & 49.41 & 49.78 & 66	& 60.42$\pm$1.97 & 61.15$\pm$0.42 & 63.37$\pm$1.03 & 43.51$\pm$1.75	& 49.93$\pm$0.24 &  54.92$\pm$1.69	\\
  & Time & 0.08 & 0.12 & 0.06 & 0.11$\pm$0.00 & 6.35$\pm$0.96 & 6.09$\pm$0.13 & 0.75$\pm$0.01 & 26.51$\pm$1.53 &  1.97$\pm$0.02\\
  \midrule
  
  \multirow{6}{*}{WDBC} & Err & 216 & 277 & 223 & 236.60$\pm$4.56 & 67.80$\pm$2.77 & 53.40$\pm$3.13 & 312.80$\pm$6.06	& 272.60$\pm$58.08 & 214.80$\pm$9.50	\\
  & Acc & 62.04 &	51.32 & 60.81 & 58.42$\pm$0.80	& 88.08$\pm$0.49 & 90.60$\pm$0.55	& 44.93$\pm$1.07 & 52.09$\pm$10.21 &  62.25$\pm$1.67	\\
  & ROC &	48.93 & 57.37 & 68.42 & 44.87$\pm$0.88	& 95.02$\pm$0.20 & 39.15$\pm$0.74 & 37.63$\pm$1.13 & 52.02$\pm$2.14 &  48.98$\pm$1.31	\\
  & PRC &	36.66 &	46.58 & 60.93 & 39.84$\pm$1.09	& 90.16$\pm$0.42 & 30.06$\pm$0.46 & 31.52$\pm$0.57 & 37.71$\pm$1.47 &  37.32$\pm$1.72	\\
  & bAcc & 50.01 & 49.23 & 53.25 & 52.40$\pm$0.75 & 87.78$\pm$0.53 & 90.47$\pm$0.51 & 41.32$\pm$1.21 & 50.52$\pm$1.26 &  50.53$\pm$1.72	\\
  & Time & 0.12 & 0.3 & 0.08 & 0.16$\pm$0.00 & 3.90$\pm$0.01 & 9.19$\pm$0.26 & 2.15$\pm$0.09 & 36.17$\pm$0.30 &  3.08$\pm$0.02\\
  \midrule

  \multirow{6}{*}{australian} & Err & 307 & 282 & 293	& 361.20$\pm$10.71 & 170.00$\pm$3.54 & 164.80$\pm$2.17 & 396.20$\pm$3.27 & 303.80$\pm$7.19 & 294.00$\pm$11.58 \\
  & Acc &	55.51 & 59.13 & 57.54 & 47.65$\pm$1.55	& 75.36$\pm$0.51 &	76.08$\pm$0.31	& 42.50$\pm$0.47 & 55.97$\pm$1.04 & 57.39$\pm$1.68	\\
  & ROC &	54.13 &	57.18 & 58.47 & 53.95$\pm$0.74	& 81.31$\pm$0.24 & 49.49$\pm$0.29 & 44.96$\pm$0.92 & 63.62$\pm$3.73 & 58.22$\pm$3.44	\\
  & PRC &	48.11 & 55.60 & 57.01 & 51.46$\pm$1.04	& 74.55$\pm$0.14 & 42.62$\pm$0.25 & 42.26$\pm$0.81 & 54.56$\pm$4.37 & 54.83$\pm$3.37	\\
  & bAcc & 53.62 & 57.01 & 56.22 & 50.76$\pm$1.18 & 74.96$\pm$0.53 & 75.88$\pm$0.23 & 44.11$\pm$0.65 & 50.61$\pm$1.32 & 55.77$\pm$1.88	 \\
  & Time & 0.08 & 0.53 & 0.1 & 0.17$\pm$0.00 & 6.67$\pm$0.89 & 10.68$\pm$0.23 & 0.82$\pm$0.01 & 20.36$\pm$0.12 & 4.33$\pm$0.11 \\
  \midrule
  
  \multirow{6}{*}{wbc} & Err & 243 & 324 & 298 & 342.60$\pm$44.20 & 78.00$\pm$0.00 & 45.60$\pm$1.52 & 517.80$\pm$13.46 & 203.20$\pm$37.89	&  105.00$\pm$5.96  \\
  & Acc & 65.24 & 53.65 & 57.37 & 50.99$\pm$6.32	& 88.84$\pm$0.00 & 93.47$\pm$0.22 & 25.82$\pm$1.93 & 70.93$\pm$5.42 & 84.98$\pm$0.85 \\
  & ROC &	50.09 & 81.31 & 76.92 & 51.17$\pm$11.69 & 94.86$\pm$0.00 & 44.72$\pm$0.45 & 38.69$\pm$1.95 & 74.30$\pm$13.16 & 89.30$\pm$1.55	\\
  & PRC & 33.40 &	76.97 & 70.42 & 48.67$\pm$10.00 & 91.97$\pm$0.00 & 30.96$\pm$0.26 & 27.93$\pm$0.82 & 60.41$\pm$20.96 & 85.95$\pm$1.74	\\
  & bAcc & 49.78 & 52.64 & 55.38 & 51.06$\pm$0.70 & 87.75$\pm$0.00 & 93.05$\pm$0.26 & 31.82$\pm$2.22 & 58.49$\pm$8.23 & 82.25$\pm$1.72	\\
  & Time & 0.05 & 0.40 & 0.09 & 0.16$\pm$0.00 & 4.26$\pm$0.04 & 28.10$\pm$0.53 & 0.37$\pm$0.00 & 16.21$\pm$0.28 &  4.23$\pm$0.13 \\
  \midrule
  
  \multirow{6}{*}{diabetes-f} & Err & 272 & 346 & 267	& 317.60$\pm$18.68 & 266.00$\pm$0.00 & 317.40$\pm$2.61 &	315.20$\pm$10.33 & 272.60$\pm$8.65 & 311.40$\pm$51.58 \\
  & Acc & 64.58 & 54.95 & 65.23 & 58.65$\pm$2.43	& 65.36$\pm$0.00 &	58.62$\pm$0.34 & 58.90$\pm$1.35 & 64.51$\pm$1.13	& 59.45$\pm$6.72	\\
  & ROC &	46.56 & 49.50 & 57.39 & 49.03$\pm$1.29	& 66.24$\pm$0.00 &	51.78$\pm$0.05	& 50.24$\pm$0.81 & 51.30$\pm$2.21	& 50.44$\pm$0.65	\\
  & PRC &	33.63 &	36.92 & 42.4 & 37.23$\pm$0.99 & 50.68$\pm$0.00 & 38.95$\pm$0.06	& 35.04$\pm$0.50 & 35.77$\pm$1.84	& 36.01$\pm$1.74	\\
  & bAcc & 50.73 & 47.05 & 50.19 & 50.79$\pm$0.73 & 59.20$\pm$0.00 & 53.82$\pm$0.42 & 49.16$\pm$1.16 & 50.08$\pm$0.35 & 51.20$\pm$0.90	  \\
  & Time & 0.06 & 0.67 & 0.13 & 0.18$\pm$0.00 & 1.82$\pm$0.02 & 10.48$\pm$0.28 & 0.58$\pm$0.01 & 17.36$\pm$0.17 & 5.41$\pm$0.28 \\
  \midrule
  
  \multirow{6}{*}{german} & Err & 304 & 277 & 454	& 436.80$\pm$14.91 & 303.00$\pm$0.00 & 279.00$\pm$2.45 & 276.00$\pm$1.22 & 276.00$\pm$1.73	& 345.80$\pm$42.20 \\
  & Acc & 68.30 & 71.12 & 52.66 & 54.45$\pm$1.55	& 68.40$\pm$0.00 &	70.88$\pm$0.26	& 71.19$\pm$0.13 & 71.22$\pm$0.18	& 63.94$\pm$4.40 \\
  & ROC &	49.92 & 55.46 & 50.53 & 51.26$\pm$1.54	& 58.68$\pm$0.00 &	50.04$\pm$0.25	& 47.13$\pm$0.58 & 54.69$\pm$0.34	& 52.00$\pm$0.36 \\
  & PRC &	70.62 & 74.55 & 72.16 & 71.85$\pm$0.93	& 74.91$\pm$0.00 &	70.39$\pm$0.07	& 69.19$\pm$0.37 & 75.26$\pm$0.12	& 72.41$\pm$0.58 \\
  & bAcc & 50.71 & 49.85 & 53.11 & 51.11$\pm$1.84 & 57.09$\pm$0.00 & 51.66$\pm$0.23 & 50.12$\pm$0.10 & 50.14$\pm$0.36 & 51.48$\pm$1.07 \\
  & Time & 0.17 & 0.35 & 0.18 & 0.26$\pm$0.00 & 17.07$\pm$0.41 & 34.33$\pm$0.87 & 1.34$\pm$0.07 & 166.09$\pm$20.93 & 6.14$\pm$0.16 \\
  \midrule
  
  \multirow{6}{*}{IPD} & Err & 534 & 548 & 237 & 361.20$\pm$9.78 & 130.40$\pm$3.44 & 166.20$\pm$5.89 & 555.80$\pm$2.77 & 301.60$\pm$25.56	& 325.40$\pm$14.12 \\
  & Acc & 51.28 & 50.00 & 78.38 & 67.04$\pm$0.89	& 88.10$\pm$0.31 &	84.82$\pm$0.54	& 49.24$\pm$0.25 & 72.48$\pm$2.33	& 70.31$\pm$1.29	\\
  & ROC &	51.61 &	50.59  & 50.3 & 74.82$\pm$1.21	& 94.59$\pm$0.06 &	51.69$\pm$0.58	& 34.81$\pm$0.40 & 79.32$\pm$4.43	& 77.20$\pm$1.84	\\
  & PRC &	51.35 &	50.13 & 51.23 & 74.44$\pm$1.10	& 93.72$\pm$0.09 & 53.65$\pm$1.12	& 41.52$\pm$0.25 & 77.29$\pm$6.35	& 75.84$\pm$1.89	\\
  & bAcc & 51.26 & 50.09 & 78.38 & 67.03$\pm$0.89 & 88.10$\pm$0.31 & 84.83$\pm$0.54 & 49.13$\pm$0.25 & 72.52$\pm$2.33 & 70.32$\pm$1.29	\\
  & Time & 0.19 & 1.45 & 0.16 & 0.29$\pm$0.00 & 14.38$\pm$3.28 & 16.86$\pm$0.22 & 16.22$\pm$0.20 & 53.76$\pm$1.76 &  6.92$\pm$0.12\\
  \midrule
  
  \multirow{6}{*}{svmguide3} & Err & 297 & 297 & 336 & 321.20$\pm$18.07 & 354.00$\pm$2.92 & 308.80$\pm$3.11 & 307.60$\pm$4.39 & 296.60$\pm$0.55	&  321.00$\pm$22.84 \\
  & Acc &	76.11 & 76.11 & 72.97 & 74.16$\pm$1.45	& 71.52$\pm$0.23 &	75.14$\pm$0.25	& 75.23$\pm$0.35 & 76.14$\pm$0.04	&  74.18$\pm$1.84	 \\
  & ROC &	51.50 &	51.45 & 50.18 & 50.07$\pm$1.00	& 58.62$\pm$0.39 &	51.65$\pm$0.29	& 46.16$\pm$0.24 & 52.78$\pm$1.11	&  48.84$\pm$0.74	\\
  & PRC &	76.06 &	77.03 & 75.69 & 74.08$\pm$0.55	& 79.34$\pm$0.38 &	77.07$\pm$0.22	& 74.69$\pm$0.27 & 78.76$\pm$0.44	&  75.09$\pm$0.77	\\
  & bAcc & 49.95 & 49.95 & 53.46 & 51.90$\pm$0.84 & 58.13$\pm$0.41 & 51.02$\pm$0.41 & 49.71$\pm$0.24 & 49.97$\pm$0.03 &  50.61$\pm$0.45	\\
  & Time & 0.18 & 0.33 & 0.19 & 0.33$\pm$0.00 & 7.96$\pm$0.04 & 42.81$\pm$1.82 & 7.93$\pm$0.14 & 52.31$\pm$0.20 &  7.44$\pm$0.08 \\
  \midrule
  
  \multirow{6}{*}{kr-vs-kp} & Err & 1575 & 1523 & 1356 & 1541.80$\pm$16.50 & 907.00$\pm$0.00 & 1327.80$\pm$6.14 & 1831.20$\pm$11.32 & 1309.40$\pm$32.65	& 1287.80$\pm$23.40	\\
  & Acc &	50.72 & 52.35 & 57.57 & 51.76$\pm$0.52	& 71.62$\pm$0.00 & 58.44$\pm$0.19	& 42.69$\pm$0.35 & 59.03$\pm$1.02	& 59.71$\pm$0.73\\
  & ROC &	51.1 & 46.20 & 50.87 & 51.25$\pm$0.30	& 77.87$\pm$0.00 &	50.67$\pm$0.10	& 44.97$\pm$0.07 & 62.78$\pm$0.53	& 62.96$\pm$0.66	\\
  & PRC &	52.69 &	49.32 & 52.64 & 52.85$\pm$0.34	& 76.53$\pm$0.00 &	53.61$\pm$0.04	& 47.93$\pm$0.08 & 63.34$\pm$0.62	& 63.56$\pm$0.62	\\
  & bAcc & 50.7 & 50.29 & 57.44 & 51.50$\pm$0.55	& 71.50$\pm$0.00 &	57.97$\pm$0.19	& 44.18$\pm$0.38 & 58.00$\pm$1.11 & 59.35$\pm$0.75	\\
  & Time & 0.86 & 16.99 & 0.46 & 1.00$\pm$0.08 & 68.52$\pm$0.02 & 75.84$\pm$1.07 & 7.54$\pm$0.08 & 462.21$\pm$322.50 & 22.58$\pm$1.53\\
  \midrule
  
  \multirow{6}{*}{spambase} & Err & 1804.00 & 1815 & 1546 & 2433.40$\pm$18.06 & 1139.80$\pm$13.16 & 1157.00$\pm$7.78 & 1834.20$\pm$4.15 & 1727.20$\pm$86.84 & 1610.00$\pm$98.42\\
  & Acc &	60.79 & 60.55 & 66.4 & 47.11$\pm$0.39 & 75.23$\pm$0.29 & 74.85$\pm$0.17 & 60.13$\pm$0.09 & 62.46$\pm$1.89 & 65.01$\pm$2.14	\\
  & ROC &	49.88 & 49.96 & 58.09 & 59.00$\pm$1.43	& 81.00$\pm$0.17 & 44.11$\pm$0.06	& 46.44$\pm$0.20 & 72.03$\pm$3.63	& 63.35$\pm$1.17	\\
  & PRC &	40.88 & 39.62 & 49.5 & 53.89$\pm$1.04 & 73.39$\pm$0.26 & 34.84$\pm$0.05 & 36.87$\pm$0.17 & 60.11$\pm$4.87 & 57.15$\pm$1.20	\\
  & bAcc & 51.11 & 50.06 & 61.56 & 51.69$\pm$0.44 & 73.38$\pm$0.30 & 71.00$\pm$0.22 & 49.73$\pm$0.09 & 52.85$\pm$2.60 & 61.16$\pm$1.20	\\
  & Time & 1.7 & 31.20 & 0.76 & 1.81$\pm$0.07 & 117.10$\pm$12.53 & 180.98$\pm$0.99 & 32.14$\pm$0.11 & 960.59$\pm$464.82 & 29.88$\pm$1.78 \\

\bottomrule
  \end{tabular}
  }
\end{table*}

%% file: Tables/Tab_Result_Synthetic_p75_small.tex
\begin{table*}[!th]
  \caption{Comparison of models on Small Synthetic Datasets (\underline{$p$ = 0.75}): The deterministic models --- NB3, FAE, and OLVF --- underwent a single execution. In contrast, the non-deterministic models were executed 5 times, with the mean $\pm$ standard deviation reported. Here, Err, Acc, ROC, PRC, bAcc, and Time stand for the number of errors, accuracy, AUROC, AUPRC, balanced accuracy, and execution time, respectively.}
  \label{tab:results_synthetic_p75_small}
  \centering
  \resizebox{\linewidth}{4.3in}{%
  \begin{tabular}{lrccccccccc}
  \toprule
  Dataset & Metric & NB3 & FAE & OLVF & OCDS & OVFM & DynFo & ORF\textsuperscript{3}V & Aux-Net & Aux-Drop\\
  \midrule

  \multirow{6}{*}{WPBC} & Err & 48 & 57 & 63	& 49.00$\pm$0.00 & 82.00$\pm$12.51 & 51.20$\pm$0.84 & 53.00$\pm$0.00 & 49.80$\pm$2.95	& 49.00$\pm$2.55 \\
  & Acc &  75.76 & 71.21  & 68.18	& 75.25$\pm$0.00 & 58.59$\pm$6.32	& 74.01$\pm$0.42 & 73.10$\pm$0.00	& 74.85$\pm$1.49	&  75.25$\pm$1.29 \\
  & ROC &	 52.15 & 51.77  & 52.67	& 42.08$\pm$1.70 & 47.07$\pm$5.65 & 44.38$\pm$0.34 & 45.69$\pm$2.44	& 44.63$\pm$1.33	& 50.06$\pm$2.21 \\
  & PRC &	 25.24 & 24.34	& 32.01	& 20.62$\pm$0.75 & 23.38$\pm$4.52	& 20.09$\pm$0.20 & 22.99$\pm$1.14	& 23.08$\pm$0.63	& 24.17$\pm$1.69	\\
  & bAcc & 49.67 & 47.42	& 48.37	& 49.34$\pm$0.00 & 46.18$\pm$2.05	& 49.64$\pm$0.91 & 50.71$\pm$0.00	& 50.25$\pm$0.81	& 50.51$\pm$0.65	\\
  & Time & 0.07  & 0.12   & 0.03  & 0.05$\pm$0.00  & 1.28$\pm$0.00 & 3.43$\pm$0.09 & 1.00$\pm$0.05 & 22.73$\pm$0.58 & 1.10$\pm$0.06 \\
  \midrule
  
  \multirow{6}{*}{ionosphere } & Err & 127 & 127 & 99	& 121.00$\pm$2.92 & 87.20$\pm$3.56 & 86.60$\pm$2.51	& 180.60$\pm$2.19 & 126.60$\pm$4.04 & 102.40$\pm$5.08	\\
  & Acc &  63.82 &  63.82	& 71.79	& 65.53$\pm$0.83 & 75.16$\pm$1.02	& 75.26$\pm$0.72 & 48.40$\pm$0.63	& 63.93$\pm$1.15 & 70.83$\pm$1.45	\\
  & ROC &	 51.73 &  44.09	& 57.98	& 65.53$\pm$1.37 & 73.13$\pm$1.57	& 68.10$\pm$1.08 & 44.23$\pm$1.36	& 52.84$\pm$1.58 & 70.38$\pm$1.65	\\
  & PRC &	 66.72 &  58.78	& 67.2	& 73.90$\pm$1.20 & 74.49$\pm$0.96	& 76.82$\pm$0.95 & 60.00$\pm$1.05	& 67.24$\pm$0.61 & 78.89$\pm$1.68	\\
  & bAcc & 49.78 &  49.78	& 65.6	& 57.96$\pm$0.75 & 70.25$\pm$0.82	& 66.26$\pm$0.76 & 43.40$\pm$1.02	& 50.32$\pm$1.44 & 61.36$\pm$0.89	\\
  & Time & 0.08 &   0.17  & 0.05  & 0.08$\pm$0.00  & 2.69$\pm$0.01 & 6.06$\pm$0.06 & 1.09$\pm$0.01 & 40.80$\pm$0.35 & 1.87$\pm$0.00\\
  \midrule
  
  \multirow{6}{*}{WDBC} & Err & 214 & 278 & 237 & 257.80$\pm$7.19 & 279.20$\pm$64.07 & 49.40$\pm$1.82	& 312.60$\pm$5.73 & 333.60$\pm$52.88 & 211.80$\pm$3.03 \\
  & Acc &  62.39 & 51.14  & 58.35	& 54.69$\pm$1.26 & 50.93$\pm$11.26 & 91.30$\pm$0.32 & 44.96$\pm$1.01 & 41.37$\pm$9.29 & 62.78$\pm$0.53 \\
  & ROC &	 48.45 & 57.35	& 68.52	& 44.89$\pm$1.41 & 45.35$\pm$14.03 & 35.92$\pm$0.80 & 38.53$\pm$1.15 & 59.87$\pm$0.97 & 45.48$\pm$6.99 \\
  & PRC &	 36.47 & 48.68	& 66.09	& 38.75$\pm$2.23 & 47.47$\pm$11.39 & 28.60$\pm$0.37	& 31.86$\pm$0.57 & 44.78$\pm$3.32 & 36.31$\pm$3.38 \\
  & bAcc & 50.1  & 49.95	& 50.43	& 50.27$\pm$1.43 & 50.53$\pm$11.97 & 90.64$\pm$0.34	& 41.90$\pm$1.32 & 50.04$\pm$0.24	& 50.85$\pm$0.86 \\
  & Time & 0.14  & 0.39   & 0.08  & 0.13$\pm$0.00  & 3.39$\pm$0.01 & 9.73$\pm$0.10 & 3.14$\pm$0.06 & 56.44$\pm$0.54 & 3.06$\pm$0.04 \\
  \midrule

  \multirow{6}{*}{australian} & Err & 289	& 270 & 285 & 364.00$\pm$6.32 & 320.80$\pm$48.46 & 147.00$\pm$2.35 & 404.60$\pm$8.68 & 307.40$\pm$0.55	& 268.40$\pm$21.78 \\
  & Acc &  58.12 & 60.87	& 58.7 &  47.25$\pm$0.92 & 53.51$\pm$7.02	& 78.66$\pm$0.34 & 41.28$\pm$1.26	& 55.45$\pm$0.08 & 61.10$\pm$3.16	\\
  & ROC &	 54.76 & 66.02	& 60.19	& 54.91$\pm$0.95 & 49.10$\pm$9.10	& 39.60$\pm$0.52 & 43.47$\pm$1.03	& 68.07$\pm$5.20 & 64.18$\pm$4.95	\\
  & PRC &	 50.07 & 67.77	& 58.66	& 54.45$\pm$1.09 & 53.57$\pm$7.56	& 36.04$\pm$0.26 & 40.63$\pm$0.93	& 57.71$\pm$6.46	& 60.36$\pm$5.87	\\
  & bAcc & 54.74 & 57	    & 57.11	& 50.30$\pm$0.72 & 53.45$\pm$7.34	& 78.66$\pm$0.36 & 42.86$\pm$1.20	& 49.95$\pm$0.07 & 59.22$\pm$3.86	\\
  & Time & 0.09  & 0.76   & 0.08  & 0.13$\pm$0.00  & 5.24$\pm$0.02 & 11.72$\pm$0.09 & 1.22$\pm$0.08 & 27.54$\pm$0.20 & 3.68$\pm$0.08 \\
  \midrule
  
  \multirow{6}{*}{wbc} & Err & 243 & 297 & 201 & 335.80$\pm$38.46 & 75.00$\pm$0.00	& 39.20$\pm$1.30 & 518.40$\pm$5.81 & 116.80$\pm$23.59	& 102.20$\pm$10.62 \\
  & Acc &  65.24 & 57.51	& 71.24	& 51.96$\pm$5.50 & 89.27$\pm$0.00	& 94.38$\pm$0.19 & 25.73$\pm$0.83	& 83.29$\pm$3.38 & 85.38$\pm$1.52	\\
  & ROC &	 45.46 & 89.54	& 74.92	& 47.97$\pm$4.57 & 94.69$\pm$0.00	& 38.97$\pm$0.56 & 37.69$\pm$0.44	& 87.18$\pm$4.43 & 90.46$\pm$1.50	\\
  & PRC &	 31.18 & 87.41	& 69.03	& 46.29$\pm$6.66 & 92.29$\pm$0.00	& 27.67$\pm$0.26 & 27.22$\pm$0.11	& 82.70$\pm$5.71 & 87.77$\pm$1.81	\\
  & bAcc & 49.88 & 53.32	& 66.46	& 50.35$\pm$3.07 & 88.37$\pm$0.00	& 93.91$\pm$0.35 & 31.18$\pm$1.18	& 78.46$\pm$4.57 & 83.79$\pm$2.17	\\
  & Time & 0.06  & 0.55   & 0.08  & 0.12$\pm$0.00  & 5.06$\pm$0.03 & 26.52$\pm$0.13 & 0.50$\pm$0.00 & 19.72$\pm$0.21 & 3.72$\pm$0.04 \\
  \midrule
  
  \multirow{6}{*}{diabetes-f} & Err & 272	& 315 & 278	& 318.80$\pm$6.83 & 303.00$\pm$70.17 & 258.20$\pm$3.56 & 342.00$\pm$8.57 & 268.60$\pm$0.55	& 304.80$\pm$40.27 \\
  & Acc &  64.58 & 58.98	& 63.8	& 58.49$\pm$0.89 & 60.55$\pm$9.14	& 66.34$\pm$0.46 & 55.41$\pm$1.12	& 65.03$\pm$0.07 & 60.31$\pm$5.24	\\
  & ROC &	 48.02 & 51.85	& 59.49	& 49.54$\pm$1.30 & 59.60$\pm$15.08 & 47.35$\pm$0.19 & 47.30$\pm$0.74 & 49.03$\pm$1.68	& 48.06$\pm$2.54 \\
  & PRC &	 33.08 & 40.33	& 46.1	& 36.17$\pm$0.94 & 46.71$\pm$9.73	& 32.17$\pm$0.11 & 32.72$\pm$0.64	& 34.35$\pm$0.51	& 34.83$\pm$1.66	\\
  & bAcc & 49.6  & 49.28	& 49.17	& 50.34$\pm$0.73 & 55.66$\pm$8.93	& 57.34$\pm$0.63 & 46.70$\pm$1.20	& 49.94$\pm$0.05	& 50.37$\pm$0.85	\\
  & Time & 0.07  & 0.68   & 0.11  & 0.13$\pm$0.00  & 1.55$\pm$0.01 & 12.88$\pm$0.19 & 0.81$\pm$0.01 & 20.33$\pm$0.27 & 4.11$\pm$0.08 \\
  \midrule
  
  \multirow{6}{*}{german} & Err & 284	& 278 & 427	& 537.20$\pm$5.02 & 362.00$\pm$6.08 & 282.20$\pm$2.17 & 274.60$\pm$0.89	& 276.20$\pm$1.64	& 328.60$\pm$40.97 \\
  & Acc &  70.39 & 71.01 & 55.47	& 43.98$\pm$0.52 & 62.25$\pm$0.63	& 70.54$\pm$0.23 & 71.34$\pm$0.09	& 71.20$\pm$0.17	& 65.74$\pm$4.27	\\
  & ROC &	 46.75 & 55.46 & 47.04	& 52.24$\pm$0.84 & 53.67$\pm$0.86	& 51.91$\pm$0.35 & 49.48$\pm$0.34	& 58.29$\pm$0.82	& 51.30$\pm$1.39	\\
  & PRC &	 67.83 & 74.88 & 69.61	& 72.13$\pm$0.24 & 73.26$\pm$0.45	& 71.55$\pm$0.24 & 70.87$\pm$0.32	& 78.28$\pm$0.33	& 72.01$\pm$0.86	\\
  & bAcc & 49.89 & 49.78 & 53.13	& 51.88$\pm$0.95 & 52.56$\pm$0.62	& 50.71$\pm$0.14 & 50.18$\pm$0.15	& 49.98$\pm$0.05	& 51.24$\pm$1.08	\\
  & Time & 0.19  & 0.55  & 0.16   & 0.20$\pm$0.00  & 15.31$\pm$0.04 & 36.00$\pm$0.26 & 1.78$\pm$0.01 & 72.57$\pm$2.86 & 5.30$\pm$0.33 \\
  \midrule
  
  \multirow{6}{*}{IPD} & Err & 548 & 551 & 133 & 303.00$\pm$7.62 & 125.00$\pm$10.75 & 109.60$\pm$2.19 & 560.20$\pm$7.19 & 220.20$\pm$18.71	& 258.60$\pm$18.99 \\
  & Acc &  50	   & 49.73	& 87.86	&  72.35$\pm$0.69 & 88.59$\pm$0.98 & 89.99$\pm$0.20 & 48.84$\pm$0.66 & 79.91$\pm$1.71 & 76.41$\pm$1.73 \\
  & ROC &	 51.56 & 50.13	& 50.03	&  81.16$\pm$0.87 & 94.95$\pm$1.47 & 53.07$\pm$0.54 & 33.01$\pm$0.26 & 88.39$\pm$1.00	& 83.17$\pm$2.49 \\
  & PRC &	 52.46 & 50.37	& 50.31	&  80.79$\pm$0.86 & 93.15$\pm$2.67 & 53.89$\pm$0.69 & 40.83$\pm$0.21 & 87.22$\pm$0.92	& 81.14$\pm$3.39 \\
  & bAcc & 50.06 & 49.82	& 87.87	&  72.35$\pm$0.69 & 88.59$\pm$0.98 & 90.00$\pm$0.20 & 48.72$\pm$0.65 & 79.94$\pm$1.70	& 76.42$\pm$1.73 \\
  & Time & 0.23  & 1.73   & 0.14  &  0.23$\pm$0.00  & 5.43$\pm$0.01 & 18.73$\pm$0.09 & 24.29$\pm$0.06 & 82.54$\pm$1.82 & 6.41$\pm$0.48 \\
  \midrule
  
  \multirow{6}{*}{svmguide3} & Err & 297 & 297 & 359 & 311.60$\pm$7.57 & 356.40$\pm$3.51 & 303.60$\pm$0.55 & 298.60$\pm$1.82 & 296.60$\pm$0.55	& 326.60$\pm$31.82 \\
  & Acc &  76.11 & 76.11	& 71.12	& 74.93$\pm$0.61 & 71.33$\pm$0.28	& 75.56$\pm$0.04 & 75.96$\pm$0.15	& 76.14$\pm$0.04	& 73.72$\pm$2.56	\\
  & ROC &	 48.07 & 51.37	& 53.37	& 49.59$\pm$0.88 & 61.41$\pm$0.44	& 52.19$\pm$0.26 & 46.60$\pm$0.41	& 51.76$\pm$0.50	& 48.80$\pm$1.13	\\
  & PRC &	 75.13 & 76.37	& 75.75	& 73.93$\pm$0.77 & 79.88$\pm$0.45	& 77.55$\pm$0.14 & 74.15$\pm$0.18	& 77.00$\pm$1.44	& 75.02$\pm$0.95	\\
  & bAcc & 49.95 & 49.95	& 53.64	& 50.36$\pm$0.46 & 59.70$\pm$0.30	& 50.60$\pm$0.09 & 49.86$\pm$0.10	& 49.97$\pm$0.03	& 50.68$\pm$0.68	\\
  & Time & 0.18  & 0.4    & 0.16  & 0.25$\pm$0.00  & 7.74$\pm$0.04 & 46.43$\pm$0.20 & 10.71$\pm$0.16 & 81.69$\pm$3.73 & 7.49$\pm$0.27 \\
  \midrule
  
  \multirow{6}{*}{kr-vs-kp} & Err & 1611 & 1529 & 1090 & 1565.00$\pm$23.30 & 620.00$\pm$0.00 & 1306.00$\pm$4.30 & 1838.40$\pm$7.02 & 1138.60$\pm$64.20 & 1135.00$\pm$7.35 \\
  & Acc &  49.59 & 52.16	& 65.89	& 51.03$\pm$0.73 & 80.60$\pm$0.00	& 59.12$\pm$0.13 & 42.46$\pm$0.22	& 64.37$\pm$2.01 & 64.49$\pm$0.23 \\
  & ROC &	 50.47 & 44.8 & 49.95	& 50.67$\pm$0.30 & 87.55$\pm$0.00	& 50.99$\pm$0.08 & 42.79$\pm$0.09	& 70.71$\pm$1.08 & 69.55$\pm$0.85 \\
  & PRC &	 52.82 & 47.48	& 52.65	& 52.46$\pm$0.27 & 86.78$\pm$0.00	& 53.71$\pm$0.09 & 46.74$\pm$0.13	& 70.55$\pm$0.82 & 69.72$\pm$1.51 \\
  & bAcc & 49.87 & 50.08	& 65.74	& 50.67$\pm$0.75 & 80.53$\pm$0.00	& 58.72$\pm$0.14 & 43.96$\pm$0.23	& 63.66$\pm$2.23 & 64.22$\pm$0.31 \\
  & Time & 1.04  & 16.42  & 0.43  & 0.74$\pm$0.00  & 335.12$\pm$123.47 & 83.40$\pm$1.08 & 9.74$\pm$0.08 & 390.14$\pm$2.77 & 18.10$\pm$0.82 \\
  \midrule
  
  \multirow{6}{*}{spambase} & Err & 1829 & 1820 & 1605 & 2567.20$\pm$35.21 & 2293.60$\pm$419.01 & 1068.80$\pm$7.01 & 1831.60$\pm$4.45 & 1477.20$\pm$64.36 & 1549.00$\pm$46.02 \\
  & Acc &  60.25 & 60.44  & 65.12	& 44.20$\pm$0.77 & 50.15$\pm$9.11	& 76.77$\pm$0.15 & 60.18$\pm$0.10	& 67.89$\pm$1.40 & 66.33$\pm$1.00	\\
  & ROC &	 42.33 & 50.18	& 57.59	& 60.63$\pm$1.33 & 51.12$\pm$3.08	& 38.56$\pm$0.16 & 44.77$\pm$0.41	& 74.90$\pm$3.78 & 66.29$\pm$1.71	\\
  & PRC &	 35.09 & 40.13	& 50.04	& 55.34$\pm$1.33 & 56.41$\pm$15.69 & 31.68$\pm$0.10 & 35.32$\pm$0.19 & 66.13$\pm$2.75 & 59.41$\pm$1.46 \\
  & bAcc & 49.81 & 49.94	& 59.8	& 49.72$\pm$0.68 & 51.64$\pm$3.01	& 72.59$\pm$0.21 & 49.69$\pm$0.09	& 61.23$\pm$1.66 & 62.75$\pm$1.18	\\
  & Time & 1.85  & 39.06  & 0.65  & 1.35$\pm$0.00  & 490.74$\pm$349.53 & 176.00$\pm$0.77 & 42.81$\pm$0.10 & 1071.04$\pm$9.10 & 27.26$\pm$2.21 \\

\bottomrule
  \end{tabular}
  }
\end{table*}

%% file: Tables/Tab_Result_Synthetic_Medium.tex
\begin{table*}[!th]
  \caption{Comparison of models on medium Synthetic Datasets (\underline{$p$ = 0.25, 0.5, and 0.75}):  The deterministic models --- NB3, FAE, and OLVF --- underwent a single execution. In contrast, the non-deterministic models were executed 5 times, with the mean $\pm$ standard deviation reported. Here, Err, Acc, ROC, PRC, bAcc, and Time stand for the number of errors, accuracy, AUROC, AUPRC, balanced accuracy, and execution time, respectively.}
  \label{tab:results_synthetic_medium}
  \centering
  \resizebox{\linewidth}{!}{%
  \begin{tabular}{lrccccccccc}
  \toprule
  Dataset & Metric & NB3 & FAE & OLVF & OCDS & OVFM & DynFo & ORF\textsuperscript{3}V & Aux-Net & Aux-Drop\\
  \midrule

  \rowcolor{lightgray} \multicolumn{11}{c}{$p$ = 0.25} \vspace{1mm}\\

  \multirow{6}{*}{magic04} & Err & 6694 & 6687 & 7686	& 7039.60$\pm$5.13 & 6341.00$\pm$0.00 & 7789.40$\pm$69.64 & 11800.00$\pm$36.69 & 6978.40$\pm$79.97 & 6156.60$\pm$57.79 \\
  & Acc &	64.81 & 64.84 & 59.59 & 62.99$\pm$0.03	& 66.66$\pm$0.00 & 59.04$\pm$0.37	& 37.96$\pm$0.19 & 63.31$\pm$0.42	& 67.63$\pm$0.30 \\
  & ROC &	49.91 & 48.66 & 44.02 & 48.79$\pm$0.26	& 61.26$\pm$0.00 & 50.29$\pm$0.15	& 48.00$\pm$0.09 & 50.13$\pm$0.25	& 57.99$\pm$0.76 \\
  & PRC &	64.73 & 62.53 & 59.34 & 62.96$\pm$0.22	& 72.67$\pm$0.00 & 65.56$\pm$0.11	& 63.04$\pm$0.05 & 64.80$\pm$0.19	& 68.46$\pm$0.59 \\
  & bAcc & 50.01 & 50.01 & 53.18 & 51.89$\pm$0.10 & 55.76$\pm$0.00 & 52.75$\pm$0.30	& 47.94$\pm$0.22 & 50.09$\pm$0.07	& 56.04$\pm$0.53 \\
  & Time & 1.46 & 72.7 &  2.09 & 4.30$\pm$0.01 & 129.82$\pm$14.06 & 46.59$\pm$5.63 & 952.28$\pm$23.98 & 376.94$\pm$3.42 & 111.51$\pm$7.59 \\
  \midrule

  \multirow{6}{*}{a8a} & Err & 7847 & 7843 & 9056 & 8768.60$\pm$109.75 & 6980.00$\pm$0.00 & 7903.00$\pm$6.71 & 7846.00$\pm$0.00 & 7843.80$\pm$0.45	& 7865.00$\pm$22.56 \\
  & Acc &	75.9 &  75.91 & 72.19 & 73.07$\pm$0.34 & 78.56$\pm$0.00 & 75.73$\pm$0.02	& 75.90$\pm$0.00 & 75.91$\pm$0.00	& 75.85$\pm$0.07 \\
  & ROC &	51.98 & 49.7  & 63.25 & 61.04$\pm$1.41	& 79.10$\pm$0.00 & 50.86$\pm$0.09	& 47.88$\pm$0.05 & 54.45$\pm$0.18	& 55.92$\pm$3.39 \\
  & PRC &	77.2 &	75.82 & 85.07 & 82.31$\pm$0.88	& 91.96$\pm$0.00 & 76.35$\pm$0.06	& 74.49$\pm$0.03 & 77.78$\pm$0.23	& 81.27$\pm$2.08 \\
  & bAcc & 50.01 & 50	& 60.67 & 54.75$\pm$0.87	& 61.58$\pm$0.00 & 50.01$\pm$0.03	& 49.99$\pm$0.00 & 50.00$\pm$0.00	& 50.00$\pm$0.01 \\
  & Time & 22.94 & 25.7  & 4.96  & 19.17$\pm$0.02 & 2207.11$\pm$111.43 & 4394.51$\pm$72.96 & 207.23$\pm$4.25 & 5484.63$\pm$54.75 & 421.91$\pm$119.74 \\
  \midrule

  \rowcolor{lightgray} \multicolumn{11}{c}{$p$ = 0.50} \vspace{1mm}\\

  \multirow{6}{*}{magic04} & Err & 6693 & 6690 & 7407	& 8699.00$\pm$380.40 & 5867.80$\pm$11.34 & 6824.40$\pm$10.95	& 12351.40$\pm$22.33 & 6965.60$\pm$113.62 & 5912.80$\pm$70.86 \\
  & Acc & 64.81 & 64.83 & 61.06 & 54.26$\pm$2.00	& 69.15$\pm$0.06 & 64.12$\pm$0.06 & 35.06$\pm$0.12 & 63.38$\pm$0.60 & 68.91$\pm$0.37\\
  & ROC &	50.51 &	47.97 & 45.38 & 55.43$\pm$1.54	& 67.58$\pm$0.12 &	53.24$\pm$0.04	& 47.27$\pm$0.07 & 49.90$\pm$0.34	& 62.39$\pm$0.34	\\
  & PRC &	65.21 &	61.36 & 60.29 & 66.40$\pm$1.20	& 77.06$\pm$0.13 &	67.65$\pm$0.02	& 62.24$\pm$0.09 & 64.61$\pm$0.22	& 71.25$\pm$0.31	\\
  & bAcc & 50.02 & 50.00 & 54.6 & 53.40$\pm$0.45	& 61.08$\pm$0.06 &	55.12$\pm$0.06	& 48.56$\pm$0.11 & 50.09$\pm$0.03 & 59.29$\pm$0.48	 \\
  & Time & 1.6 & 72.16 & 2.69 & 5.44$\pm$0.65 & 87.05$\pm$3.38 & 1110.78$\pm$38.39 & 1109.44$\pm$17.12 & 478.10$\pm$30.42 & 116.15$\pm$1.19 \\
  \midrule

  \multirow{6}{*}{a8a} & Err & 7842 & 7843 & 7512 & 8390.60$\pm$207.37 & 6236.00$\pm$0.00 & 7865.40$\pm$3.65 & 7844.00$\pm$0.00 & 7843.80$\pm$0.84 & 7377.00$\pm$178.05 \\
  & Acc & 75.92 & 75.91 & 76.93 & 74.23$\pm$0.64	& 80.85$\pm$0.00 &	75.84$\pm$0.01	& 75.91$\pm$0.00 & 75.91$\pm$0.00	& 77.34$\pm$0.55\\
  & ROC &	51 & 49.74 & 69.64 & 73.56$\pm$0.97 & 84.44$\pm$0.00 & 51.96$\pm$0.08 & 47.28$\pm$0.04 & 58.66$\pm$0.65	& 72.34$\pm$2.56\\
  & PRC &	76.55 &	75.84 & 88.6 & 89.40$\pm$0.51	& 94.41$\pm$0.00 & 76.88$\pm$0.06	& 74.27$\pm$0.02 & 79.40$\pm$0.26	& 89.28$\pm$0.93\\
  & bAcc & 50.01 & 50.00 & 66.46 & 64.04$\pm$1.01 & 68.57$\pm$0.00 & 50.11$\pm$0.01 & 50.01$\pm$0.00 & 50.00$\pm$0.00 & 55.33$\pm$1.99\\
  & Time & 23.52 & 26.18 & 6.59 & 51.67$\pm$3.29 & 4120.49$\pm$415.07 & 4858.85$\pm$133.63 & 395.44$\pm$0.80 & 17639.23$\pm$2014.52 & 235.54$\pm$8.39\\
  \midrule

  \rowcolor{lightgray} \multicolumn{11}{c}{$p$ = 0.75} \vspace{1mm}\\

  \multirow{6}{*}{magic04} & Err & 6691 & 6689 & 7168	& 8591.00$\pm$406.63 & 6743.20$\pm$28.31 & 6356.60$\pm$1.52 & 12397.40$\pm$6.58 & 6977.20$\pm$136.68	& 5580.80$\pm$72.82 \\
  & Acc &  64.82 & 64.83	& 62.31	& 54.83$\pm$2.14 & 64.55$\pm$0.15	& 66.58$\pm$0.01 & 34.82$\pm$0.03 & 63.32$\pm$0.72	& 70.66$\pm$0.38	\\
  & ROC &	 49.34 & 46.5	& 46.89	& 55.90$\pm$1.74 & 61.15$\pm$0.02	& 53.79$\pm$0.04 & 44.26$\pm$0.13 & 50.16$\pm$0.24	& 68.48$\pm$0.92	\\
  & PRC &	 64.15 & 59.67	& 61.4	& 67.60$\pm$1.39 & 75.98$\pm$0.01	& 67.60$\pm$0.01 & 59.68$\pm$0.14 & 64.88$\pm$0.33	& 75.93$\pm$0.43	\\
  & bAcc & 49.99 & 50	& 56.19	& 53.76$\pm$1.07 & 60.78$\pm$0.12	& 56.75$\pm$0.02 & 49.32$\pm$0.04 & 50.05$\pm$0.07	& 63.18$\pm$0.61	\\
  & Time & 1.78  & 51.8   & 2.16  & 3.20$\pm$0.00  & 129.95$\pm$23.78 & 1134.59$\pm$13.96 & 1259.35$\pm$21.46 & 693.89$\pm$19.41 & 122.99$\pm$2.65 \\
  \midrule

  \multirow{6}{*}{a8a} & Err & 7842 & 7843 & 6622	& 8279.80$\pm$238.28 & 5594.00$\pm$0.00 & 7863.00$\pm$4.06 & 7852.00$\pm$0.00	& 7843.40$\pm$0.55 & 6533.00$\pm$111.55 \\
  & Acc &  75.92 & 75.91	& 79.66	& 74.57$\pm$0.73 & 82.82$\pm$0.00	& 75.85$\pm$0.01 & 75.88$\pm$0.00	& 75.91$\pm$0.00 & 79.94$\pm$0.34	\\
  & ROC &	 50.88 & 49.74	& 72.43	& 77.80$\pm$0.57 & 87.33$\pm$0.00	& 52.04$\pm$0.08 & 46.94$\pm$0.05	& 62.44$\pm$0.51 & 79.66$\pm$1.14	\\
  & PRC &	 76.16 & 75.84	& 90	& 91.64$\pm$0.31 & 95.57$\pm$0.00	& 77.11$\pm$0.06 & 74.07$\pm$0.03	& 81.26$\pm$0.30 & 92.35$\pm$0.43	\\
  & bAcc & 50.01 & 50	    & 70.63	& 68.81$\pm$1.10 & 72.70$\pm$0.00	& 50.13$\pm$0.01 & 49.99$\pm$0.00	& 50.00$\pm$0.00 & 62.87$\pm$0.93	\\
  & Time & 28.16 & 37.84  & 6.37  & 32.28$\pm$3.32 & 2929.01$\pm$1001.51 & 5263.24$\pm$92.75 & 569.20$\pm$11.54 & 28776.93$\pm$2045.19 & 577.31$\pm$768.95 \\  

  \bottomrule
  \end{tabular}
  }
\end{table*}

%% file: Tables/Tab_Result_Synthetic_Large.tex
\begin{table*}[!th]
  \caption{Comparison of models on large synthetic Datasets (\underline{$p$ = 0.25, 0.5, and 0.75}): The deterministic models --- NB3, FAE, and OLVF --- underwent a single execution. In contrast, the non-deterministic models were executed 5 times, with the mean $\pm$ standard deviation reported. Here, Err, Acc, ROC, PRC, bAcc, and Time stand for the number of errors, accuracy, AUROC, AUPRC, balanced accuracy, and execution time, respectively. A \textsuperscript{$\ddagger$} symbol indicates non-deterministic models that were run only once on specific datasets due to substantial time constraints.}
  \label{tab:results_synthetic_large}
  \centering
  \resizebox{\linewidth}{!}{%
  \begin{tabular}{lrccccccccc}
  \toprule
  Dataset & Metric & NB3 & FAE & OLVF & OCDS & OVFM & DynFo & ORF\textsuperscript{3}V & Aux-Net & Aux-Drop\\
  \midrule

  \rowcolor{lightgray} \multicolumn{11}{c}{$p$ = 0.25} \vspace{1mm}\\

  \multirow{6}{*}{SUSY} & Err & 457957 & 529137 & 504062 & 497382.00$\pm$2960.08 & 388255.40$\pm$24.74 & 451496.67$\pm$76.94 & 495106.60$\pm$134.17 & 456143.60$\pm$9290.25 & 362096.60$\pm$871.73 \\
  & Acc &	54.2 &  47.09   &  49.59 & 50.26$\pm$0.30 & 61.17$\pm$0.00 & 54.85$\pm$0.01 & 50.49$\pm$0.01	& 54.39$\pm$0.93 & 63.79$\pm$0.09 \\
  & ROC &	49.88 & 50.41   & 56.26	 & 55.93$\pm$0.32 & 65.38$\pm$0.00 & 50.18$\pm$0.00	& 49.42$\pm$0.01	& 51.03$\pm$1.80 & 69.26$\pm$0.12 \\
  & PRC &	45.68 &	46.06   & 57.31	 & 53.64$\pm$0.94 & 62.93$\pm$0.00 & 45.94$\pm$0.00 & 45.23$\pm$0.01	& 46.93$\pm$2.26 & 67.50$\pm$0.09 \\
  & bAcc & 50 &   49.9    & 51.12	 & 52.11$\pm$0.19 & 59.72$\pm$0.00 & 54.68$\pm$0.01 & 49.37$\pm$0.01	& 50.53$\pm$1.17 & 61.98$\pm$0.10 \\
  & Time & 68.97 &12100.93& 105.85 & 163.96$\pm$1.52& 3309.30$\pm$184.51 & 654852.68$\pm$8956.39 & 24252.71$\pm$37.64 & 16455.42$\pm$167.84 & 5773.05$\pm$129.01 \\
  \midrule

  \multirow{6}{*}{HIGGS} & Err & 470403 & 480197 & 482630 & 500440.40$\pm$475.98 & 472764.20$\pm$31.21 & 492724\textsuperscript{$\ddagger$}	& 503204.00$\pm$312.46 & 472809.80$\pm$233.06 & 464302.00$\pm$186.18 \\
  & Acc &	52.96 & 51.98 & 51.74 & 49.96$\pm$0.05	& 52.72$\pm$0.00 & 50.73\textsuperscript{$\ddagger$} & 49.68$\pm$0.03	& 52.72$\pm$0.02 & 53.57$\pm$0.02 \\
  & ROC &	50.03 & 50.09 & 49.99 & 49.96$\pm$0.07	& 51.54$\pm$0.00 & 49.99\textsuperscript{$\ddagger$} & 50.00$\pm$0.01	& 50.04$\pm$0.02 & 52.36$\pm$0.17 \\
  & PRC &	52.97 & 52.98 & 53.1  & 52.82$\pm$0.06	& 54.09$\pm$0.00 & 52.98\textsuperscript{$\ddagger$} & 53.02$\pm$0.01	& 53.01$\pm$0.02 & 54.93$\pm$0.16 \\
  & bAcc & 50 &   50.16 & 50.57 & 49.97$\pm$0.07	& 50.97$\pm$0.00 & 50.18\textsuperscript{$\ddagger$} & 49.86$\pm$0.03	& 49.99$\pm$0.00 & 51.17$\pm$0.05 \\
  & Time & 136.12 & 17777.05 & 112.34 & 201.20$\pm$12.85 & 8675.39$\pm$335.33 & 1845662.94\textsuperscript{$\ddagger$} & 42814.64$\pm$94.35 & 28363.61$\pm$153.32 &  5784.12$\pm$159.85 \\
  \midrule

  \rowcolor{lightgray} \multicolumn{11}{c}{$p$ = 0.50} \vspace{1mm}\\

  \multirow{6}{*}{SUSY} & Err & 457957 & 531338 & 483988 & 482305.60$\pm$3558.98 & 336800.80$\pm$14.81 & 412369.00$\pm$36.87 & 515823.80$\pm$285.70 & 392908.20$\pm$60770.38 & 300952.40$\pm$1155.92\\
  & Acc & 54.2 & 46.87 & 51.6	& 51.77$\pm$0.36 & 66.32$\pm$0.00	& 58.76$\pm$0.00 & 48.42$\pm$0.03 & 60.71$\pm$6.08 & 69.90$\pm$0.12\\
  & ROC &	50.03 &	50.22 & 61.86 & 57.95$\pm$0.32	& 71.56$\pm$0.00 & 50.04$\pm$0.00 & 49.06$\pm$0.01 & 60.14$\pm$8.30 & 76.48$\pm$0.15\\
  & PRC &	45.79 &	45.51 & 62.65 & 54.01$\pm$0.65	& 69.05$\pm$0.00 & 45.88$\pm$0.00	& 44.85$\pm$0.01 & 57.49$\pm$10.17 & 75.44$\pm$0.10\\
  & bAcc & 50	& 50.01 & 53.21	& 54.03$\pm$0.28 & 65.60$\pm$0.00	& 58.27$\pm$0.00 & 48.33$\pm$0.02 & 57.89$\pm$7.19 & 68.79$\pm$0.14\\
  & Time & 75.86 & 10731.91 & 135.96 & 255.58$\pm$0.50 & 4210.01$\pm$453.61 & 343724.64$\pm$5176.04 & 23928.85$\pm$169.87 & 19612.72$\pm$567.55 & 6054.62$\pm$660.84\\
  \midrule

  \multirow{6}{*}{HIGGS} & Err & 470388 & 474407 & 477760 & 499377.20$\pm$487.22 & 467153.60$\pm$33.89 & 492219\textsuperscript{$\ddagger$}	& 512329.40$\pm$189.51 & 472704.40$\pm$358.46 & 451891.40$\pm$365.32\\
  & Acc & 52.96 & 52.56 & 52.22 & 50.06$\pm$0.05	& 53.28$\pm$0.00 & 50.78\textsuperscript{$\ddagger$} & 48.77$\pm$0.02 & 52.73$\pm$0.04 & 54.81$\pm$0.04\\
  & ROC &	49.97 &	50.12 & 50.29 & 50.04$\pm$0.08	& 52.97$\pm$0.00 & 50.01\textsuperscript{$\ddagger$} & 49.98$\pm$0.01 & 50.05$\pm$0.01 & 55.60$\pm$0.06\\
  & PRC &	52.9 & 53 & 53.29 & 52.86$\pm$0.05	& 55.29$\pm$0.00 & 52.99\textsuperscript{$\ddagger$} & 52.96$\pm$0.01 & 53.01$\pm$0.02	& 57.81$\pm$0.12\\
  & bAcc & 50 & 50.01 & 51.21	& 50.06$\pm$0.06 & 51.83$\pm$0.00 & 50.21\textsuperscript{$\ddagger$} & 49.82$\pm$0.02 & 49.99$\pm$0.01	& 53.09$\pm$0.05\\
  & Time & 142.97 & 7885.78 & 145.83 & 267.35$\pm$0.54 & 10342.77$\pm$1686.50 & 1788308.41\textsuperscript{$\ddagger$} & 42623.97$\pm$97.07 & 44123.61$\pm$283.06 & 6039.45$\pm$565.24\\

  \midrule

  \rowcolor{lightgray} \multicolumn{11}{c}{$p$ = 0.75} \vspace{1mm}\\

  \multirow{6}{*}{SUSY} & Err & 457952 & 529567 & 452821 & 477624.00$\pm$5394.23	& 302880.20$\pm$3.11 & 383919.4$\pm$129 & 524984.20$\pm$345.24 & 429402.00$\pm$69311.90	& 255994.60$\pm$801.85	\\
  & Acc &  54.2  & 47.04	& 54.72	& 52.24$\pm$0.54 & 69.71$\pm$0.00	& 61.61$\pm$0.01 & 47.50$\pm$0.03	& 57.06$\pm$6.93 & 74.40$\pm$0.08	\\
  & ROC &	 49.96 & 49.51	& 65.15	& 58.41$\pm$0.32 & 72.26$\pm$0.00	& 49.91$\pm$0.00 & 48.36$\pm$0.02 & 55.41$\pm$10.29	& 81.07$\pm$0.07	\\
  & PRC &	 45.75 & 46.43	& 65.74	& 53.39$\pm$0.40 & 74.04$\pm$0.00	& 45.8$\pm$0.00 & 44.28$\pm$0.02	& 50.87$\pm$10.38	& 80.44$\pm$0.11	\\
  & bAcc & 50 & 50.12	& 55.98	& 54.84$\pm$0.48 & 68.52$\pm$0.00	& 60.94$\pm$0.01 & 47.53$\pm$0.03 & 53.67$\pm$8.13 & 73.55$\pm$0.11	\\
  & Time & 83.59 & 12324.28 & 111.37 & 166.07$\pm$0.20 & 2201.85$\pm$12.91 & 114179.9$\pm$2433.89 & 25690.16$\pm$79.67 & 23987.22$\pm$1326.00 & 5787.78$\pm$123.95 \\
  \midrule

  \multirow{6}{*}{HIGGS} & Err & 470336 & 507591	& 472523 & 500300.00$\pm$475.25 & 462743.40$\pm$2.30	& 492196\textsuperscript{$\dagger$} & 521619.00$\pm$301.39 & 472790\textsuperscript{$\ddagger$} & 432298.00$\pm$791.48 \\
  & Acc &  52.97 & 49.24	& 52.75	& 49.97$\pm$0.05 & 53.73$\pm$0.00	& 50.78\textsuperscript{$\dagger$} & 47.84$\pm$0.03	& 52.72\textsuperscript{$\ddagger$} & 56.77$\pm$0.08	\\
  & ROC &	 49.95 & 50.1 & 50.66	& 49.98$\pm$0.02 & 54.01$\pm$0.00	& 49.94\textsuperscript{$\dagger$} & 49.88$\pm$0.01 & 50.06\textsuperscript{$\ddagger$} & 58.96$\pm$0.18	\\
  & PRC &	 52.94 & 53.48 & 53.66	& 52.80$\pm$0.02 & 56.07$\pm$0.00	& 52.91\textsuperscript{$\dagger$} & 52.89$\pm$0.00 & 53.04\textsuperscript{$\ddagger$} & 60.78$\pm$0.22	\\
  & bAcc & 50	& 50.55	& 51.98	& 49.97$\pm$0.05 & 52.66$\pm$0.00	& 50.16\textsuperscript{$\dagger$} & 49.75$\pm$0.03 & 49.98\textsuperscript{$\ddagger$} & 55.55$\pm$0.11	\\
  & Time & 153.05 & 549606.27 & 125.45 &   200.58$\pm$2.95 & 7407.39$\pm$295.04 & 801655.55\textsuperscript{$\dagger$} & 44079.06$\pm$134.05 & 65500.41\textsuperscript{$\ddagger$} & 5762.40$\pm$170.11 \\

  \bottomrule
  \end{tabular}
  }
\end{table*}

%% file: Tables/Tab_Result_Real.tex
\begin{table*}[!ht]
  \caption{Comparison of models on \underline{Real} Datasets: The deterministic models --- NB3, FAE, and OLVF --- underwent a single execution. In contrast, the non-deterministic models were executed 5 times, with the mean $\pm$ standard deviation reported. Here, Err, Acc, ROC, PRC, bAcc, and Time stand for the number of errors, accuracy, AUROC, AUPRC, balanced accuracy, and execution time, respectively. A \textsuperscript{$\ddagger$} symbol indicates non-deterministic models that were run only once on specific datasets due to substantial time constraints.}
  \label{tab:results_real_data}
  \centering
  \begin{adjustbox}{width=\textwidth,center}
  \begin{tabular}{lrccccccccc}
  \toprule
  
  Dataset & Metric & NB3 & FAE & OLVF & OCDS & OVFM & DynFo & ORF\textsuperscript{3}V & Aux-Net & Aux-Drop\\
  \midrule

  \rowcolor{lightgray} \multicolumn{11}{c}{Small} \vspace{1mm}\\
  \multirow{6}{*}{crowdsense(c3)} & Err & 8.00 & 7.00 & 7 & 12.80$\pm$0.45 & 312.60$\pm$28.34 & 10.00$\pm$1.58 & 8.60$\pm$0.55 & 164.60$\pm$57.87 & 48.60$\pm$26.93 \\
  & Acc &	98.99 & 99.11 & 99.11 & 98.37$\pm$0.06	& 60.23$\pm$3.61 & 98.73$\pm$0.20 & 98.90$\pm$0.07 & 79.06$\pm$7.36 &  93.82$\pm$3.43\\
  & ROC &	99.65 &	2.35  & 40.97 & 99.78$\pm$0.01	& 87.64$\pm$4.25 & 89.55$\pm$0.92 & 98.88$\pm$0.11  & 87.06$\pm$2.68 & 97.00$\pm$0.78\\
  & PRC &	99.97 &	77.70 & 85.72 & 99.98$\pm$0.00	& 98.91$\pm$0.37 & 99.00$\pm$0.10 & 99.90$\pm$0.01  & 98.78$\pm$0.27 &  99.18$\pm$1.11\\
  & bAcc & 99.45 & 99.51 & 99.51 & 99.11$\pm$0.03 & 78.23$\pm$1.97 &  95.79$\pm$1.11 & 99.40$\pm$0.04	& 87.87$\pm$3.76 & 93.15$\pm$2.16\\
  & Time & 4.14 & 9.66 & 0.28 & 39.84$\pm$0.47 & 354.31$\pm$4.99 & 69.89$\pm$0.62 & 33.59$\pm$0.06 & 4909.36$\pm$752.39 &  464.23$\pm$0.92\\
  \midrule

  \multirow{6}{*}{crowdsense(c5)} & Err & 15.00 & 32.00 & 8 & 14.00$\pm$1.22 & 310.60$\pm$42.72 & 23.40$\pm$1.52 & 11.40$\pm$0.89 & 87.00$\pm$1.00 & 78.00$\pm$32.11\\
  & Acc &	98.10 & 95.94 & 98.99 & 98.22$\pm$0.16	& 60.48$\pm$5.44 & 97.02$\pm$0.19 & 98.55$\pm$0.11 & 88.93$\pm$0.13	& 90.08$\pm$4.09\\
  & ROC &	97.38 &	56.46 & 44.05 & 94.65$\pm$0.28	& 62.90$\pm$4.19 & 27.51$\pm$1.76 & 71.07$\pm$1.51 & 68.71$\pm$25.08 & 90.95$\pm$3.36\\
  & PRC &	86.63 &	12.48 & 9.91 & 92.21$\pm$0.33 & 25.16$\pm$11.06 & 7.08$\pm$0.30 & 36.31$\pm$0.37 & 19.81$\pm$9.75	& 46.17$\pm$7.10\\
  & bAcc & 94.85 & 93.64 & 97.9 & 94.10$\pm$0.35	& 61.90$\pm$6.16 & 94.15$\pm$0.27 & 95.62$\pm$0.41 & 49.93$\pm$0.07	& 83.72$\pm$5.48\\
  & Time & 4.02 & 58.52 & 0.27 & 43.73$\pm$0.61 & 354.56$\pm$5.32 & 68.67$\pm$0.39 & 33.80$\pm$0.09 & 5563.30$\pm$498.90 & 35.72$\pm$4.56\\
  \midrule

  \multirow{6}{*}{spamassasin} & Err & 267 & 152 & 15	& 137.00$\pm$0.00 & 31.60$\pm$7.23 & 182.20$\pm$11.90 & 225.60$\pm$1.67 & 103\textsuperscript{$\ddagger$} & 1439.20$\pm$15.45\\
  & Acc & 95.58 & 97.49 & 99.75 & 97.73$\pm$0.00	& 99.48$\pm$0.12 & 96.99$\pm$0.20 & 96.27$\pm$0.03	& 98.30\textsuperscript{$\ddagger$} & 76.19$\pm$0.26\\
  & ROC &	93.06 & 2.56 & 51.34 & 99.47$\pm$0.00 & 99.37$\pm$0.03 & 60.48$\pm$1.31 & 98.48$\pm$0.04 & 98.31\textsuperscript{$\ddagger$} & 59.34$\pm$0.40\\
  & PRC &	95.05 &	19.36 & 31.84 & 99.59$\pm$0.00	& 99.40$\pm$0.03 & 50.79$\pm$1.92 & 97.94$\pm$0.04	& 98.09\textsuperscript{$\ddagger$} & 53.07$\pm$0.46\\
  & bAcc & 92.96 & 95.99 & 99.6 & 96.39$\pm$0.00	& 99.28$\pm$0.09 & 97.61$\pm$0.14 & 94.05$\pm$0.04	& 97.28\textsuperscript{$\ddagger$} & 62.08$\pm$0.39\\
  & Time & 254.32	& 5880.89 & 4.96 & 8809.23$\pm$856.44 & 32739.41$\pm$3369.92 & 2176.94$\pm$22.49 & 1741.31$\pm$22.76 & 67934.42\textsuperscript{$\ddagger$} & 18071.34$\pm$685.19\\
  \midrule

  \rowcolor{lightgray} \multicolumn{11}{c}{Medium} \vspace{1mm}\\
  \multirow{6}{*}{imdb} & Err & 4609 & 4455 & 4980 & 12464.20$\pm$6.14 & 5646\textsuperscript{$\ddagger$} & 10504.00$\pm$73.27 & 5882.80$\pm$27.38 & 8148\textsuperscript{$\ddagger$} & 6725.60$\pm$48.12\\
  & Acc & 81.56 & 82.18 & 80.08 & 50.14$\pm$0.02	& 77.42\textsuperscript{$\ddagger$} & 57.98$\pm$0.29	& 76.47$\pm$0.11 & 67.41\textsuperscript{$\ddagger$} & 73.10$\pm$0.19\\
  & ROC &	88.08 & 46.17 & 48.57 & 50.04$\pm$0.01	& 83.9\textsuperscript{$\ddagger$} & 43.11$\pm$0.14	& 55.53$\pm$0.02 & 71.34\textsuperscript{$\ddagger$} & 79.67$\pm$0.29\\
  & PRC &	84.93 &	47.95 & 48.29 & 50.13$\pm$0.02	& 82.46\textsuperscript{$\ddagger$} & 45.68$\pm$0.15	& 55.28$\pm$0.01 & 71.27\textsuperscript{$\ddagger$} & 77.74$\pm$0.40\\
  & bAcc & 81.56 & 82.18 & 80.08 & 50.14$\pm$0.02 & 77.42\textsuperscript{$\ddagger$} & 57.98$\pm$0.29 & 76.47$\pm$0.11 & 67.41\textsuperscript{$\ddagger$} & 73.10$\pm$0.19\\
  & Time & 1257.01 & 4117.33 & 22.17 & 47818.07$\pm$2783.21 & 134556.64\textsuperscript{$\ddagger$} & 4072.86$\pm$138.29 & 2768.24$\pm$23.26 & 223699.04\textsuperscript{$\ddagger$} & 46437.42$\pm$18114.58\\
  \midrule

  \rowcolor{lightgray} \multicolumn{11}{c}{Large} \vspace{1mm}\\
  \multirow{6}{*}{diabetes\_us} & Err & 11357 & 11508 & 15971 & 31298.80$\pm$851.84 & 31608.20$\pm$10591.15 & 11357.00$\pm$0.00 & 11357.00$\pm$0.00 & 11357.20$\pm$0.45 & 11357.60$\pm$0.89 \\
  & Acc &	88.84 & 88.69 & 84.31 & 69.24$\pm$0.84	& 68.94$\pm$10.41 & 88.84$\pm$0.00 & 88.84$\pm$0.00 & 88.84$\pm$0.00	& 88.84$\pm$0.00\\
  & ROC &	50.84 &	53.29 & 49.09 & 50.55$\pm$0.10	& 49.88$\pm$1.11 & 47.23$\pm$0.45 & 49.18$\pm$0.17	& 49.04$\pm$0.21	& 50.04$\pm$0.26\\
  & PRC &	11.43 &	13.02 & 11.02 & 11.30$\pm$0.06	& 22.34$\pm$6.71 & 10.19$\pm$0.15 & 10.98$\pm$0.05	& 10.94$\pm$0.07	& 11.10$\pm$0.08\\
  & bAcc & 50	& 50.08 & 50.11 & 50.27$\pm$0.14	& 50.35$\pm$0.74 & 50.00$\pm$0.00 & 50.00$\pm$0.00	& 50.00$\pm$0.00	& 50.00$\pm$0.00\\
  & Time & 41.55 & 32727.48 & 16.47 & 47.96$\pm$0.24 & 4251.94$\pm$210.06 & 487.49$\pm$1.34 & 491.84$\pm$11.12 & 25760.59$\pm$105.48 & 567.92$\pm$3.34\\

\bottomrule
  \end{tabular}
  \end{adjustbox}
\end{table*}